\pdfoutput=1

\documentclass[acmsmall, nonacm]{acmart} % for arXiv

\usepackage{kotex}
\usepackage{comment}
\usepackage{graphicx}
\usepackage{subcaption}
\usepackage{bbm}
\usepackage[algo2e,ruled,vlined]{algorithm2e}
\usepackage{svg}
\usepackage{algorithm}
\usepackage{todonotes}
\usepackage{subcaption}
\usepackage{booktabs}
\usepackage{multirow}
\usepackage{color, colortbl}
\usepackage{mathtools}
\usepackage{booktabs,caption}
\usepackage[flushleft]{threeparttable}
\usepackage{lipsum}

\usepackage{multirow}
\usepackage{amsmath}
\usepackage{makecell}
\usepackage{listings}
\usepackage{xcolor}
\usepackage{array}
\usepackage{algpseudocode}
\usepackage{caption}
\usepackage{fancyvrb}
\usepackage{hyperref}
\usepackage{tabularx}
\usepackage{longtable}
\usepackage{xtab}
\usepackage{microtype}
\usepackage{varwidth} % varwidth 패키지 선언
\usepackage[breakable]{tcolorbox}
\usepackage{wrapfig}

\newcommand{\revise}[1]{\textcolor{black}{#1}}

\AtBeginDocument{%
  \providecommand\BibTeX{{%
    \normalfont B\kern-0.5em{\scshape i\kern-0.25em b}\kern-0.8em\TeX}}}

\setcopyright{rightsretained}
\begin{document}

%%
%% The "title" command has an optional parameter,
%% allowing the author to define a "short title" to be used in page headers.
% \title{LLM on Abstraction and Reasoning Corpus}%
\title{Reasoning Abilities of Large Language Models: In-Depth Analysis on the Abstraction and Reasoning Corpus}

%%
%% The "author" command and its associated commands are used to define
%% the authors and their affiliations.
%% Of note is the shared affiliation of the first two authors, and the
%% "authornote" and "authornotemark" commands
%% used to denote shared contribution to the research.

\author{Seungpil Lee}
\authornote{The first three authors contributed equally and should be considered co-first authors.}
\email{iamseungpil@gm.gist.ac.kr}
\orcid{0009-0007-1133-0848}
\affiliation{%
  \institution{Electrical Engineering and Computer Science, GIST}
  \streetaddress{123 Cheomdangwagi-ro}
  \city{Gwangju}
  \country{Republic of Korea}
  \postcode{61005}
}

\author{Woochang Sim}
\authornotemark[1]
\email{woochang@gm.gist.ac.kr}
\orcid{0009-0001-9215-0123}
\affiliation{%
  \institution{AI Graduate School, GIST}
  % \streetaddress{123 Cheomdangwagi-ro}
  \city{Gwangju}
  \country{Republic of Korea}
  \postcode{61005}
}

\author{Donghyeon Shin}
\authornotemark[1]
\email{dong97411@gm.gist.ac.kr}
\orcid{0009-0005-1587-1268}
\affiliation{%
  \institution{AI Graduate School, GIST}
  % \streetaddress{123 Cheomdangwagi-ro}
  \city{Gwangju}
  \country{Republic of Korea}
  \postcode{61005}
}

\author{Wongyu Seo}
\email{won8617@gmail.com}
\orcid{0009-0003-8982-1031}
\affiliation{%
  \institution{Electrical Engineering and Computer Science, GIST}
  \city{Gwangju}
  \country{Republic of Korea}
  \postcode{61005}
}

\author{Jiwon Park}
\email{parkjohn58@gm.gist.ac.kr}
\orcid{0009-0000-4323-5684}
\affiliation{%
  \institution{AI Graduate School, GIST}
  \city{Gwangju}
  \country{Republic of Korea}
  \postcode{61005}
}

\author{Seokki Lee}
\email{sklee1103@gm.gist.ac.kr}
\orcid{0009-0001-4070-6927}
\affiliation{%
  \institution{AI Graduate School, GIST}
  \city{Gwangju}
  \country{Republic of Korea}
  \postcode{61005}
}

\author{Sanha Hwang}
\email{hsh6449j@gm.gist.ac.kr}
\orcid{0009-0005-4697-2360}
\affiliation{%
  \institution{AI Graduate School, GIST}
  \city{Gwangju}
  \country{Republic of Korea}
  \postcode{61005}
}

\author{Sejin Kim}
\affiliation{%
  \institution{AI Graduate School, GIST}
  \city{Gwangju}
  \country{Republic of Korea}}
\email{sejinkim@gist.ac.kr} 
\orcid{0000-0002-3328-5757}

\author{Sundong Kim}
% \authornote{Corresponding author}
\affiliation{%
  \institution{AI Graduate School, GIST}
  \city{Gwangju}
  \country{Republic of Korea}}
\email{sundong@gist.ac.kr} 
\orcid{0000-0001-9687-2409}
  
%% later remove (added for arXiv)
\makeatletter
\let\@authorsaddresses\@empty
\makeatother

\renewcommand{\shortauthors}{Lee, Sim, and Shin et al.}

\begin{abstract}
\revise{The existing methods for evaluating the inference abilities of Large Language Models (LLMs) have been predominantly results-centric, making it challenging to assess the inference process comprehensively. We introduce a novel approach using the Abstraction and Reasoning Corpus (ARC) benchmark to evaluate the inference and contextual understanding abilities of LLMs in a process-centric manner, focusing on three key components from the Language of Thought Hypothesis (LoTH): Logical Coherence, Compositionality, and Productivity. Our carefully designed experiments reveal that while LLMs demonstrate some inference capabilities, they still significantly lag behind human-level reasoning in these three aspects. The main contribution of this paper lies in introducing the LoTH perspective, which provides a method for evaluating the reasoning process that conventional results-oriented approaches fail to capture, thereby offering new insights into the development of human-level reasoning in artificial intelligence systems.}
\end{abstract}

\begin{CCSXML}
<ccs2012>
   <concept>
       <concept_id>10010147.10010178.10010187.10010198</concept_id>
       <concept_desc>Computing methodologies~Reasoning about belief and knowledge</concept_desc>
       <concept_significance>500</concept_significance>
       </concept>
   <concept>
       <concept_id>10010147.10010257.10010293.10010297</concept_id>
       <concept_desc>Computing methodologies~Logical and relational learning</concept_desc>
       <concept_significance>500</concept_significance>
       </concept>
 </ccs2012>
\end{CCSXML}

\ccsdesc[500]{Computing methodologies~Reasoning about belief and knowledge}
\ccsdesc[500]{Computing methodologies~Logical and relational learning}

\newcommand{\ej}[1]{\textcolor{orange}{#1}}
\newcommand{\sh}[1]{\textcolor{cyan}{#1}}
\newcommand{\fix}[1]{\textcolor{red}{#1}}
%%
%% This command processes the author and affiliation and title
%% information and builds the first part of the formatted document.
\maketitle

% add when submit to arXiv
\begingroup
\footnotesize
\tableofcontents
\endgroup
\vspace{-10mm}
\newpage

\section{Introduction}

Recent Large Language Models (LLMs) have demonstrated performance levels close to that of humans, but experimental results showed that they lacked planning ability through thought or reasoning~\cite{bubeck2023sparks}. Consequently, a key question in recent language model research is: Can LLMs think? To address this question, new benchmarks for measuring reasoning abilities, such as MathVista~\cite{lu2024math}, Bongard-Logo~\cite{nie2020bongard}, and Raven~\cite{zhang2019raven}, have been proposed. Among these, the Abstraction and Reasoning Corpus (ARC)~\cite{chollet2019ARC} emerged to be one of the representative benchmarks for assessing reasoning abilities. As shown in Fig.~\ref{fig:intro/arc_examples} below, each task in ARC consists of 2--5 demonstration example pairs and a test example input grid. The goal is to infer rules from the given demonstration example pairs and apply them to the test example. Input and output grid sizes can vary from a minimum of $1 \times 1$ to a maximum of $30 \times 30$, with each grid having up to 10 different colors.

\begin{figure}[ht]
    \centering
    \includegraphics[width=\columnwidth]{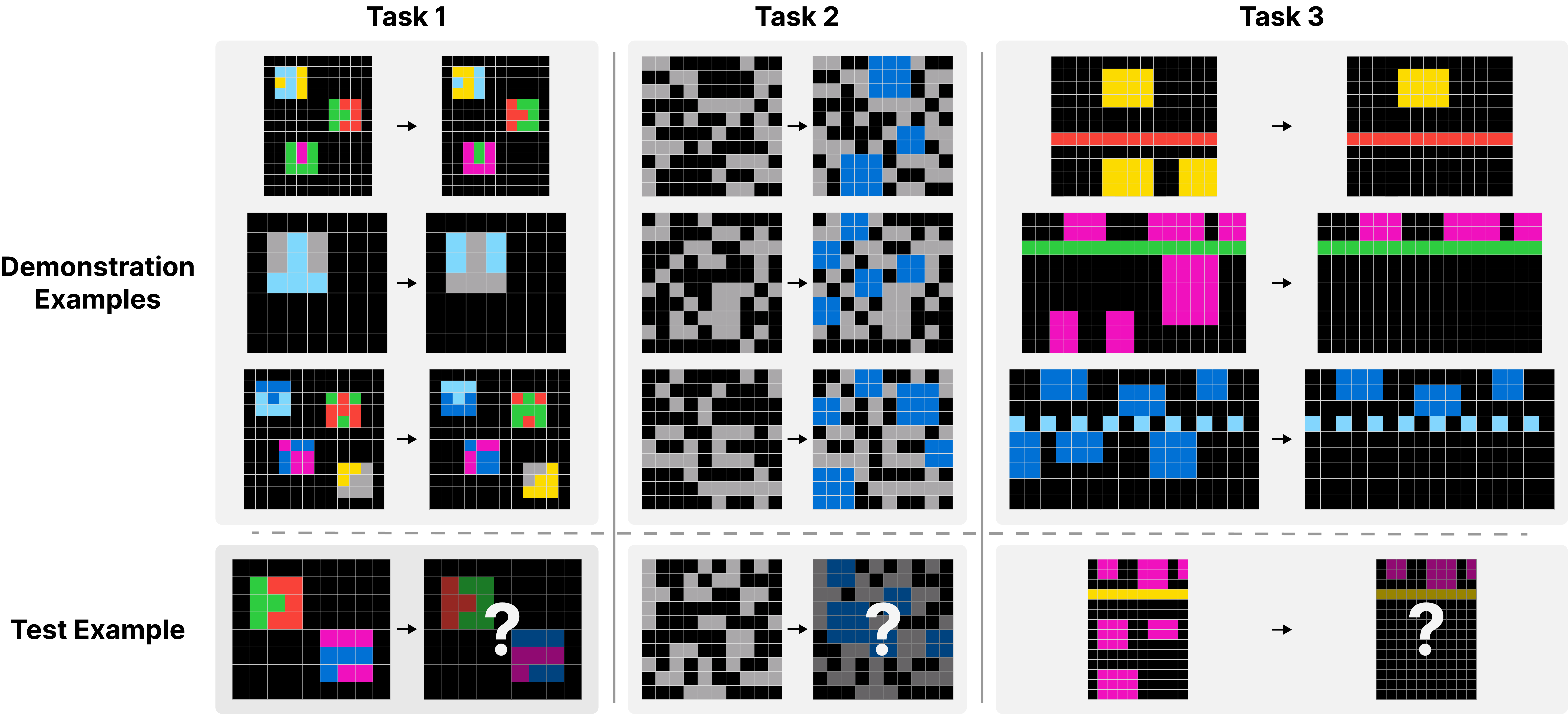}
    \caption{Three different ARC tasks. Each task involves demonstration examples of input and output grids that exemplify the required transformation. Solvers must generate the correct output grid for the test example's input grid by applying the same transformation. ARC is a straightforward benchmark that can be solved using only four types of prior knowledge: objectness, goal-directedness, arithmetic, and geometric topology. Despite the small amount of prior knowledge required to solve the tasks, it presents a high level of reasoning difficulty. These characteristics enable ARC to serve as a benchmark that fairly measures reasoning abilities.}
    \label{fig:intro/arc_examples}
\end{figure}

The ARC remains an unsolved challenge despite its seemingly simple content and evaluation methods. It demands a high level of abstraction and multiple reasoning steps, which explains why conventional deep learning techniques have not achieved success. The best-performing models to date have only achieved an accuracy of 40-55\%~\cite{lab422024arcprize}, while LLMs (GPT-4, PaLM) have shown an accuracy of around 10-20\%~\cite{mirchandani2023large}. Compared to the average human accuracy of 80\%~\cite{johnson2021fast}, these results suggest significant differences in reasoning and abstraction capabilities between humans and LLMs. However, in-depth research into how LLMs reason and how their reasoning differs from human reasoning is lacking. This gap has led to calls for a shift from a results-focused evaluation to a more nuanced analysis of the process~\cite{chang2024llmsurvey, huang2023towards, xue2023phy, zhu2023physicsofLLM-3-2}, indicating a need for a new perspective that evaluates reasoning abilities based on the process rather than just the outcome.

To overcome the limitations of result-oriented analysis in artificial intelligence, this study adopts an existing theory on what constitutes human reasoning ability. According to the Language of Thought Hypothesis (LoTH)~\cite{fodor1975language}, human reasoning encompasses three essential characteristics: \textbf{Logical Coherence}, the ability to maintain consistency in reasoning; \textbf{Compositionality}, the capability to construct complex ideas from simpler components; and \textbf{Productivity}, the capacity to formulate an indefinite number of thoughts or solutions using a finite set of elements.

\begin{figure}[ht]
    \centering
    \includegraphics[width=\columnwidth]
    {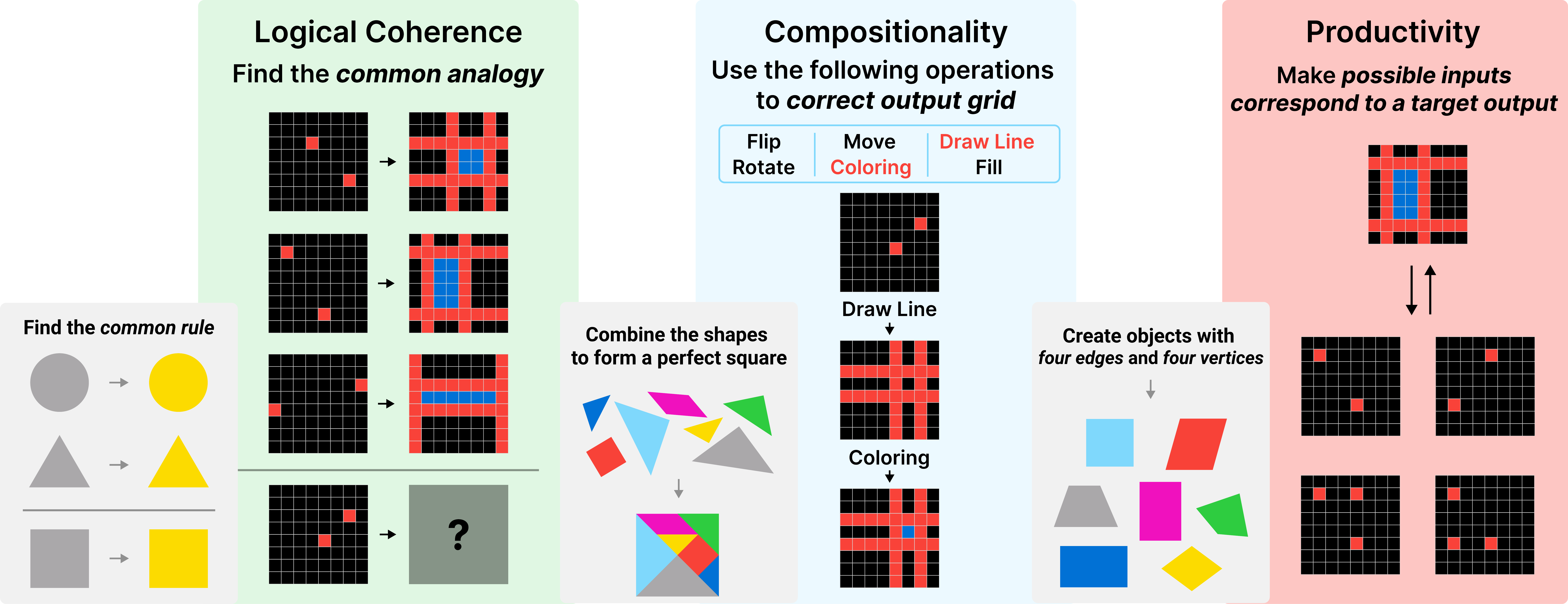}
    \caption{Three concepts of the Language of Thought Hypothesis (LoTH).}
    \label{fig:three_concepts}
\end{figure}

While attempts to evaluate logical coherence, compositionality, and productivity have existed before~\cite{bubeck2023sparks, valmeekam2024planbench}, there were limitations in that the definitions of each component varied across papers and existing benchmarks showed insufficient performance in assessing each aspect. This study differs from previous research in two key ways: 1) by redefining concepts borrowed from psychology to fit into the field of computer science, and 2) by evaluating all elements through the visual reasoning benchmark ARC. To achieve these goals, we have designed three experiments:

\begin{enumerate}
\item \textbf{Logical Coherence:} LoTH identifies two types of coherence. These are Inferential Coherence -- the ability to apply logical reasoning across related instances coherently -- and Semantic Coherence --- the ability to maintain logical coherence in the reasoning process and results~\cite{fodor1988connectionism}. To verify both types of logical coherence, we augmented each solved ARC task with 100 similar test examples and evaluated the LLM's performance on these related instances. Additionally, we analyzed the solution processes, identifying cases where correct answers were derived from flawed reasoning, to measure the LLM's semantic coherence.

\item \textbf{Compositionality:} 
Compositionality refers to a system's capacity to express one proposition being inherently linked to its ability to express related propositions~\cite{fodor1988connectionism}. In this study, we define compositionality as the ability to combine given semantics. Therefore, to evaluate compositionality, it is necessary to verify whether semantics can be combined as desired. Consequently, this study provided LLMs with step-by-step functions and examined whether they could identify the appropriate functions to solve ARC problems. Subsequently, we conducted an additional analysis to determine if the LLM could accurately predict the results from the given step-by-step functions and to understand the reasons for the failure.

\item \textbf{Productivity:} Productivity refers to the ability to infinitely create unseen expressions by combining a limited set of semantics~\cite{fodor1988connectionism}. However, it is difficult to quantitatively measure whether one can make an infinite number of unseen expressions. Therefore, previous studies have evaluated productivity by assessing whether rule-compliant unseen expressions can be created~\cite{van2004lack, lake2018generalization, hupkes2020compositionality}. Similarly, in this study, to evaluate the ability to generate unseen expressions, we examined whether unseen ARC tasks that comply with the rules could be generated when given a set of functions.
\end{enumerate}

As a result, we have confirmed that the current level of LLMs possesses a basic understanding of images and is capable of simple types of compositional object manipulations. However, compared to human reasoning abilities, LLMs lag in three areas: 1) They are not inferentially and semantically coherent; 2) Their logical reasoning abilities, especially in a step-by-step manner, are weak; 3) They struggle with understanding and generating unseen representations under complex constraints.

Finally, this study summarizes and presents recent trends proposed to address the weaknesses in abstraction abilities and reasoning capabilities. Analyzing the reasoning abilities of LLMs according to the components of human reasoning and discussing how to enhance each component represents a differentiated approach from previous research. It offers a fresh perspective for measuring and advancing the reasoning capabilities of LLMs in the future.

\section{Preliminaries}
\label{preliminaries}

This section aims to explain why we chose the LoTH perspective and ARC before starting a detailed evaluation of LLMs' reasoning capabilities. First, we will look at existing definitions of reasoning abilities and show why LoTH is useful in the perspective of measuring intelligence in Section~\ref{sec:LOTH}. Then, in Section~\ref{sec:ARC as reasoning benchmark}, we show that the ARC is an appropriate benchmark for studying LLMs from the perspective of human reasoning, as it 1) utilizes abstract semantics that can be generalized, and 2) is easy to modify.

\subsection{Limitation on Assessing Reasoning Ability of LLMs}
\label{sec:LOTH}
Efforts to evaluate LLMs' capabilities continue, highlighting their strengths in image and text generation. Especially, analyses confirm that LLMs possess elements of a World Model~\cite{gurnee2023language}, indicating potential in inference tasks. However, challenges in reasoning persist~\cite{valmeekam2024planbench}, with errors such as distortion and incomplete reasoning being frequently observed~\cite{li2024understanding}. Studies indicate that complex compositionality remains a significant challenge~\cite{Nouha2023faith}.

The divergent claims about LLMs' reasoning abilities stem from result-centric measurement methods. Turing first shifted the approach toward a consequential direction~\cite{turing1950computing}, followed by others who focused on performance measurement~\cite{wiener1950cybernetics, mcculloch1943logical, rosenblatt1958perceptron}. Recently, Chollet attempted to quantify inference abilities from a consequential perspective~\cite{chollet2019ARC}. However, these studies focus on what reasoning can achieve without specifying its constituent elements. West et al.~\cite{west2024generative} raised concerns about evaluating LLMs' reasoning abilities solely from this perspective.

To address these limitations, we propose adopting LoTH perspective. LoTH enhances discussions by integrating reasoning components with quantitative metrics, positing that inference involves the manipulation of mental representations with compositional syntax and combinatorial semantics. Our study evaluates LLMs' inference capabilities through LoTH, focusing on logical coherence, compositionality, and productivity.

Previous research has evaluated these aspects independently. Logical coherence refers to the ability to construct coherent logic in problem-solving~\cite{zhang2024ratt}. Compositionality involves understanding and combining complex expressions~\cite{lake2018generalization}. Productivity is assessed by the accuracy and efficiency of output generated from limited resources~\cite{van2004lack, hupkes2020compositionality}. However, these attempts lack unified criteria and fail to provide a direct comparison to human reasoning processes.

Adopting the LoTH perspective offers strong justification for improving reasoning capabilities. It helps develop the ability to process information and solve tasks in a manner similar to human reasoning. Logical coherence ensures reasoning without contradictions, compositionality allows the adaption of known knowledge to new scenarios, and productivity enhances the capacity to generate results based on given rules. Thus, this approach aids LLMs in achieving more human-like reasoning, enabling them to address complex problems with innovative and valid results.

\subsection{Advantages of using ARC as Reasoning Benchmark}
\label{sec:ARC as reasoning benchmark}

The Abstraction and Reasoning Corpus (ARC) emerges as a compelling candidate for evaluating inference abilities from the perspective of LoTH. ARC aligns with LoTH by requiring the combinations of semantics to solve problems and allowing flexible task modifications.

\subsubsection{Core Properties of ARC}
ARC's key characteristic is its requirement to extract and combine compositional semantics, necessitating sophisticated problem-solving approaches. Two research findings support this:

\begin{enumerate}
    \item \textbf{Importance of Semantic Information:} Studies show that supplementary semantic information significantly improves ARC task performance. For instance, integrating graph-represented object information nearly doubled the success rate~\cite{xu2023llms}.
    \item \textbf{High Abstraction Level of ARC:} ARC's abstraction level surpasses that of other benchmarks~\cite{malkinski2023review}. Chollet argues that conventional feature extraction methods are insufficient for ARC, given its demand for complex shape interpretation and transformation comprehension~\cite{chollet2019ARC}.
\end{enumerate}

These observations highlight the need to develop approaches that can effectively extract and utilize complex, abstract information for solving ARC tasks. Such properties fit with the perspective of LoTH, which views reasoning ability as arising from a combination of semantics.

\subsubsection{Flexibility in benchmark adaptation}

Despite its simple rules, ARC remains challenging, with LLMs achieving 15\% accuracy~\cite{qiu2023phenomenal}, traditional program synthesis models reaching 26\%~\cite{Icecuber2020dsl}, and the human average accuracy at 80\%~\cite{johnson2021fast}. Various ARC variants have emerged to address this challenge:

\begin{enumerate}
    \item \textbf{1D-ARC}~\cite{xu2023llms}: Reduces dimensionality from 2D to 1D, simplifying complexity while retaining core knowledge. It effectively addresses object cohesion challenges, achieving high LLM accuracy (approximately 90\%).
    
    \item \textbf{MC-LARC}~\cite{Shin2023mclarc}: Adopts a multiple-choice format, transitioning from generative tasks to selection tasks. GPT-4 demonstrated strong performance (approximately 75\%).
    
    \item \textbf{Mini-ARC}~\cite{Kim2022playgrounds}: Limits grid size to 5x5, simplifying input while retaining 2D generative characteristics. Performance remains challenging, similar to the original ARC (approximately 15\%).

    \item \textbf{ConceptARC}~\cite{arseny2023ConceptARC}: Organizes tasks into concept groups that focus on specific spatial and semantic concepts. Performance remains challenging, similar to the original ARC (approximately 20\%).
\end{enumerate}

\revise{These variations demonstrate ARC's transformation flexibility and emphasize the necessity of composition in solving ARC tasks. MC-LARC and 1D-ARC reduced reasoning step complexity, while Mini-ARC focused on reducing image complexity. The performance differences among these variants imply that reducing the need for complex transformation combinations can significantly improve results, highlighting the importance of combinatorial syntax in solving ARC.}

\revise{In summary, the ARC emerges as a compelling benchmark for evaluating inference abilities through the lens of the LoTH. ARC's core strength lies in its requirement to extract and combine compositional semantics to solve tasks, as evidenced by improved performance with additional semantic information. The various ARC variants demonstrate flexibility for different experimental purposes, with their performance differences highlighting the necessity of combinatorial syntax in solving ARC tasks. Furthermore, ARC's high level of abstraction and reasoning complexity, shown by the significant gap between human and AI performance, validates its use as an effective tool for exploring inference abilities in the context of the LoTH.}

\section{Evaluating the Inferential Capabilities of LLMs Using the ARC Benchmark}
\label{sec:experiments}

To evaluate whether LMs possess inferential capabilities, one could compare these capabilities to human reasoning. As explained in Section~\ref{sec:LOTH}, according to LoTH, human reasoning can be broadly categorized into three main components: Logical Coherence (Section~\ref{sec:logical_coherence}), Compositionality (Section~\ref{sec:compositionality}), and Productivity (Section~\ref{sec:productivity}). We utilized ARC to examine each aspect of the LLMs' reasoning capabilities from the perspective of LoTH.

\subsection{Capability of LLMs 1: Logical Coherence}
\label{sec:logical_coherence}

\subsubsection{Motivation}

Section~\ref{sec:logical_coherence} aims to evaluate the logical coherence of LLMs. This is a fundamental aspect of LoTH, which considers coherence in two dimensions: inferential coherence and semantic coherence~\cite{fodor1988connectionism}. \textbf{Semantic Coherence} refers to the ability to maintain logical coherence in the process and results of reasoning. \textbf{Inferential Coherence}, on the other hand, is a system's ability to consistently apply a specific type of logical inference across all relevant instances, given it can perform that inference in some cases. These concepts are crucial in human cognitive processes and relevant to the rule inference required in ARC tasks.

Our initial experiments primarily focused on measuring semantic coherence by evaluating whether the results produced by LLMs logically followed their problem-solving steps. This evaluation was conducted using various prompt techniques such as Chain of Thought (CoT)~\cite{wei2022chain}, Least to Most (LtM)~\cite{zhou2023least}, and Tree of Thought (ToT)~\cite{yao2023tree}, similar to previous ARC solving attempts~\cite{xu2023llms, mirchandani2023large}. We compared the coherence levels these different prompting strategies achieved, aiming to identify which techniques yielded the most semantically coherent results across diverse problem-solving scenarios. However, recognizing the limitations of this approach in addressing inferential coherence, we introduced supplementary experiments using augmented ARC tasks. These tasks, created through the Re-ARC program~\cite{hodel2024addressing}, allowed us to assess how consistently LLMs can apply logical patterns across variations of originally solved problems, providing a more comprehensive evaluation of their logical reasoning capabilities.

\begin{figure}[h!]
    \centering
    \begin{subfigure}[b]{0.32\textwidth}
        \includegraphics[width=\textwidth]{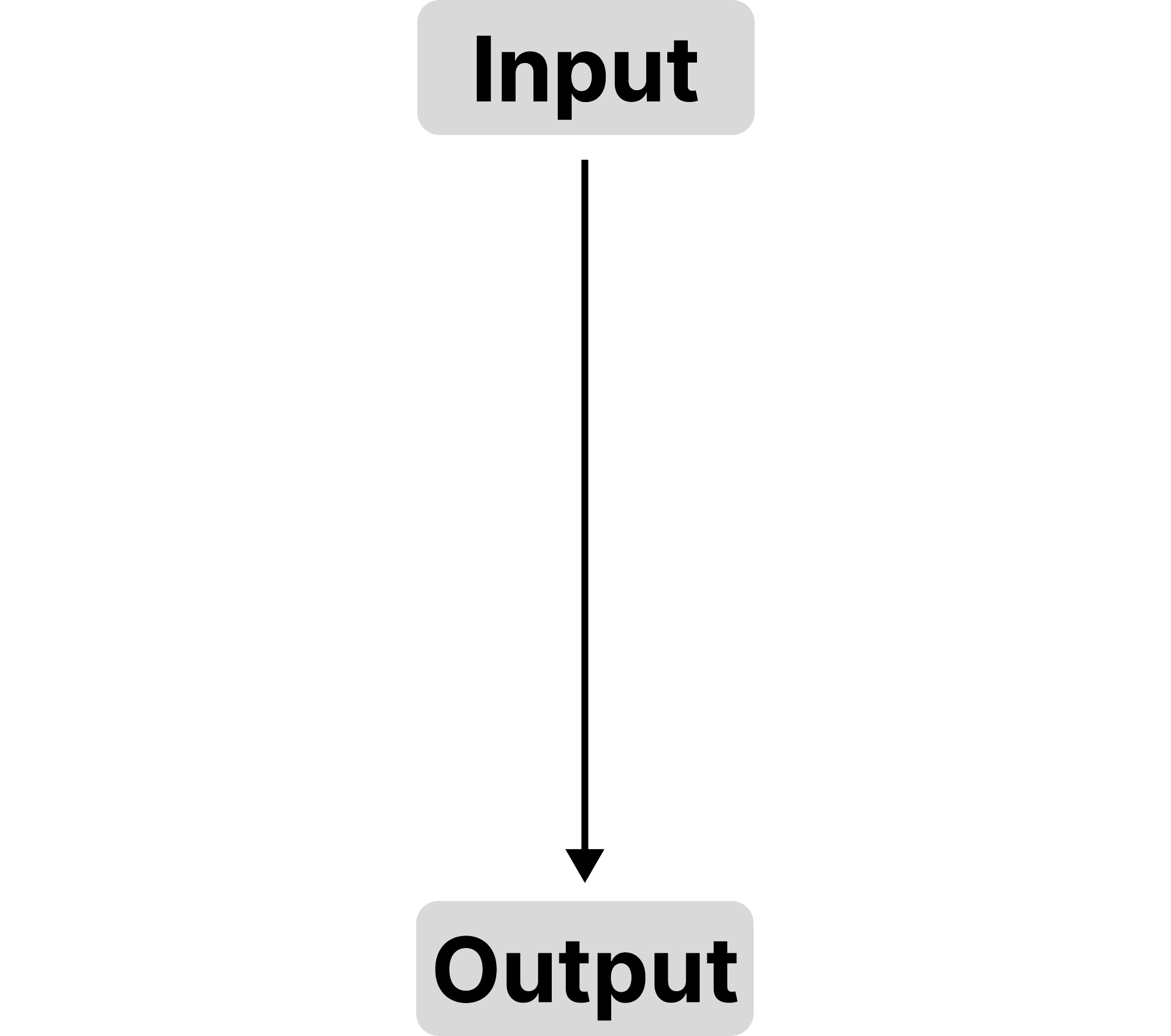}
        \caption{Chain of Thought (CoT)}
        \label{fig:logical_coherence/cot}
    \end{subfigure}
    \hfill
    \begin{subfigure}[b]{0.32\textwidth}
        \includegraphics[width=\textwidth]{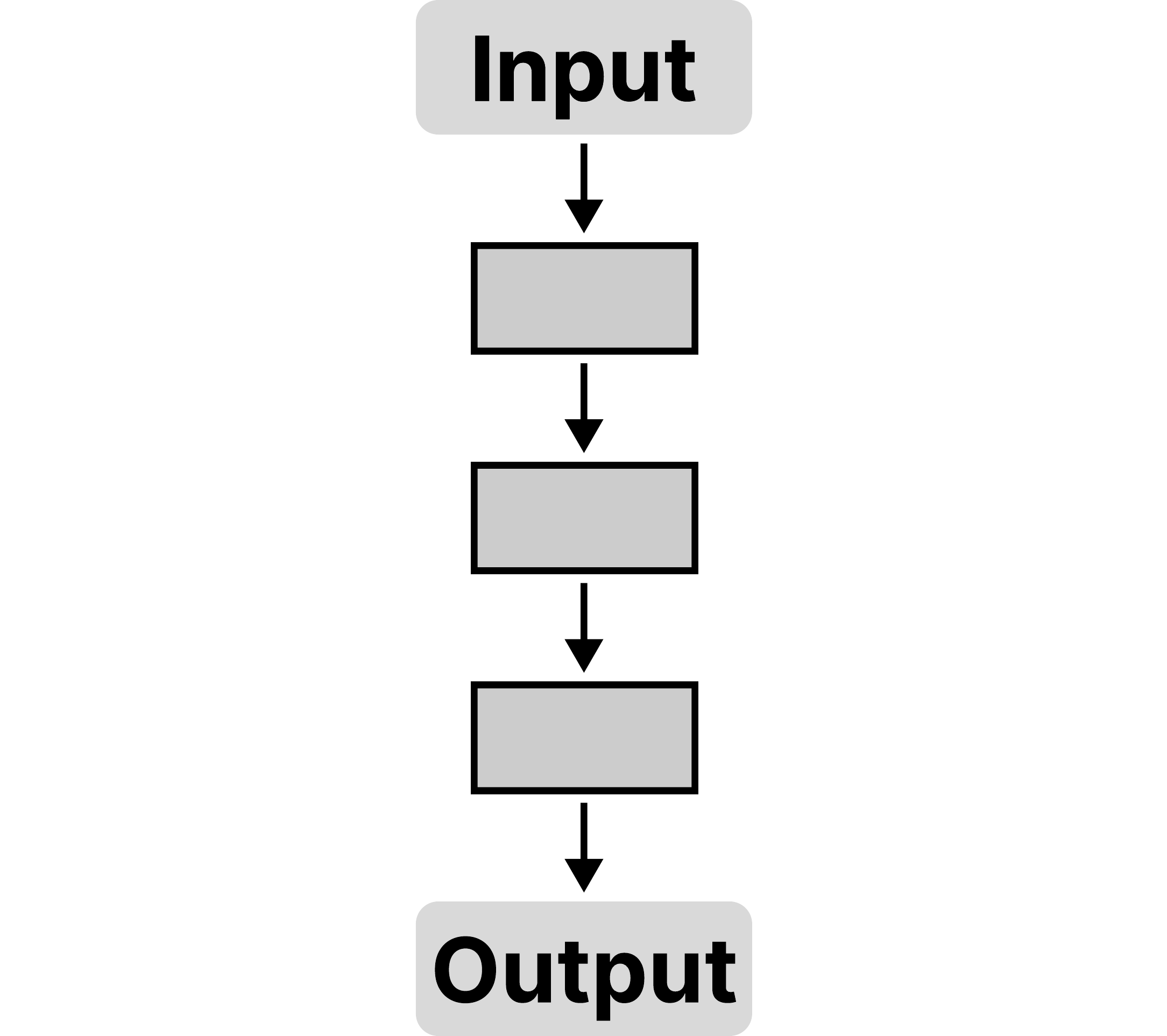}
        \caption{Least to Most (LtM)}
        \label{fig:logical_coherence/ltm}
    \end{subfigure}
    \hfill
    \begin{subfigure}[b]{0.32\textwidth}
        \includegraphics[width=\textwidth]{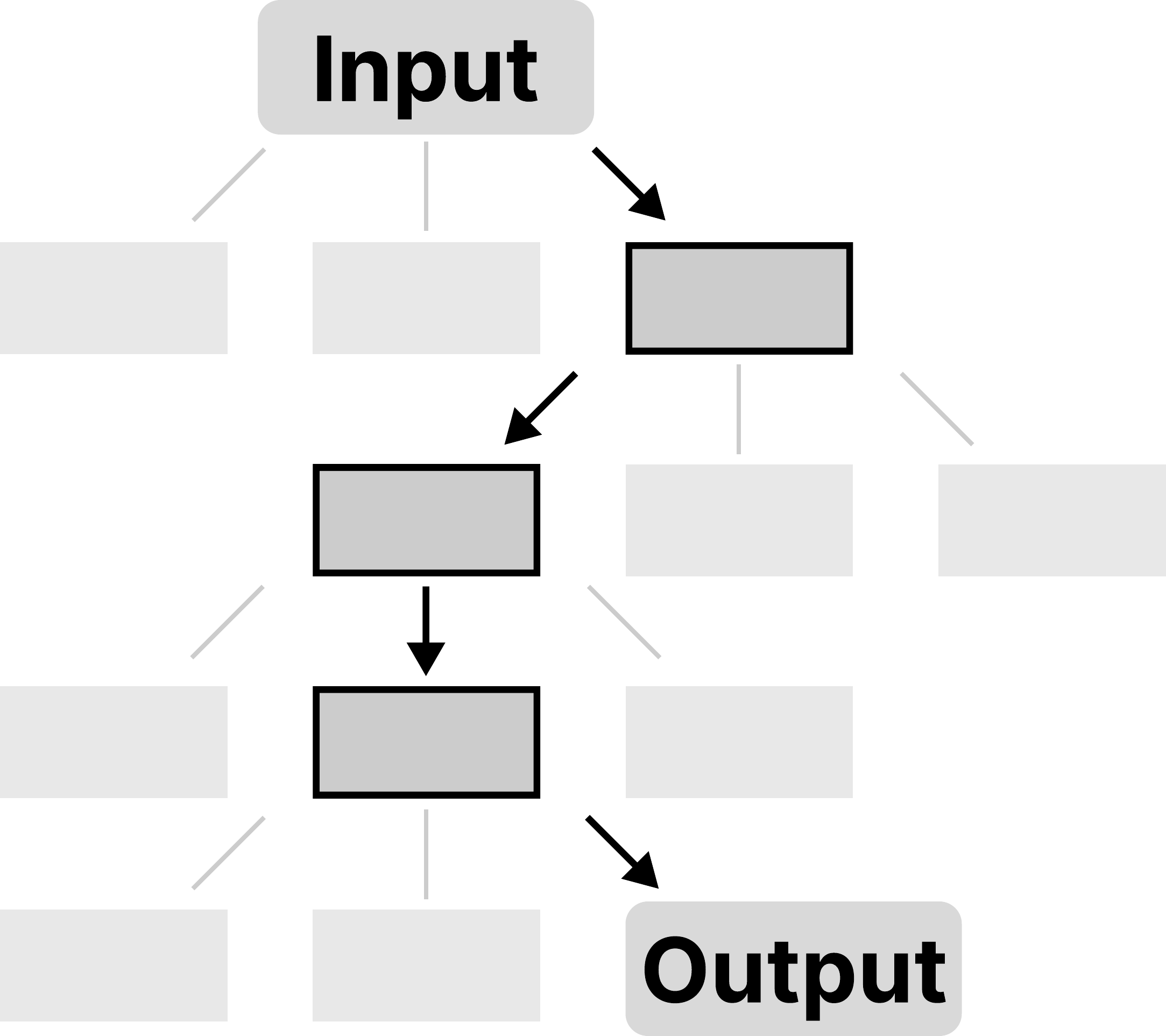}
        \caption{Tree of Thoughts (ToT)}
        \label{fig:logical_coherence/tot}
    \end{subfigure}
    \caption{Three prompting techniques in the experiment about logical coherence: (a) CoT, (b) LtM, and (c) ToT.}
    \label{fig:logical_coherence/three_prompting_techniques}
\end{figure}

\subsubsection{Comparison Across Prompting Techniques}
\label{sec:logical_coherence/prompt_techniques}

The perceived deficiency in LLMs' logical reasoning has been a recurrent critique, with direct attempts to solve ARC tasks yielding success rates below 10\%~\cite{mirchandani2023large}. To address this issue, enhancements in LLMs' logical reasoning are being pursued through prompting techniques such as CoT, LtM, and oT. 
These strategies have been shown to effectively leverage LLMs' reasoning capabilities~\cite{wang2019multi} and offer the advantage of providing a more transparent analysis for humans, as they involve a step-by-step reasoning process. Therefore, in this experiment, we assess the impact of these prompting strategies on LLMs' logical coherence by solving ARC tasks.

We applied three major prompting techniques -- CoT, LtM, and ToT -- to solve 100 ARC evaluation tasks using the GPT-4-32k model. Each technique was tested across five iterations. ARC tasks follow a few-shot learning paradigm, requiring the model to infer task rules from given example pairs and apply them to test examples. The CoT method enhances reasoning performance by generating answers through a structured chain of thought, which systematically connects steps required for solving ARC tasks and provides examples in the prompt. Similar contextual information was provided for LtM and ToT. LtM decomposes tasks into manageable steps and executes them sequentially, while ToT generates multiple candidates at each step post-decomposition, selecting the best candidate through a voting mechanism before proceeding to the next step.

\begin{figure}[ht]
    \includegraphics[width=0.89\textwidth]{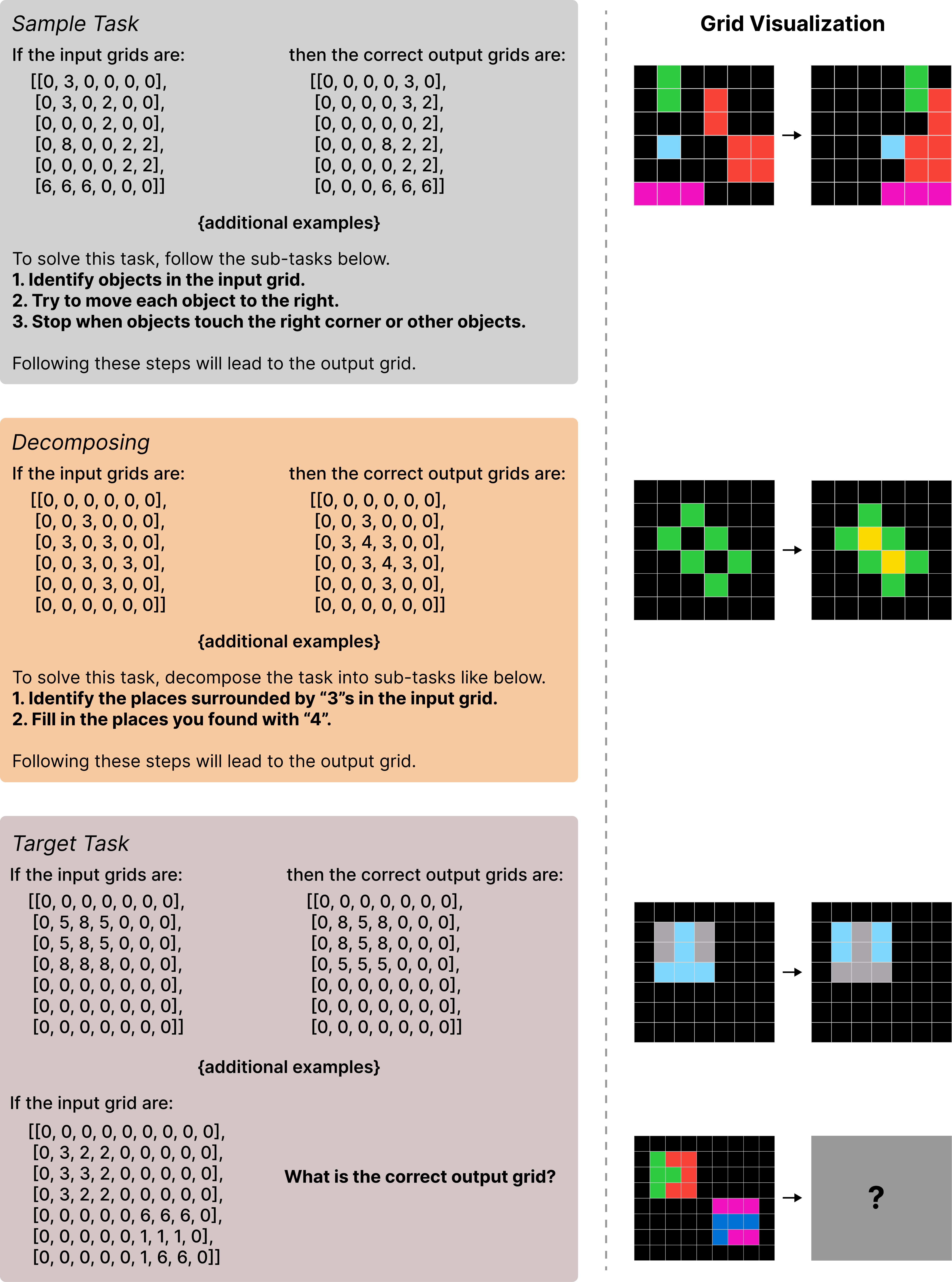}
    \captionsetup{width=\linewidth}
    \caption{Three types of prompts are shown on the left. Although all prompts are described as a 2D array of grids, we visualized them on the right for clarity. By default, all three techniques use prompts with two main components: a sample task and a target task. However, LtM and ToT use a different combination of the target task and its decomposition command. This difference arises because CoT strictly follows the given sub-task, while LtM and ToT decompose the task on their own.}
    \label{fig:logical_coherence/Components_of_Prompting}
\end{figure}

\begin{figure}[ht]
    \includegraphics[width=0.89\textwidth]{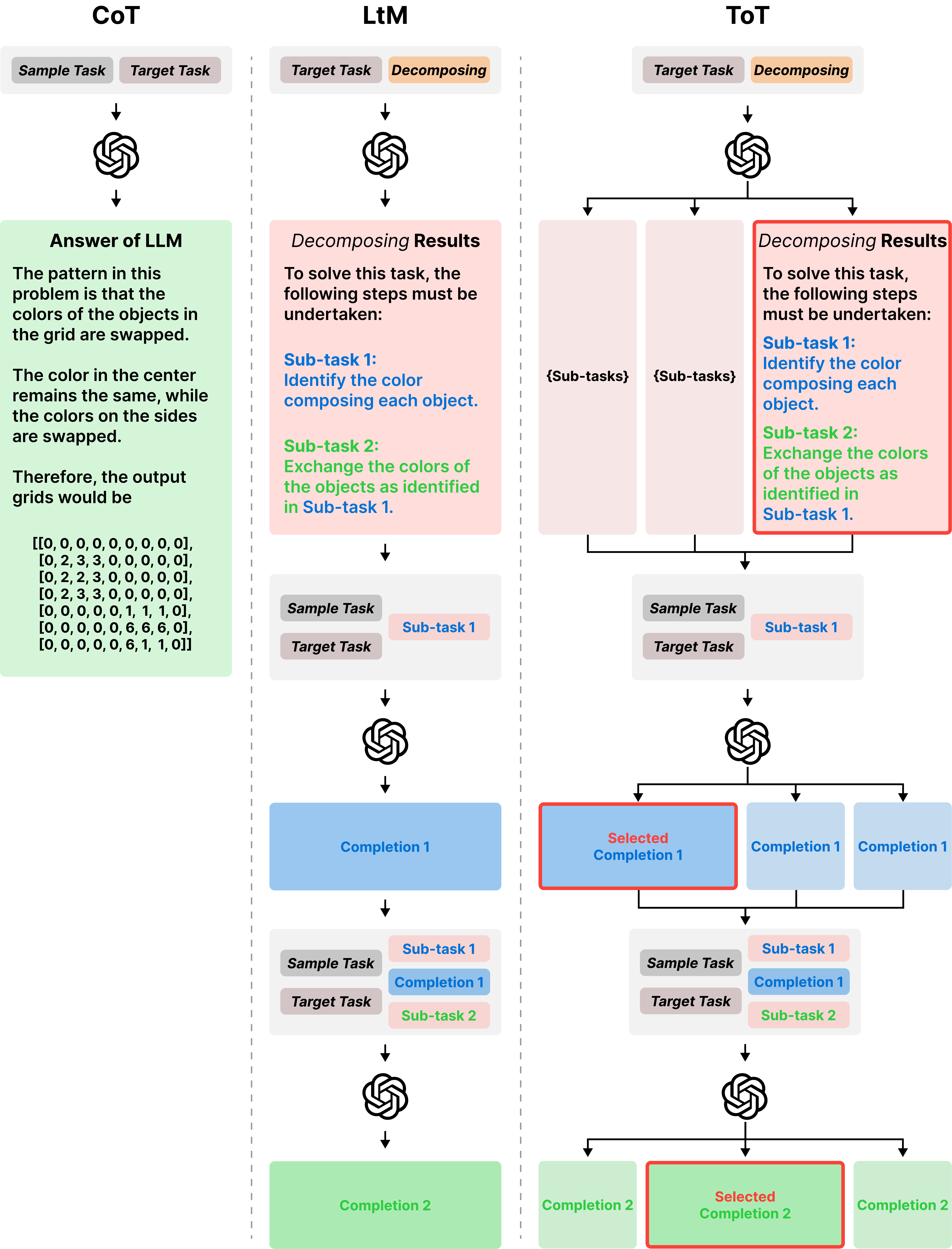}
    \captionsetup{width=\linewidth}
    \caption{Grey blocks illustrate prompt sets delivered to the LLM, including the sample task, target task, and LLM's prior responses, as shown in Fig.~\ref{fig:logical_coherence/Components_of_Prompting}. Green blocks denote the final answer. CoT relies on a single grey block, indicating that the LLM strictly follows the provided sub-tasks. Conversely, LtM and ToT prompt the LLM to generate and address sub-tasks sequentially, represented by decomposed results (red) and intermediate responses (blue). ToT further distinguishes itself from LtM by evaluating multiple suggestions for sub-task handling and selecting the most effective one through a voting mechanism.}
    \label{fig:logical_coherence/allinone}
\end{figure}

\begin{table}[ht]
\small
\centering
\caption{Averaged performance of each prompting technique. The accuracy is based on solving 100 random ARC tasks with CoT, LtM, and ToT prompts, each repeated five times. The accuracy outside the parentheses refers to the accuracy when only the results are correct, while the accuracy inside the parentheses indicates the accuracy when both the results and the process are correct.}
\begin{tabular}{c c c c c} 
\toprule
\textbf{Iteration} &
\textbf{CoT} &
\textbf{LtM} &
\textbf{ToT} \\
   \midrule
1 & 11\% (3\%) & 6\% (4\%) & 7\% (3\%) \\
2 & 10\% (2\%) & 7\% (4\%)& 4\% (1\%) \\
3 & 10\% (5\%) & 6\% (3\%) & 7\% (2\%) \\
4 & 10\% (4\%) & 4\% (2\%) & 7\% (4\%) \\
5 & 10\% (6\%) & 5\% (2\%) &  6\% (2\%) \\ \midrule
Average & 10.2\% (4.0\%) & 5.6\% (3.0\%) &  6.2\% (2.4\%) \\
\bottomrule
\end{tabular}
\label{tbl:logical_coherence/prompt_techniques_performance}
\end{table}

Comparing ARC accuracy across prompts, CoT outperformed LtM and ToT in accuracy. Table~\ref{tbl:logical_coherence/prompt_techniques_performance} presents the results of applying LtM, CoT, and ToT to 100 randomly selected tasks from the ARC evaluation set. The experiment was repeated five times, with the percentage of correct answers included for each iteration. CoT achieved approximately 10\% accuracy, while LtM and ToT showed about 6\% accuracy. CoT demonstrates superior performance, while ToT and LtM suffer from cumulative error propagation, where small mistakes in one step of their multi-step answer generation process can lead to compounded errors in subsequent steps. Given CoT's accuracy (\textasciitilde 11\%) compared to LtM and ToT (\textasciitilde 7\%) and its resilience to error propagation, we exclusively used the CoT prompt in subsequent experiments.

However, when we checked the correctness of the solution process, all three prompting techniques showed low accuracy, with no significant difference at around 3\%, as indicated in parentheses. These results demonstrate that while accuracy may differ depending on the prompting technique, there is little variation in semantic coherence. This consistency across prompting methods suggests that the issue lies not in the method of eliciting responses but in the fundamental reasoning capabilities of LLMs. It is also important to note that both the results and processes fall far short of the average human accuracy of 80\%. These low performance metrics, particularly when compared to human benchmarks, cannot be attributed to the limitations of specific prompting techniques. These findings suggest that LLMs lag behind humans in terms of logical coherence. To analyze the specific reasons for this, we conducted follow-up experiments. Section~\ref{sec:logical_coherence/inferential_coherence} analyzes inferential coherence, one aspect of logical coherence, while Section~\ref{sec:logical_coherence/semantic_coherence} examines the semantic coherence of LLMs through case studies.

\newpage

\subsubsection{Inferential Coherence of LLMs}
\label{sec:logical_coherence/inferential_coherence}

% In our second experiment, we designed a test to evaluate the inferential coherence of LLMs. Inferential coherence assesses whether an LLM can maintain the same type of logical inference across relevant tasks. To assess this, we examined whether the LLM could solve problems similar to the ARC tasks it had previously solved successfully.

In our second experiment, we tested the inferential coherence of LLMs, which measures their ability to maintain the same logical inference across tasks sharing an underlying analogical rule. To assess this, we examined whether the LLM could solve new problems defined by the same rules as previously solved ARC tasks.

\begin{figure}[ht]
    \includegraphics[width=0.8\textwidth]{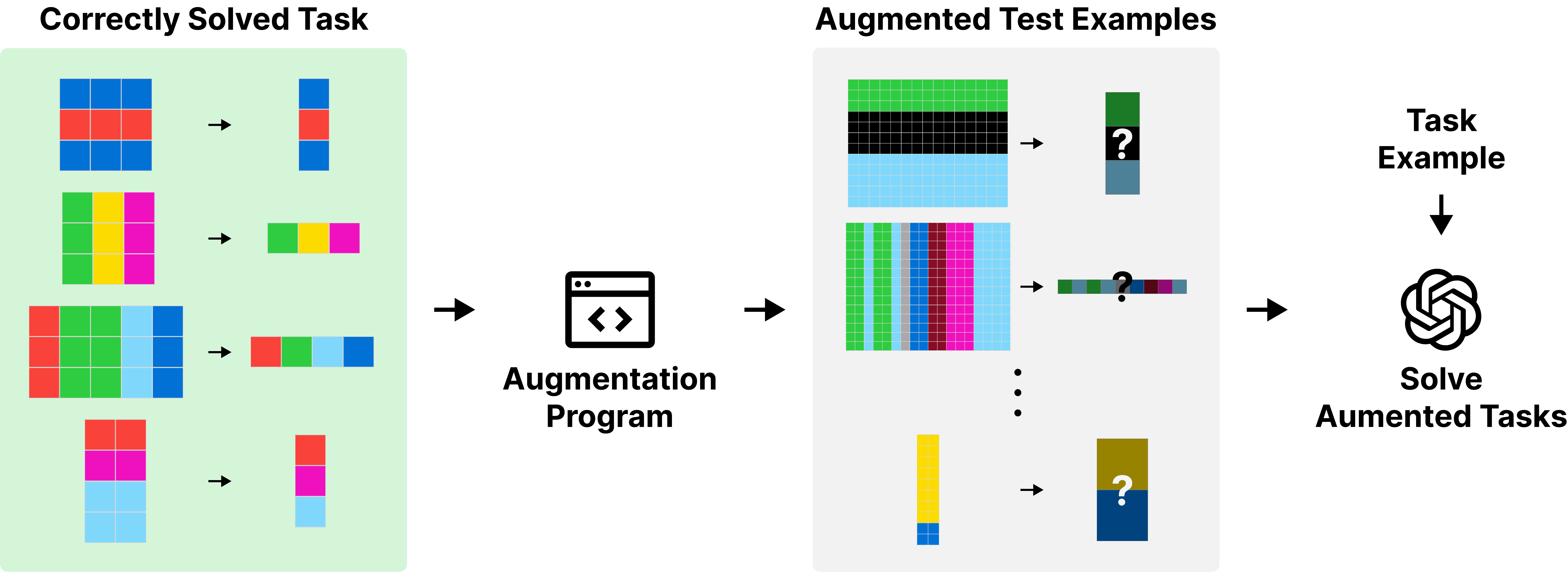}
    \captionsetup{width=\linewidth}
    \caption{\revise{Inferential coherence testing via augmentation: Using the Re-ARC augmentation program~\cite{hodel2024addressing}, we generated 100 new test examples per task. These examples retained the original analogical rule, allowing us to evaluate LLMs’ inferential coherence across varied instances.}}
    \label{fig:logical_coherence/logical_coherence_experiment}
\end{figure}

\begin{algorithm}[htbp]
\begin{flushleft}
\resizebox{0.9\textwidth}{!}{ % 알고리즘 전체 크기를 90%로 줄이기
\begin{varwidth}{\textwidth} % varwidth 사용하여 오류 방지
\DontPrintSemicolon
\SetNoFillComment
\caption{Calculating Inferential Coherence}
\label{alg:training-evaluation}
\textbf{Input:} Training Tasks $\mathcal{T}$, LLM Model $\mathrm{L}()$, Re-ARC Program $R()$, CoT Prompt $\mathrm{P}()$ \\
\smallskip
\textbf{Output:} Inferential Coherence of LLM $\mathcal{C}$\\

{\small \tcc{Step 1: Identify Solved Tasks $\mathcal{T}^S$}}
$\mathcal{T}^S \gets \emptyset$ \small \tcp*{$\mathcal{T}^S$ is a set of solved tasks by LLM}
\For{$T_i \in \mathcal{T}$}{
    \For{$j \gets 1$ \KwTo $5$}{
        \If{$L(P(T_i)) == True$}{
            $\mathcal{T}^S \gets \mathcal{T}^S \cup \{T_i\}$ {\small \tcp*{Consider $T_i$ is solved if LLM solves in 5 tries}} 
            \textbf{break}
        }
    }
}
\medskip
{\small \tcc{Step 2: Augment Solved Tasks $\mathcal{T}^A$}}
$\mathcal{T}^A \gets \emptyset$ \small \tcp*{$\mathcal{T}^A$ is a set of $\mathcal{T}^A_i$ generated below}
\For{$T^S_i \in \mathcal{T}^S$}{
    $\mathcal{T}^A_i \gets \emptyset$ \small \tcp*{$\mathcal{T}^A_i$ is a set of augmented tasks from $T^S_i$} 
    \For{$j \gets 1$ \KwTo $100$}{
        $\mathcal{T}^A_i \gets \mathcal{T}^A_i \cup \{R(T^S_i)\}$ {\small \tcp*{Augment $T^S_i$ using Re-ARC}}
    }
    $\mathcal{T}^A \gets \mathcal{T}^A \cup \mathcal{T}^A_i$
}
\medskip
{\small \tcc{Step 3: Calculate Inferential Coherence $\mathcal{C}$}}
$\mathcal{C} \gets [c_1, c_2, \dots, c_{|\mathcal{T}^A|}]$    \small \tcp*{$\mathcal{C}$ is a list of the number of solved tasks in $\mathcal{T}^A_i$}
\For{$\mathcal{T}^A_i \in \mathcal{T}^A$}{
    $c_i \gets 0$ \\
    \For{$T_{i,j}^A \in \mathcal{T}^A_i$}{ 
% \For{$j \gets 1$ \KwTo $100$}{
        \If{$L(P(T_{i,j}^A)) == True$}{
            $c_i \gets c_i + 1$ \small \tcp*{Count solved $T_{i,j}^A$ in $T_i^A$}
        }
    }
    %$\text{results} \gets \text{results} \cup \{L(P(T_{i,j}^A))\}$ \small \tcp*{Evaluate inferential coherence with $T_{i,j}^A$}
}

% \medskip
% {\small \tcc{Step 4: Calculate Exact Task Distribution}}
% $\mathcal{D}^{exact} \gets [e_1, e_2, \dots, e_{100}]$ \small \tcp*{$\mathcal{D}^{exact}$ is the distribution of tasks solved exactly $c_i$ times}

% \For{$i \gets 1$ \KwTo 100}{
%     $e_i \gets 0$ \\
% }

% \For{$c_i \in \mathcal{C}$}{ 
%     $e_{c_i} \gets e_{c_i} + 1$ \small \tcp*{Count tasks solved exact $i$ times}
% }

% \medskip
% {\small \tcc{Step 5: Calculate Reverse-Accumulated Task Distribution}}
% $\mathcal{D}^{accumulated} \gets [a_1, a_2, \dots, a_{100}]$ \small \tcp*{$\mathcal{D}^{accumulated}$ is the count of tasks solved at least $i$ times}

% $a_{100} \gets e_{100}$ \\
% \For{$i \gets 99$ \KwTo $1$}{ 
%     $a_i \gets e_i + a_{i+1}$ \small \tcp*{Accumulate tasks solved at least $i$ times in reverse order}
% }

\medskip
\textbf{return} $\mathcal{C}$
% \textbf{Return} $\mathcal{D}^{accumulated}$, $\mathcal{D}^{exact}$

\end{varwidth} % varwidth 사용하여 너비 지정
}
\end{flushleft}

\end{algorithm}

\revise{Fig.~\ref{fig:logical_coherence/logical_coherence_experiment} summarizes the experiment, and the detailed procedure is in Algorithm~\ref{alg:training-evaluation}. We began by using GPT-4o to solve examples from 400 ARC tasks,\footnote{We selected 400 tasks from the ARC training set, as Re-ARC can only augment the training set} repeating this five times to identify consistently solvable tasks. For tasks solved correctly at least once, we used Re-ARC~\cite{hodel2024addressing} to generate 100 additional examples that mirror the original approach. We hypothesized that a model demonstrating inferential coherence would solve all augmented examples, allowing us to rigorously test its generalization ability across similar tasks.}

% Fig.~\ref{fig:logical_coherence/cdf_pdf} shows the results. Fig.~\ref{fig:logical_coherence/cdf} is a graph that represents all five iterations for the number of cumulatively correct augmented test examples. It's important to note that the graphs for all five iterations show an exponential decrease. This graph trend doesn't vary significantly between iterations, demonstrating low coherence regardless of the iteration. \revise{Fig.~\ref{fig:logical_coherence/pdf} shows the distribution of accuracy, averaged over five iterations, on 100 augmented test examples for each task. The notable points in this graph are that more than two-thirds of the tasks, 54 tasks, are concentrated below the average accuracy of 0.2. These figures demonstrate that LLMs have low inferential coherence for most ARC tasks.}

\begin{figure}[ht]
    \centering
    \begin{subfigure}[b]{0.4848\textwidth}
        \includegraphics[width=\textwidth]{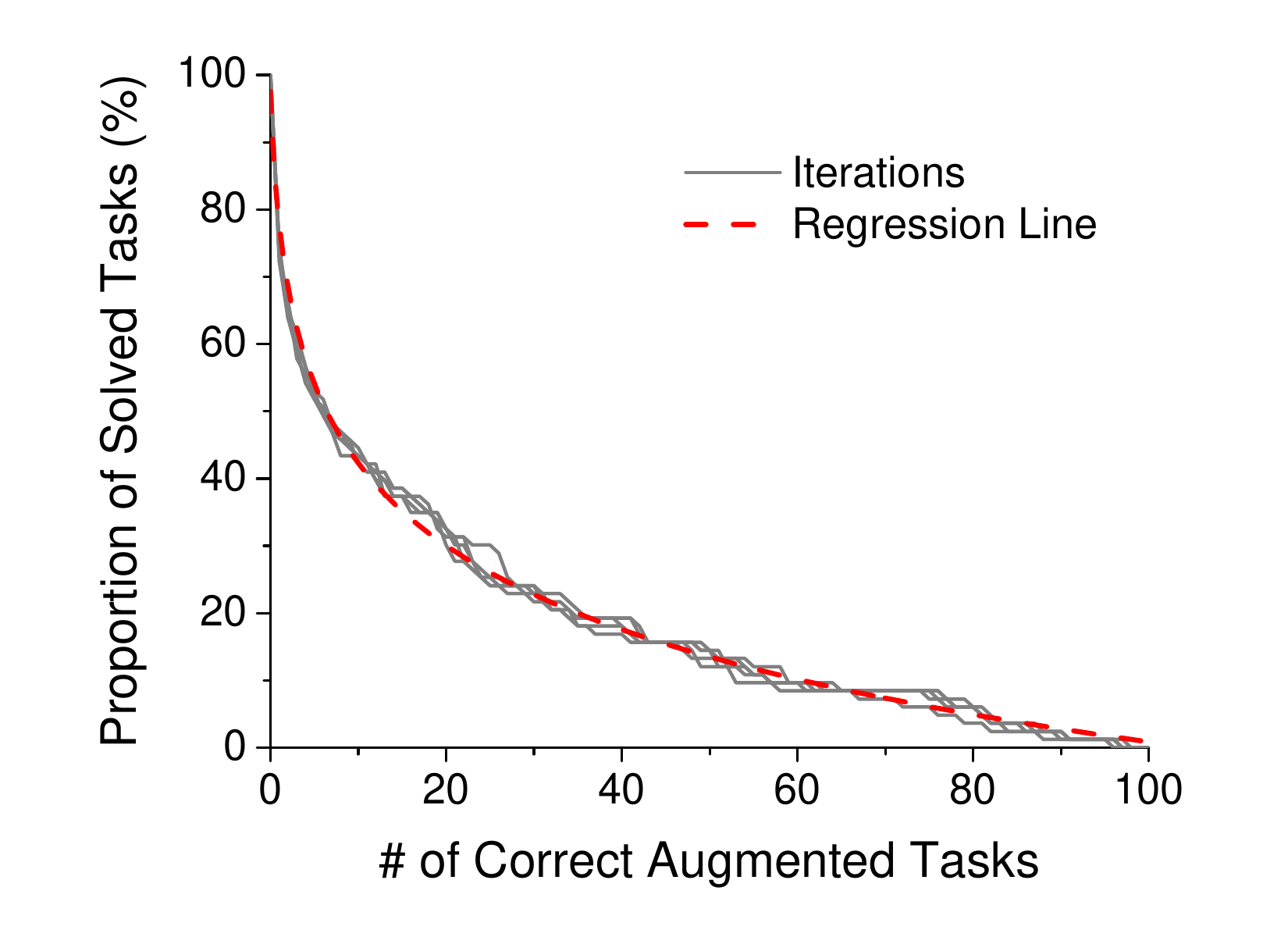}
        \caption{\revise{Cumulative distributions of successful solutions across 100 augmented examples per task. The y-axis shows the cumulative proportion of tasks achieving each success rate. The consistent exponential decay across five iterations demonstrates persistent low coherence.}}
        \label{fig:logical_coherence/cdf}
    \end{subfigure}
    \hfill
    \begin{subfigure}[b]{0.4848\textwidth}
        \includegraphics[width=\textwidth]{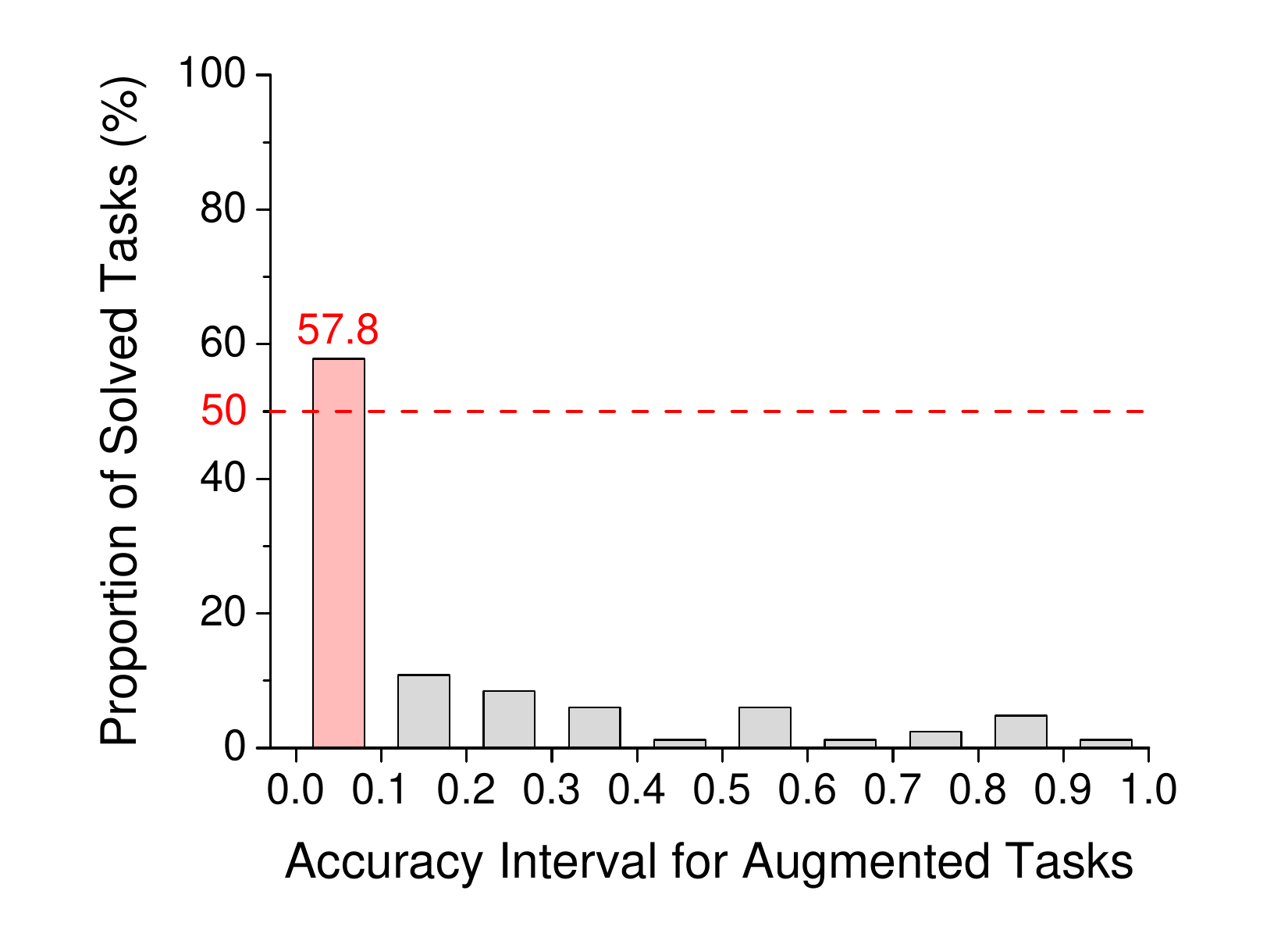}
        \caption{\revise{Distribution of ARC tasks by accuracy intervals of 0.1 on augmented examples. The y-axis shows the proportion of tasks per accuracy interval, averaged over five iterations. More than half (57.8\%) of tasks achieved below 10\% accuracy, demonstrating limited inferential coherence.}}
        \label{fig:logical_coherence/pdf}
    \end{subfigure}
    \caption{\revise{Analysis of GPT-4o's performance on augmented test examples. For each of the 83 tasks successfully solved from the ARC training set, we generated 100 additional examples per task using Re-ARC~\cite{hodel2024addressing} to evaluate inferential coherence.}}
    \label{fig:logical_coherence/cdf_pdf}
\end{figure}

\revise{Fig.~\ref{fig:logical_coherence/cdf_pdf} presents two key analyses of the results. The cumulative distribution (Fig.~\ref{fig:logical_coherence/cdf}) shows consistent exponential decay patterns across all five iterations, indicating persistent low coherence regardless of iteration. The accuracy distribution (Fig.~\ref{fig:logical_coherence/pdf}) reveals that 57.8\% of tasks achieved below 10\% accuracy on augmented examples. Together, these results demonstrate LLMs' limited inferential coherence on ARC tasks.}

\subsubsection{Case Study: Semantic Coherence of LLMs}
\label{sec:logical_coherence/semantic_coherence}

\begin{figure}[ht!]
    \includegraphics[width=0.9\textwidth]{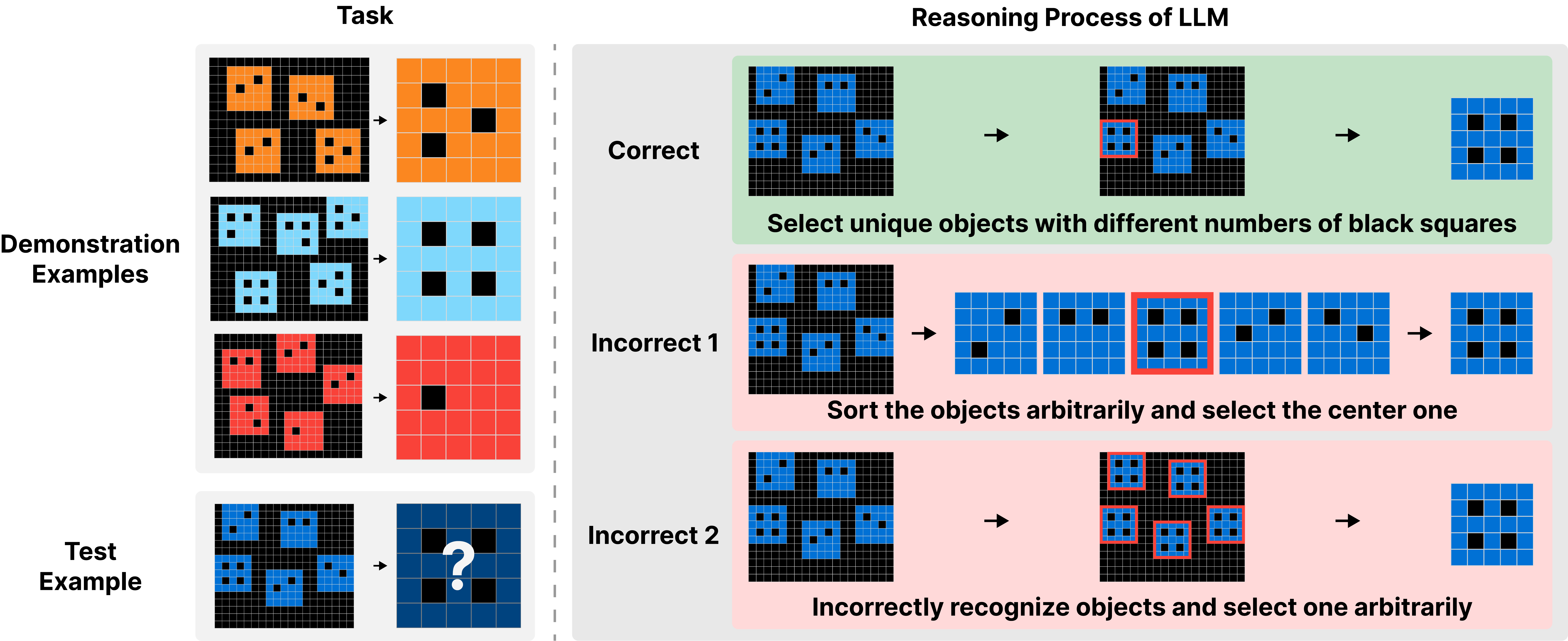}
    \captionsetup{width=\linewidth}
    \caption{Presenting instances where LLMs reach the correct answer but use flawed reasoning, highlighting the challenge of applying consistent logical rules across different ARC tasks. The task involves identifying a unique $5 \times 5$ object within a grid based on the number of black squares. The `Correct' process shows the LLM correctly identifying the unique object, while `Incorrect 1' and `Incorrect 2' represent failed reasoning—one due to arbitrary selection and the other due to misidentification.}
    \label{fig:logical_coherence/task_solving_example}
\end{figure}

Finally, we analyzed how LLMs solved tasks in the two experiments described in Section \ref{sec:logical_coherence/prompt_techniques} and Section~\ref{sec:logical_coherence/inferential_coherence}. When evaluating not only the answers but also the process for the three prompts CoT, LtM, and ToT, we found that regardless of the prompt, the accuracy was about 3\%, indicating that correct answers were being derived from incorrect processes, as shown in Fig.~\ref{fig:logical_coherence/task_solving_example}.

To solve the task, 1) identify $5 \times 5$ objects within the input grid, 2) count the number of black squares in each object, and 3) extract the object with the highest number of black squares. However, CoT, LtM, and ToT attempted to solve the task incorrectly. For CoT, objects in the input grid were sorted, and then the object in the middle was selected. Although CoT arrived at the correct answer, the method of sorting the objects lacked clarity. For LtM and ToT, there was an understanding that a specific object from the input grid needed to be selected to solve the task, but they mistakenly recognized objects from the test input grid. These solutions share a common flaw: they fail to establish a logically consistent rule across the different examples of training inputs and outputs provided. In other words, regardless of the prompting technique (CoT, LtM, or ToT), LLMs still struggle to demonstrate logical coherence in deriving a single rule that consistently applies across the examples given to solve the task.

The inconsistency of inferring correct results from incorrect processes was also observed in the second experiment conducted on the training set. Upon analyzing the natural language explanations for the 83 tasks solved at least once out of 400 training tasks, we found that in 35 of these cases, the solutions proposed by the LLM could not produce the correct answer. This finding suggests that LLMs lack semantic coherence regardless of the prompting technique or tasks. In other words, LLMs derive outcomes unrelated to their reasoning process, as evidenced by generating correct answers from incorrect solutions.

\begin{figure}[ht!]
    \includegraphics[width=0.95\textwidth]{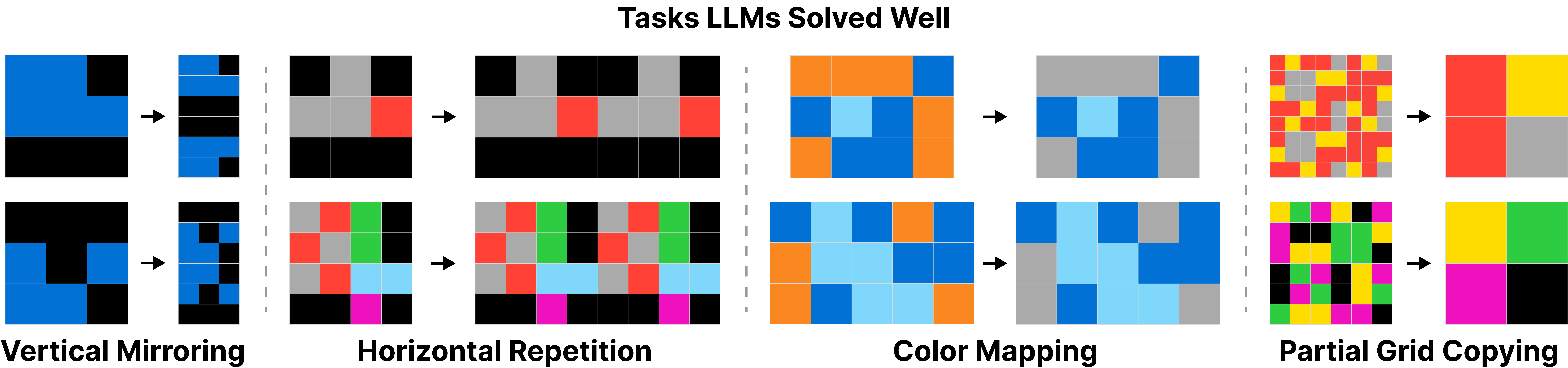}
    \captionsetup{width=\linewidth}
    \caption{Types of tasks where the LLM showed high accuracy. The LLM showed high accuracy in simple tasks such as pattern mirroring, pattern repetition, color mapping, and partial grid copying.}
    \label{fig:logical_coherence/well_solved}
\end{figure}

Nevertheless, in Section~\ref{sec:logical_coherence/inferential_coherence}, we identified eight tasks that the LLM could solve with an accuracy of 0.6 or higher. As shown in Fig.~\ref{fig:logical_coherence/well_solved}, these eight tasks consist of simple solutions such as mirroring, color mapping, and partial grid copying. These tasks shared a common characteristic of conceptual simplicity, utilizing only one of the four prior knowledge domains included in ARC: objectness, goal-directedness, numbers and counting, and basic geometry~\cite{chollet2019ARC}. For the 17 tasks that required the use of two or more prior knowledge domains, the LLM failed to solve any of the 100 augmented examples. The fact that the LLM could not solve any of the augmented examples, despite having solved the original ones, suggests that LLMs are not semantically coherent and may even indicate potential data leakage.

This comprehensive analysis demonstrates that while LLMs can solve certain simple pattern recognition tasks, they struggle with more complex reasoning that requires the integration of multiple concepts. The inability to coherently apply rules across augmented test examples, coupled with the generation of correct answers through incorrect reasoning, highlights significant limitations in both the inferential and semantic coherence of current LLM systems when tackling abstract reasoning tasks like those presented in ARC.

\subsubsection{Conclusion}
In Section~\ref{sec:logical_coherence}, we evaluated the logical coherence of LLMs by solving 100 ARC tasks using three different prompting techniques. Our results, showing accuracies ranging from 4\% to 12\%, demonstrate variability in reasoning performance depending on the prompting approach. Additionally, when experimenting with GPT-4o on 400 training tasks, the LLM showed a high accuracy of 20\%.

However, through an in-depth qualitative review, we demonstrated that the LLM's results may not be logically coherent. For the augmented test examples (100 for each solved task), the LLM only managed to achieve performance above 60\% in eight out of the 83 solved tasks. Furthermore, for 35 out of the 83 solved tasks, nearly half of the solution processes provided by the LLM were incorrect and could not derive the correct results. This analysis suggests that the LLM has failed to achieve human-level logical coherence.

The results of this study align with previous research asserting that logical problem-solving remains challenging for LLMs alone. One study~\cite{wang2023towards} found that LLMs can generate logically consistent reasoning with CoT prompting, even when their reasoning steps are flawed. Another study~\cite{zhang2024self} showed that LLMs struggle with accurate self-reflection in tasks like mathematical reasoning and translation. Additionally, research~\cite{tyen2024llms} revealed that LLMs often fail to detect errors in intermediate steps, exposing flaws in their reasoning process. While these studies suggest that providing more context or enforcing stronger self-reflection might improve logical reasoning~\cite{wei2022chain, wang2023towards, zhang2024self}, our findings indicate that these challenges persist, suggesting the issue may not be simply a lack of information about the problem. 
\subsection{Capability of LLMs 2: Compositionality}
\label{sec:compositionality}

\subsubsection{Motivation}
In Section~\ref{sec:compositionality}, we investigate compositionality, the second concept of LoTH.\footnote{While the Language of Thought Hypothesis principally uses the term `systematicity', this study employs `compositionality' as used in Fodor's paper. We use this term because compositionality encompasses a broader concept than systemicity.} Compositionality refers to the ability to generate complex linguistic expressions given simpler ones~\cite{fodor1988connectionism}. This characteristic allows individuals to effectively tackle more complex tasks by breaking sub-tasks down into simpler steps, supporting the notion that humans can solve more complex tasks when faced with them. Strong compositionality enables the resolution of complex tasks and facilitates transparent descriptions of the process, which is also an important aspect of LLMs. 

This section uses ARC to test the compositionality of LLMs. Previous studies have tested a model's compositionality by providing functions in the prompt that can be combined to solve tasks and then checking if the model can solve them~\cite{sinha2024survey}. Similarly, in this study, we also provide step-by-step functions, which we refer to as DSL (Domain Specific Language), and then conduct experiments to verify whether they can solve ARC tasks. Additionally, to understand why tasks might not be solved, we conducted further experiments on the model's comprehension of these functions. Therefore, we verify whether LLMs understand the meaning of the functions provided for ARC tasks and whether they can combine the functions appropriately to produce the desired results. The result of this experiment indicates that while LLMs sufficiently understand the functions and their relationship with images, their ability to decompose and combine functions to achieve the desired outcome is weak.

\subsubsection{Compositionality of LLMs}

\begin{figure}[ht]
    \centering
    \includegraphics[width=\columnwidth]{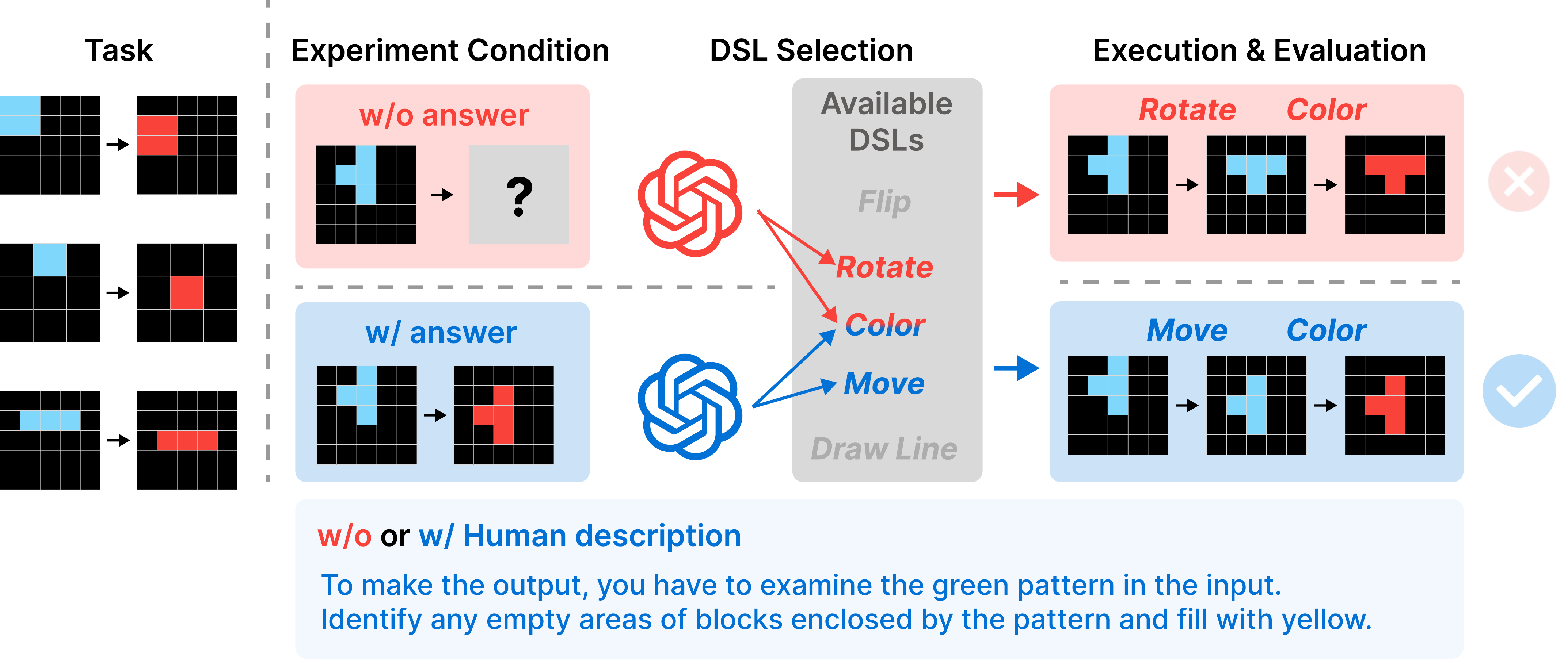}
    \caption{Overall process of DSL compositionality experiments. Before conducting the experiment, decisions are made on whether to provide 1) the test output and 2) a human description. During execution, the LLM analyzes the given demo examples to infer the rules and then selects the appropriate DSL steps from the DSL list to solve the test example. The chosen DSL steps are then applied to the test input grid within the DSL environment, which determines whether the answer is correct.}
    \label{fig:compositionality/compositionality_main_experiment}
\end{figure}

In the first experiment, to measure compositionality, we provided LLM with information about DSL and asked them to solve given ARC tasks. Fig.~\ref{fig:compositionality/compositionality_main_experiment} illustrates the structure of the entire experiment. If an LLM possesses sufficient compositionality, it should be able to select appropriate DSLs and their arguments for a given goal. However, in cases where the LLM failed to choose the correct DSL, we divided the conditions further to identify the cause. These conditions were whether the LLM understood the goal and the solution process. To analyze the results according to each condition, four types of experiments were conducted: 1) given only DSL, 2) given correct output along with DSL, 3) given human descriptions~\cite{Shin2023mclarc} to ARC test examples along with DSL, and 4) given both correct output grid and human descriptions along with DSL. Providing the correct output grid demonstrates compositionality based on knowing or not knowing the goal, while providing human descriptions shows the impact of natural language descriptions on compositionality.

We provided each DSL as a Python function. In this experiment, we used 19 types of DSL capable of solving ARC tasks. The prompts commonly included a brief explanation of ARC, DSL function code with comments, DSL usage examples, demonstration examples of tasks, inputs for the test examples, and object information of the test inputs. Object information is one of the crucial parameters in solving ARC tasks, which is why we added it to the prompt. We used the PnP algorithm~\cite{park2023unraveling} to extract object information from ARC tasks. The LLM returned a JSON-formatted string representing the chosen DSL and arguments at each step, which was used to verify whether the LLM reached the correct test output with an appropriate combination of DSL and arguments. We used the most recent model, GPT-4o, for this experiment.

\begin{figure}[ht]
   \centering
   \includegraphics[width=\columnwidth]{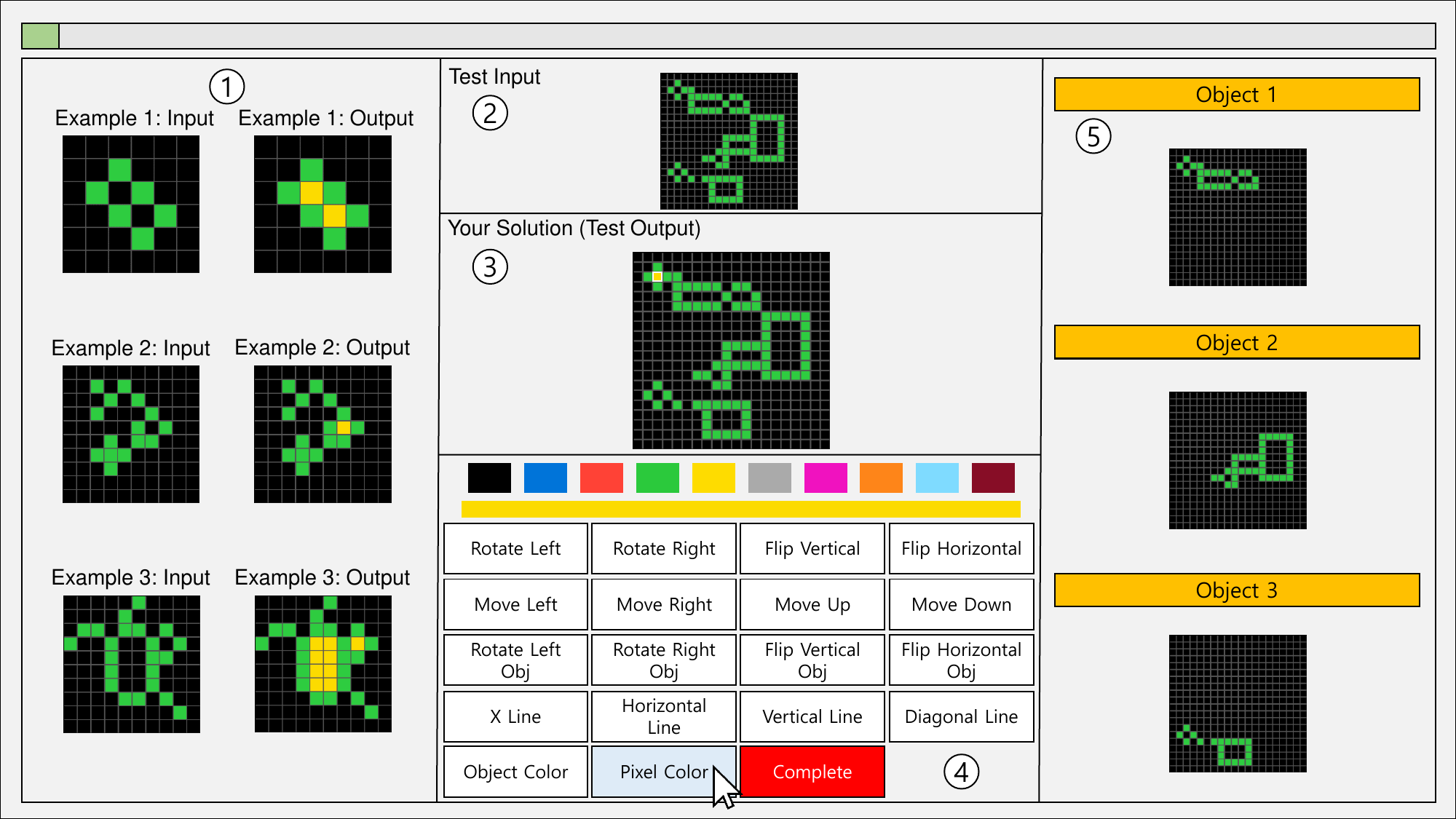}
   \caption{Performance measurement tool for human participants: The interface consists of five numbered sections: 1) Demonstration examples from the given ARC task; 2) Test input grid for the given task; 3) Current grid for participants' responses; 4) Available DSLs; and 5) Object information automatically extracted using the PnP method. Participants must utilize the provided DSLs and object information to construct appropriate solutions, with both grid pixels and objects available as function parameters. Participants are meant to click complete if the current grid seems to be the solution.}
   \label{figures/compositionality/3.2_Tool}
\end{figure}

\revise{Lastly, to establish a baseline, we conducted human experiments. We developed a specialized tool (Fig.~\ref{figures/compositionality/3.2_Tool}) that provides the same information available to LLMs: ARC task demonstration examples, initial test input, current grid state, DSL functions, and object information extracted through PnP. Seven participants were constrained to solve the tasks using only the same DSLs given to the LLMs. Through these experiments, we identified that among the 800 publicly available ARC tasks, 158 tasks were solvable within 10 DSL steps using the given operations. Subsequently, all experiments in Section~\ref{sec:compositionality} were conducted on this subset of solvable tasks.}

The experimental results are shown in Table~\ref{tbl:compositionality/main_experiment_result}. For LLM experiments, an average accuracy of 9\% was observed when the test output was provided, and 3\% without the test output. Compositionality strengthened when human explanations were included in the prompt, showing a similar improvement rate to the test output condition. Cronbach's alpha measurements showed consistency in responses, with all four experiments scoring above 0.7. 

\revise{For human experiments, participants solved an average of 137 tasks, achieving approximately 86\% accuracy on solvable tasks. This significant performance gap between LLMs (3--14\%) and humans (86\%) suggests that despite having access to the same information and tools, LLMs face fundamental challenges in DSL composition that humans can naturally overcome.}

\begin{table}[ht]
\small
\centering
\caption{Average accuracy from 10 repeated experiments based on the presence or absence of test output and human descriptions. The values in parentheses are Cronbach's alpha, and `X' indicates that this condition was not applicable for human experiments as they performed the tasks without test outputs. In all the results in the table, Cronbach's alpha is greater than 0.7, indicating consistency.}
\begin{tabular}{c c c c} 
\toprule
\textbf{} & \textbf{w/o Human Description} & \textbf{w/ Human Description} & \textbf{Accuracy of Human}\\
\midrule
\textbf{w/o Test Output} & 3\% (0.93) & 8\% (0.97) & 86\% \\
\textbf{w/ Test Output}  & 9\% (0.96) & 14\% (0.96) & X \\
% \midrule
% Cronbach's alpha & 0.960(w/ testoutput) & 0.928(w/o testoutput)  \\
 \bottomrule
\end{tabular}
\label{tbl:compositionality/main_experiment_result}
\end{table}

\newpage

\subsubsection{Analysis of compositional failures resulting from DSL misinterpretation}

The issue is that the average accuracy described in Table~\ref{tbl:compositionality/main_experiment_result} doesn't solely reflect compositionality. DSL provides a step-by-step manner to represent solution steps in ARC tasks. When we use a DSL to solve these tasks, we can think about the likelihood of choosing the right DSL for each step in two parts: 1) How well LLMs understand the DSL: This is reflected in how accurately it can predict the next grid when given the DSL instructions. 2) How necessary each predicted grid is in creating the final solution: This relates to how well the steps fit together to solve the task.

The overall chance of picking the correct DSL for all steps depends on both of these factors working together. To solve a task, all DSLs must be correct for 10 steps. Based on our preliminary analysis, we modeled this relationship as a multiplicative interaction between DSL understanding and compositional difficulty, as shown in Eq.~\ref{eq:calculate_compositional_ability}. In this equation, \(n\) represents the DSL sequence length, \(w_n\) represents the number of tasks that need \(n\) steps to solve, \(p\) represents the single-step accuracy, and \(x\) represents the difficulty of composition for each task. We assumed that the LLM's compositionality could vary depending on the information provided to the LLM and the task.

% The issue is that the average accuracy described in Table~\ref{tbl:compositionality/main_experiment_result} doesn't solely reflect compositionality. When we use a DSL to solve ARC tasks, we can think about the likelihood of choosing the right DSL for each step in two parts: 1) How well LLMs understand the DSL: This is reflected in how accurately it can predict the next grid when given the DSL instructions. 2) How necessary each predicted grid is in creating the final solution: This relates to how well the steps fit together to solve the task. The overall chance of picking the correct DSL for all steps depends on both of these factors working together. To solve a task, all DSLs must be correct for 10 steps. Reflecting this, we can estimate as shown in Eq.~\ref{eq:calculate_compositional_ability} below. In this equation, \(n\) represents the DSL sequence length, \(w_n\) represents the number of tasks at step \(n\), \(p\) represents the single-step accuracy, and \(x\) represents the difficulty of composition for each task. We assumed that the compositionality of the LLM might vary depending on the task complexity and the information provided to it.

\begin{equation}
y = \frac{\sum_{n=1}^{10} w_n \cdot (p \cdot x)^n}{\sum_{n=1}^{10} w_n}
\label{eq:calculate_compositional_ability}
\end{equation}

\begin{figure}[ht]
    \centering
    \includegraphics[width=1\columnwidth]{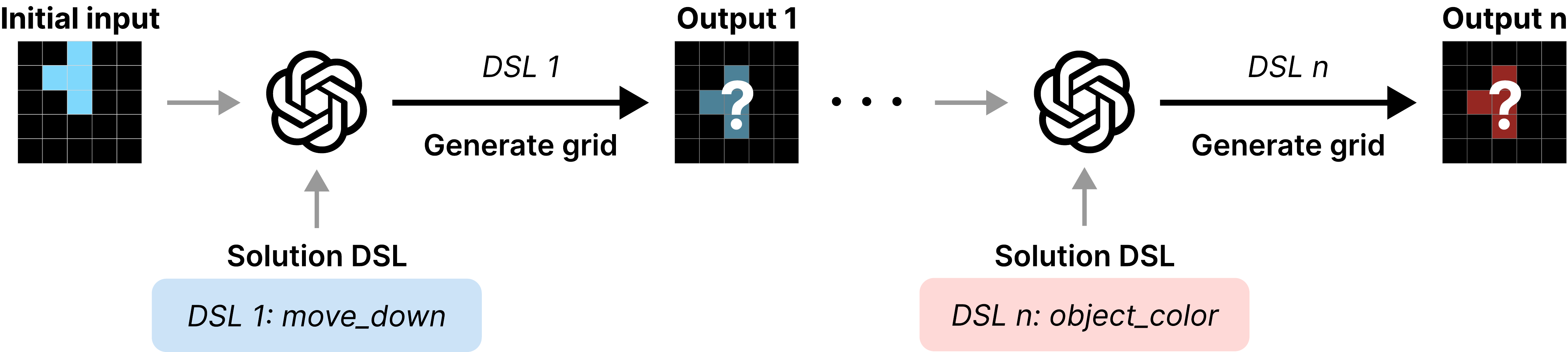}
    \caption{Overall process of an experiment in understanding DSL. The task for the LLM is to accurately generate a grid transformed by the DSL when given a grid and its corresponding DSL. Each task involves a DSL sequence ranging from 1 to 10 steps, using trajectories previously solved by humans.}
    \label{fig:compositionality/understands_dsl}
\end{figure}

To determine the task accuracy considering only the compositional difficulty, we must estimate the \(y\) value when \(p=1\). Therefore, we conducted an additional experiment, as shown in Fig.~\ref{fig:compositionality/understands_dsl} to verify the probability of not finding an appropriate DSL due to the inability to predict the output grid when selecting a DSL.

In the additional experiment, we focused on 158 tasks selected from the 800 publicly available ARC tasks, specifically choosing those that could be solved within 10 DSL steps. We checked how accurately the LLM could generate the correct output grid when given both the DSL and ARC input grid. Each task was repeated 10 times to ensure reliability. For these experiments, we provided the LLM with correct DSL operations and argument chains created by human solvers. Among multiple human solutions, we prioritized those with the shortest step length to minimize complexity. Since both the input grid and DSL instructions were provided, LLM should be able to produce the correct output grid regardless of the sequence length, assuming perfect DSL comprehension.

% In the additional experiment, we checked how accurately the LLM could generate the correct output grid when given the DSL and ARC input grid. The experiment was conducted on 158 tasks, each repeated 10 times. The correct DSL and its argument chain for ARC tasks, created by humans while solving the tasks, were provided to the LLM. We prioritized using the solution with the shortest step length among the solutions provided by the human solvers. If the LLM can predict all subsequent output grids when given a specific DSL, it should be able to correctly produce the output grid regardless of the step length, since both the input grid and DSL were provided.

\noindent
\begin{minipage}{0.46\textwidth}  % 너비를 0.5에서 0.45로 줄임
\revise{The relationship between DSL sequence length and LLM's prediction accuracy is shown in Fig.~\ref{fig:compositionality/understanding_dsl_accuracy}. As the required sequence length increases, we observe a clear decline in the model's ability to predict correct output grids. Based on these observations, we calculated a weighted average single-step accuracy \(p\) using Eq.~\ref{eq:calculate_dsl_understaing_degree}, where \(w_n\) represents the number of tasks with sequence length \(n\), and \(a_n\) represents the prediction accuracy for that length. This calculation yielded an estimated single-step accuracy of 81\%, indicating that errors compound significantly with longer sequences.}
\begin{equation}
    p = \frac{\sum_{n=1}^{10} w_n \cdot a_n}{\sum_{n=1}^{10} w_n}
    \label{eq:calculate_dsl_understaing_degree}
\end{equation}
\end{minipage}%
\hspace{0.03\textwidth}%  % 5%의 여백 추가
\begin{minipage}{0.48\textwidth}
   \centering
   \includegraphics[width=\textwidth]{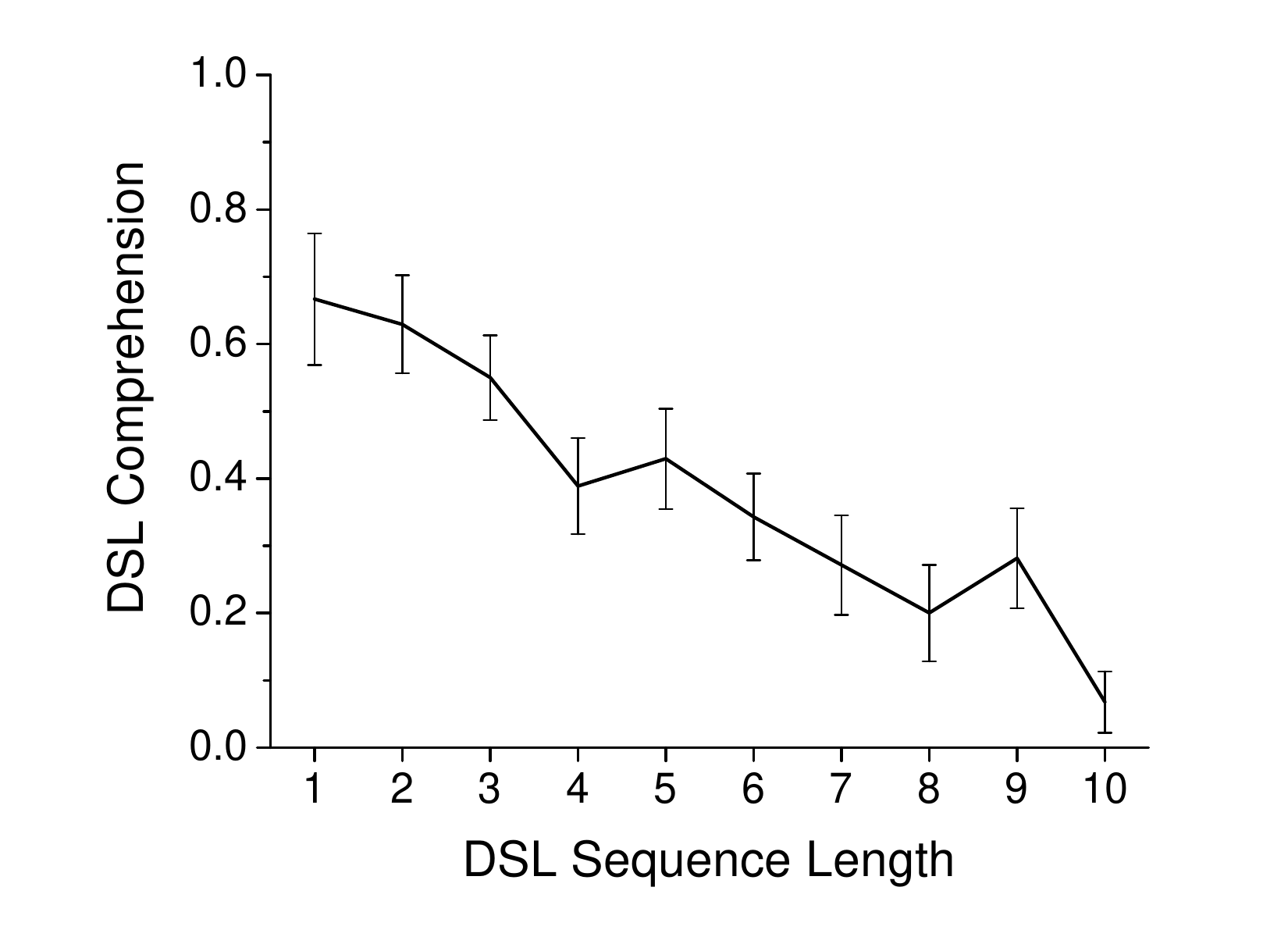}
   \captionof{figure}{\revise{Model's accuracy in predicting output grids for DSL sequences of varying lengths (1-10 operations). The y-axis shows the success rate of grid prediction after applying the given DSL operations. The DSL comprehension tends to decrease with longer sequences.}}
   \label{fig:compositionality/understanding_dsl_accuracy}
\end{minipage}

Table~\ref{tbl:compositionality/optimal_experiment} presents the estimated accuracy when assuming perfect DSL understanding (\(p\) = 1.0, adjusted from the observed \(p\) = 0.8). This adjustment isolates the impact of compositional ability alone, showing that under ideal conditions with both test output and human descriptions provided, nearly 30\% of tasks could be solved. The consistent 10 percentage point improvement observed when adding either the correct answer or natural language descriptions suggests that each element reduces the compositional difficulty (represented as \(x\) in Eq.~\ref{eq:calculate_compositional_ability}) of the tasks.

\begin{table}[ht]
\small
\centering
\caption{
The table of results shows the accuracy estimates obtained using Eq.~\ref{eq:calculate_compositional_ability}, assuming that the LLMs have a 100\% understanding of DSL, meaning the single-step accuracy \( p \) is 1.0.}
\begin{tabular}{c c c} 
\toprule
\textbf{} & \textbf{w/o Human Description} & \textbf{w/ Human Description}  \\
\midrule
\textbf{w/o Test Output} & 5\% & 15\%  \\
\textbf{w/ Test Output}  & 17\% & 29\%  \\
% \midrule
% Cronbach's alpha & 0.960(w/ testoutput) & 0.928(w/o testoutput)  \\
 \bottomrule
\end{tabular}
\label{tbl:compositionality/optimal_experiment}
\end{table}

\subsubsection{Case Study: Enhancement of Compositionality through Human Descriptions}

One notable observation was the enhanced compositionality when human descriptions of problem-solving methods were included in prompts. To investigate how LLMs could solve tasks with human descriptions, we analyzed the solution processes of 13 additional tasks solved when human descriptions were provided. Results indicate that human descriptions facilitate task input and action abstraction, thereby improving problem-solving capabilities. For instance, LLMs fail to recognize patterns in the correct output without descriptions; however, they immediately identify patterns such as an `X' shape with descriptions. These findings suggest the potential to enhance LLMs' reasoning performance by incorporating abstracted task information.

% , as illustrated in Fig.~\ref{fig:compositionality/human_description}

% \begin{figure}[ht]
%     \centering
%     \includegraphics[width=0.8\columnwidth]{figures/3.2_compositionality/human_description.pdf}
%     \caption{Comparison of LLM's DSL selection with and without human descriptions. A human description helps find the necessary DSL for problem-solving by aiding in the required abstraction. Without it, the lack of proper abstraction prevents finding the correct DSL.}
%     \label{fig:compositionality/human_description}
% \end{figure}

\subsubsection{Conclusion}

In Section~\ref{sec:compositionality}, experiments using ARC and DSL were conducted to measure the compositionality of LLMs. The results led to three conclusions. First, LLMs could predict the output grid when DSL was applied to the input with an average accuracy of about 81\%. However, as the sequence length increased, the accuracy decreased, which appears to be due to cumulative errors. Second, when not given the correct answer, LLMs selected the correct DSL only 3\% of the time, indicating a lack of ability both in inferring rules to predict the correct output grid and in selecting the appropriate DSL to reach the expected output. Finally, when human descriptions were added, the accuracy in choosing DSL increased to a level similar to when the correct answer was provided. Analysis of this process suggested that this improvement was due to linguistic abstraction of the ARC task and DSL combinations.

Previous studies have emphasized LLMs' limitations in combining simple elements to create new meanings, revealing struggles with compositionality. One study shows Transformers exhibit significant performance drops when tested on new function combinations, indicating challenges in systematically generalizing knowledge~\cite{hupkes2020compositionality}. Another study introduced datasets like SADE to evaluate LLMs' ability to process visual and textual information, suggesting they still struggle with tasks like understanding negations and grasping complex content~\cite{ma2024examination}. A further study examined how well LLMs can break down complex instructions or build them from simple ones. These found that while LLMs improve at understanding simple tasks by learning complex ones, they struggle with complex tasks when starting from simpler ones~\cite{yang2024exploring}. These findings across studies point to ongoing challenges in LLMs' ability to connect simple and complex elements, highlighting their compositionality limitations.

% \newpage
\subsection{Capability of LLMs 3: Productivity}
\label{sec:productivity}

\subsubsection{Motivation}

In Section~\ref{sec:productivity}, we investigate the third concept of LoTH: productivity. Productivity refers to the ability to generate unseen representations based on observed data~\cite{fodor1988connectionism}. This characteristic enables humans to imagine diverse situations from a single phenomenon, facilitating efficient learning without the need for repetitive data exposure. Similarly, when endowed with this ability, LLMs are expected to excel in unseen tasks, making productivity a crucial function of essential reasoning. The capacity to generate new pairs within a constrained set of rules is particularly valuable for solving ARC tasks, highlighting the need for productivity. In this section, we will assess productivity by evaluating the validity of LLM-generated examples based on given example pairs from ARC tasks.

While productivity ideally involves testing for infinite generative capacity, practical limitations necessitate alternative approaches. The challenge lies in demonstrating that a system can produce an unlimited number of novel, meaningful outputs from a finite set of inputs and rules. Previous studies have addressed this challenge by examining whether valid outputs can be produced under added constraints \cite{van2004lack, lake2018generalization, hupkes2020compositionality}. These constraints serve to create a more manageable testing environment while still allowing for the assessment of generative capabilities. Following this methodology, our study investigates how effectively LLMs can generate valid outputs when presented with an ARC task and its underlying conceptual rule. This approach allows us to evaluate productivity within a controlled framework while still capturing the essence of generative capacity.

To understand how well LLMs can generate new expressions based on inherent logical concepts, we conduct experiments using ARC tasks. Productivity in this context involves two main steps: 1) inferring specific rules for image generation from example images and natural language expressions, and 2) applying these rules to generate new, unseen images. However, as explored in previous sections, the standard approach to solving ARC tasks is insufficient to confirm these two processes. Therefore, we propose a novel experiment: \textit{Given an ARC task and a basic rule shared with similar ARC tasks, can LLMs generate valid examples of the given task?} If LLMs can understand the relationship between the given ARC task and the abstract rule, they should be able to infer specific rules for the task and generate new valid examples. Through this, we aim to determine whether LLMs can imitate the productivity of human thinking in generating novel solutions.

\begin{figure}[ht]
    \includegraphics[width=\textwidth]{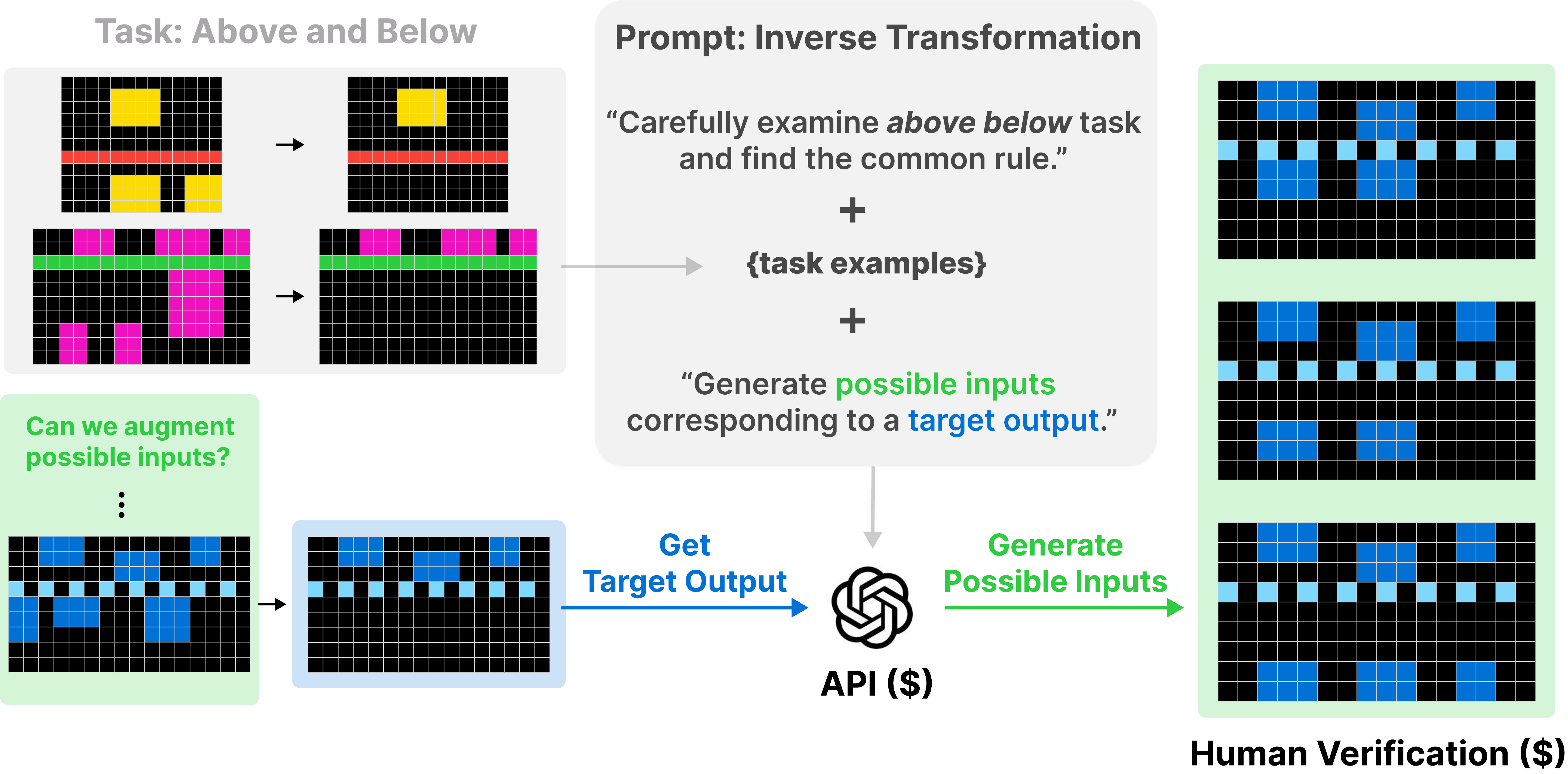}
    \vspace{-0.5cm}
    % \captionsetup{width=.85\linewidth}
    \caption{Overall process of possible input generation with the Inverse Transformation Prompt (ITP). With ITP and one example of the task, LLMs generate input candidates of the output for the given example. If these generated inputs are valid, pairs created by these inputs and the given output can become new examples.}
    \label{fig:productivity/generation_process}
\end{figure}

\subsubsection{Validity of Augmentation}

\revise{To evaluate whether LLMs can infer their own generation rules given ARC examples and create new tasks by appropriately applying these rules, we rigorously controlled the prompts. While ARC provides a diverse set of tasks, it lacks systematic categorization and explicit rules for each task. Therefore, we utilized ConceptARC~\cite{arseny2023ConceptARC}, which maintains the same format as ARC but provides categories for each task, making it more suitable for our experimental design. We provided LLMs with two types of prompts: example pairs from the ConceptARC task and abstract rule descriptions applicable to similar tasks. In this step, one example pair served as the basis for generation, while the others were used to infer task-specific rules. Based on the ConceptARC framework, the tasks are organized into 16 distinct categories. For each category within ConceptARC, a corresponding abstract rule ensures that tasks within the same category adhere to the identical abstract rule.}

% To evaluate whether LLMs can infer their own generation rules given ARC examples and create new tasks by appropriately applying these rules, we rigorously controlled the prompts. LLMs receive two types of prompts: example pairs included in the ARC task and abstract rule descriptions applicable to similar tasks. In this step, one example pair served as the basis for generation, while the others were used to infer task-specific rules. Based on the category of ConceptARC~\cite{arseny2023ConceptARC}, which follows the same format as ARC but consists of differently constructed tasks, the tasks are organized into 16 distinct categories according to human classification criteria. For each category within ConceptARC, a corresponding abstract rule ensures that tasks within the same category adhere to the identical abstract rule.

We proposed the Inverse Transformation Prompting (ITP), a prompting technique for this experiment. ITP instructs LLMs to generate multiple valid examples by leveraging both the ConceptARC task and its associated abstract rules. Fig.~\ref{fig:productivity/generation_process} demonstrates how LLMs generate new examples, given the ConceptARC task and the corresponding ITP. Using this method, LLMs produce multiple inputs that can pair with the output from one example of the task. This example for generation is excluded from the ITP. If LLMs understood the ConceptARC task rules provided through ITP, the new example pairs generated by LLMs would be suitable as examples of the task.

ITP is based on a many-to-one approach to achieve two advantages. First, the input-only generation method is more data-efficient than generating both input and output, as existing task outputs can be reused without modification. Since all tasks in ConceptARC have example pairs, reusing these examples fully utilizes the given data. ITP allows a single ConceptARC task to be reused multiple times. In particular, using ITP can further increase data efficiency by allowing one ConceptARC task to be reused multiple times by changing the order of examples. Secondly, ITP increases the likelihood of generating valid responses. Through simulations, we observed that inferring inputs from outputs is more likely to generate valid results than the reverse. Because generating input from output is subject to relatively fewer constraints, there is a wide range of acceptable outcomes.

\begin{figure}[ht]

    % Subfigure (a)
    \begin{subfigure}[b]{0.465\textwidth}
        \includegraphics[width=\textwidth]{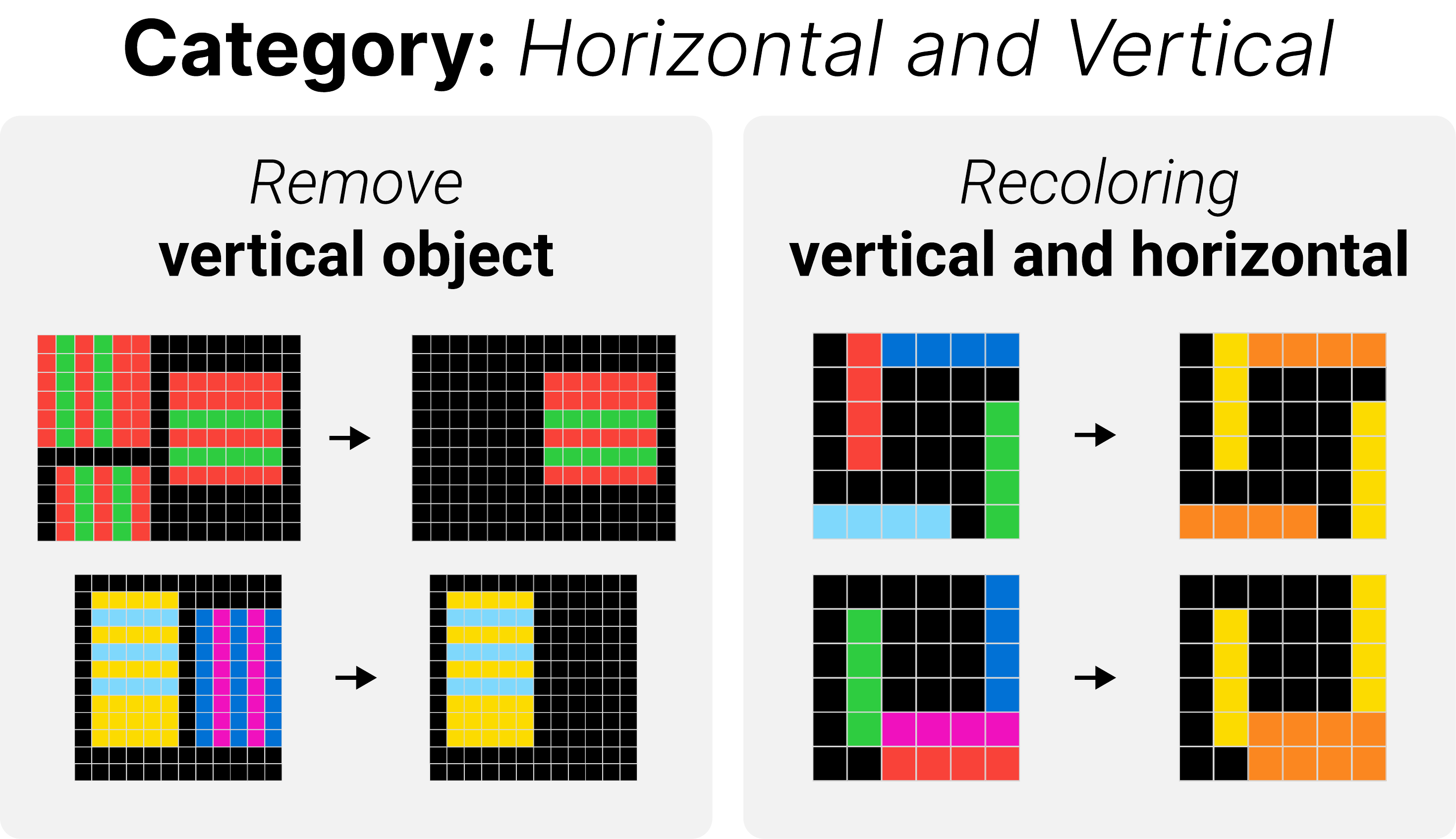}
        \captionsetup{width=.95\linewidth}
        \caption{Even within the same category, tasks can showcase varied objectives and complexities. The left task requires removing vertically striped objects, while the right focuses on recoloring objects based on their orientation.}
        \label{fig:productivity/ambiguous_category}
    \end{subfigure}
    \hfill % 서브피겨 사이의 공간을 조정
    % Subfigure (b)
    \begin{subfigure}[b]{0.465\linewidth}
        \includegraphics[width=\textwidth]{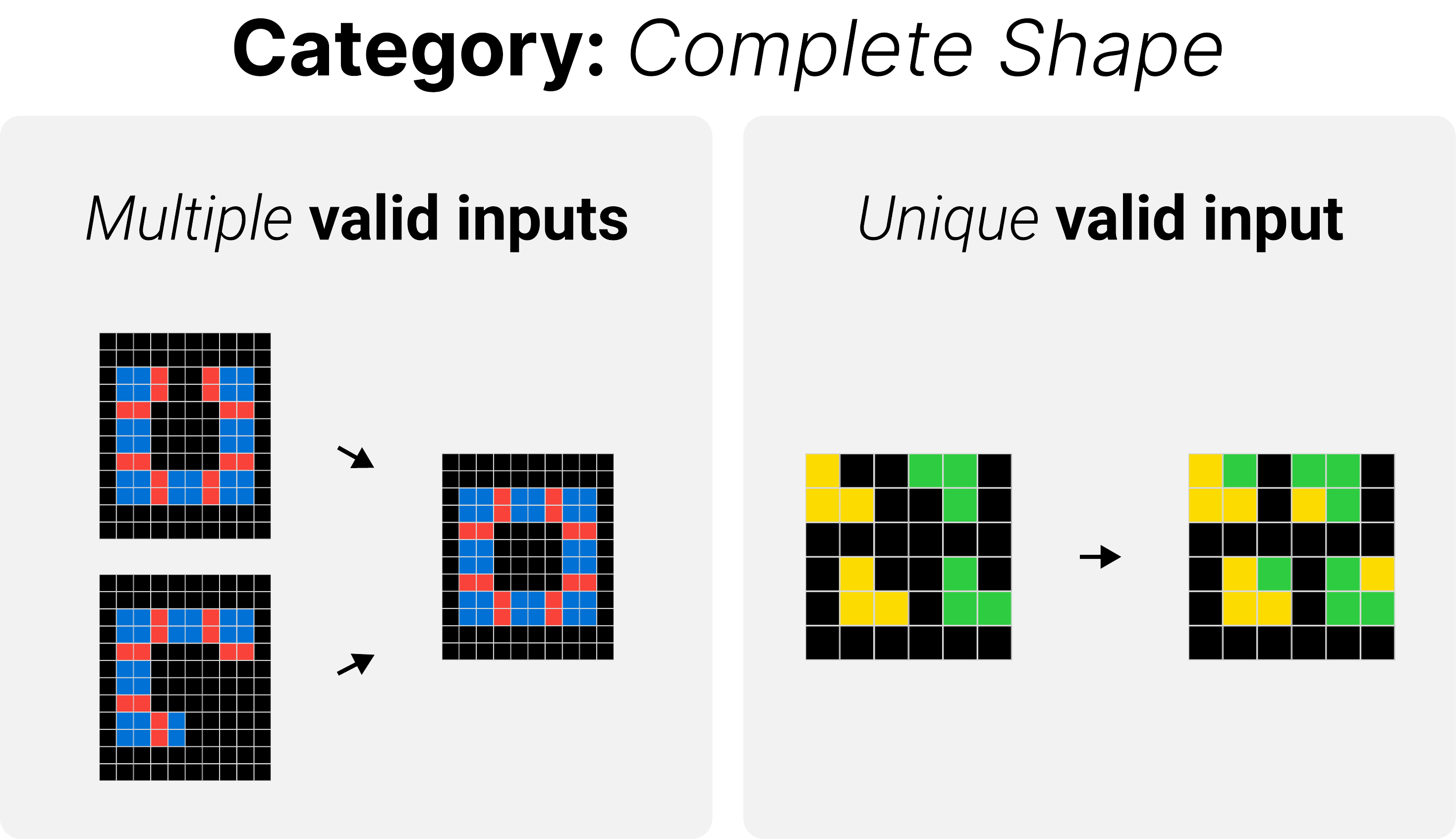}
        \captionsetup{width=.95\linewidth}
        \caption{Depending on the task, there may be multiple or a unique input for an output. The left shows a task of completing a square with various inputs, and the right combines specific shapes, leading to a unique input.}
        \label{fig:productivity/unique_input}
    \end{subfigure}
    
    \caption{There are two challenges when LLMs generate examples through ITP: (a) task diversity within categories and (b) inflexibility in task-specific examples. These may cause difficulties in the process of LLMs generating examples through ITP.}
    \label{fig:productivity/ITP_limitations}
\end{figure}

In the process of creating ITP, we encounter two challenges. First, according to the ConceptARC category, tasks within the same category can have different specific objectives. Fig.~\ref{fig:productivity/ambiguous_category} illustrates that there are various types of tasks with the same category. For example, even within the same category, the core solution for one task might involve removal, while another might focus on recoloring. This variation highlights that abstract rules given in the same sentences for each category may not be sufficient to cover various types of tasks. Second, there were ConceptARC tasks that made it impossible to infer multiple inputs from a single output (Fig.~\ref{fig:productivity/unique_input}). In such cases, there was only one valid input. Although we tried to take these cases into account while writing the ITP, these challenges nevertheless harmed the experimental results.

Before analyzing the experimental results, it was necessary to redefine the evaluation metric to reflect a shift in focus from solving tasks to generating valid examples. As previously explained, for a given example of a particular task, we generated valid inputs that could be paired with the corresponding output. To successfully generate these inputs, the LLM must derive the task's specific rules through its ITP and apply them to the output to create valid inputs. In this experiment, we evaluated whether all generated inputs were valid for each task. This metric assesses both the LLM’s understanding of the correct rules and its ability to generate valid examples based on those rules. Consequently, this experiment systematically evaluates the LLMs' capability to generate logical and valid demo pairs, enhancing our understanding of their ability to create new representations.

\begin{table}[ht]
\centering
\renewcommand{\arraystretch}{1.0}
\caption{The ratio of valid examples among examples generated for each category of ConceptARC.}
\label{tbl:productivity/generation_validity}
\resizebox{0.52\textwidth}{!}
{\begin{tabular}{l c c c} 
\toprule
\textbf{Category} & \textbf{Generated} & \textbf{Valid} & \textbf{Validity} \\ [0.5ex] 
\midrule
Above Below & 158 & 34 & 21.52\% \\ 
Center & 236 & 35 & 14.83\% \\ 
Clean Up & 183 & 83 & 45.36\% \\ 
Complete Shape & 147 & 37 & 25.17\% \\ 
Copy & 153 & 4 & 2.61\% \\ 
Count & 202 & 29 & 14.36\% \\ 
Extend To Boundary & 167 & 8 & 4.79\% \\ 
Extract Objects & 176 & 21 & 11.93\% \\ 
Filled Not Filled & 203 & 29 & 14.29\% \\ 
Horizontal Vertical & 114 & 7 & 6.14\% \\ 
Inside Outside & 191 & 24 & 12.57\% \\ 
Move To Boundary & 165 & 12 & 7.27\% \\ 
Order & 162 & 26 & 16.05\% \\ 
Same Different & 246 & 76 & 30.89\% \\ 
Top Bottom 2D & 255 & 59 & 23.14\% \\ 
Top Bottom 3D & 215 & 25 & 11.63\% \\ 
\midrule
Total & 2,913 & 509 & 17.12\% \\ 
\bottomrule
\end{tabular}
}
\end{table}

%Based on 160 ARC tasks classified by ConceptARC, we evaluated the validity of a total of 2,913 generated examples. The valid generation ratio on average was approximately 17.1\%, while the rest were invalid. As we mentioned before, the validity of results was determined by human judgment regarding whether the generated task adhered to the given rule. Results in Table~\ref{tbl:productivity/generation_validity} indicate that LLMs demonstrate a certain level of performance in generating examples consistent with the rule. However, since the criteria for validating the generated results as valid or invalid are weak, there is a limitation that results cannot be used before data post-processing even if an infinite number of results can be created.

Based on 160 ConceptARC tasks, we evaluated the validity of 2,913 generated examples. The average valid generation ratio was approximately 17.1\%, with the remaining examples deemed invalid. As previously mentioned, the validity of the generated examples was determined by human judgment, assessing whether the generated tasks adhered to the analogical rules required to solve the task. The results in Table~\ref{tbl:productivity/generation_validity} show that LLMs exhibit a degree of capability in generating examples that align with the specified rules. However, there is a limitation due to weak criteria for determining validity: even if infinite results can be generated, they cannot be reliably used without post-processing the data.

\subsubsection{Case Study: Invalid Production}
We analyzed the generated inputs to investigate the reasons behind LLMs' inability to produce valid inputs for ConceptARC tasks. Two major limitations were observed when LLMs generated new ConceptARC tasks. First, LLMs tended to simply copy inputs rather than infer meaningful rules from given example pairs. As shown in Fig.~\ref{fig:productivity/wrong_generation}, this occurred repeatedly despite attempts to prevent it through prompts. Second, LLMs failed to properly consider the steps needed to generate inputs from given outputs. This frequently resulted in the creation of examples that could not be solved by the specific rules of the task. For instance, in cases where all vertices of a square were erased in the input, it became impossible to determine the color of the vertices, making it infeasible to infer the given output. These limitations suggest that LLMs lack an understanding of the semantics applicable to ConceptARC tasks and the ability to compose these semantics according to constraints.

\begin{figure}[ht]
    \includegraphics[width=\textwidth]{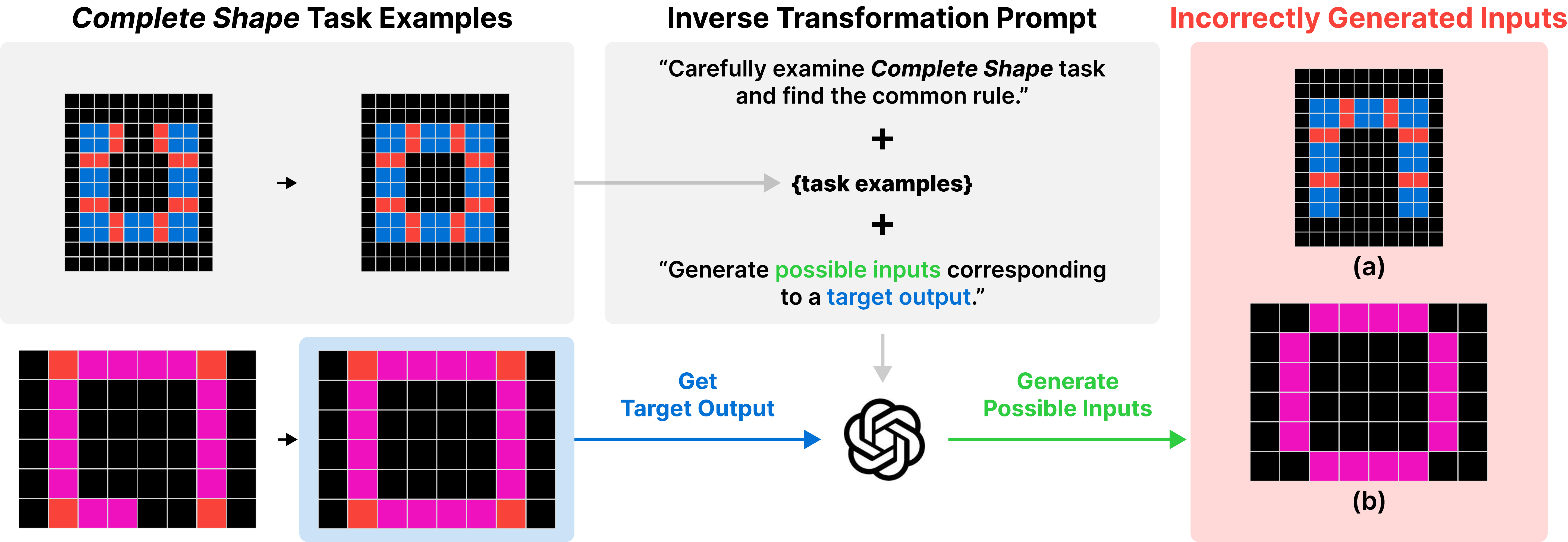}
    % \captionsetup{width=.85\linewidth}
    \caption{Two incorrect generations for the task of completing the square shape: (a) The LLM generates the input from another example's output, and (b) the input does not provide enough information to infer the square's corner colors.}
    \label{fig:productivity/wrong_generation}
\end{figure}

\iffalse
\begin{figure}[ht]
    \begin{subfigure}[b]{0.465\linewidth}
    \includegraphics[width=\textwidth]{figures/3.3_productivity/wrong_generation.pdf}
    \captionsetup{width=.95\linewidth}
    \caption{(a) LLM creates this input from the output of another example.}
    \label{fig:productivity/copying_other_input}
    \end{subfigure}
    \begin{subfigure}[b]{0.465\linewidth}
    \includegraphics[width=\textwidth]{figures/3.3_data_aug/wrong_generation_b.png}
    \captionsetup{width=.95\linewidth}
    \caption{(b) It is impossible to infer the color of the corners of the square based on this input.}
    \label{fig:productivity/lack_of_information}
    \end{subfigure}
    \caption{Two examples of the wrong generations for the task of completing the square shape.}
    \label{fig:productivity/wrong_generation}
\end{figure}
\fi

\subsubsection{Conclusion}

In Section~\ref{sec:productivity}, we conducted experiments to confirm the productivity of LLMs by assessing whether they can understand given tasks in abstracted representations and generate valid new examples based on abstracted rules. Although it is known that LLMs have great strengths in creating creative works, our experimental results reveal that LLMs are weak in understanding rules and producing creations that adhere to those rules. Moreover, the observed limitations highlight a critical gap in LLMs' ability to engage in higher-level reasoning and abstraction, which are essential for successfully solving tasks that require an understanding of underlying principles rather than surface patterns. These results suggest that when LLMs generate outputs, they tend to mimic human-created results rather than truly understanding and applying rules. This makes it difficult for LLMs to reach the level of generation that humans can achieve.

Similarly, previous studies have shown similar results in measuring the productivity of AI models. Researchers tested how well pre-LLM models generalize to novel command combinations~\cite{lake2018generalization, van2004lack}. Their findings revealed strong performance on trained data but weaknesses in generating responses to unseen commands. Some researchers argued that LLMs struggle with generation under complex constraints and proposed improved models to address this issue~\cite{koh2024generating, liu2024llm4gen}. They propose novel frameworks to enhance LLMs for generating desired outputs when complex constraints are introduced, rather than relying solely on the base models. These researches share similarities with our study, which encountered difficulties in augmenting valid tasks based on complex rules.

% Similar results have been observed in previous studies measuring the productivity of AI models. Researches tested how well pre-LLM models generalize to novel command combinations~\cite{lake2018generalization, van2004lack}. Their findings revealed strong performance on trained data but weakness in generating unseen commands.  argued that existing LLMs struggle with generation under complex constraints and proposed an improved model to address this issue~\cite{koh2024generating, liu2024llm4gen}. These studies propose novel frameworks to enhance existing LLMs for generating desired outputs when complex constraints are introduced, instead of relying solely on the base models. These research efforts share similarities with our study, which encountered difficulties in augmenting valid ARC tasks based on complex rules.

%To transparently analyze the process, it is suggested to conduct additional experiments based on the ToT prompting technique to provide step-by-step generation.

\section{Discussion}

Through the three experiments in Section~\ref{sec:experiments}, we have observed that LLMs demonstrate strengths in understanding and manipulating both image and text inputs. However, they still exhibit weaknesses in logical inference, sequential planning based on understanding, and generating unseen images according to predefined rules. We will conclude by introducing the current research directions aimed at further enhancing LLMs' ability and outlining the goals after solving ARC.

\subsection{What Should LLMs Possess to Solve ARC?}
\label{sec:discussion/arc_and_llm}

Based on the experimental results of Section~\ref{sec:experiments}, it is evident that LLMs still cannot solve ARC effectively. This is attributed to the deficiencies in logical coherence, compositionality, and productivity. How can we improve the inference capabilities of LLMs? In this section, we explore directions to enhance LLMs from the perspectives of abstraction knowledge and reasoning.

\subsubsection{Abstract Knowledge}
% To solve ARC, the first challenge is the ability to extract the implicit information contained within ARC. Xu et al.~\cite{xu2023graphs} argued that object-based representation is crucial for solving ARC and proposed ARGA, which converts given example grids into graphs. Their follow-up study~\cite{xu2023llms} involved LLM solving ARC tasks using information obtained from ARGA and showed notable performance for object-based ARC tasks. However, these studies have a fundamental weakness in that they cannot be applied to ARC tasks without objects. Since only about 40\% of ARC tasks contain object concepts~\cite{xu2023graphs}, this method cannot be applied to more than half of the tasks. Wang et al. on the other hand, improved the abstraction ability of LLMs to some extent with a graph-form dataset consisting of 221K textual descriptions called AbsPyramid~\cite{wang2023abspyramid}, and also proposed a framework called AbsInstruct~\cite{wang2024absinstruct} utilizing this dataset. Attempting to structure sentences can be an effective abstraction method for natural language, but its effectiveness cannot be seen in tasks that do not contain sentences.

To solve ARC, the first challenge lies in extracting its implicit information. Xu et al.~\cite{xu2023graphs} emphasized the importance of object-based representation and proposed ARGA, which transforms example grids into graphs. Their follow-up study~\cite{xu2023llms} demonstrated notable performance in object-based ARC tasks by leveraging ARGA-generated information. However, these methods have a critical limitation: they are inapplicable to ARC tasks without objects. Since only about 40\% of ARC tasks involve object concepts~\cite{xu2023graphs}, this approach cannot address more than half of the tasks. Wang et al.~\cite{wang2023abspyramid} partially enhanced LLM abstraction with a graph-form dataset, AbsPyramid, containing 221K textual descriptions, and proposed a framework called AbsInstruct. While structuring sentences can effectively abstract natural language, this approach is ineffective for ARC, which does not involve textual data.

\subsubsection{Reasoning}
Another challenge for LLMs in ARC is the vast search space. A promising approach involves enabling LLMs to generate DSLs themselves. Rajani et al.~\cite{rajani2019explain} introduced CAGE, which prompts LLMs to generate explanations before generating answers. Subsequently, Wang et al.~\cite{wang2024hypothesis} reported improved results by having LLMs generate DSLs based on hypotheses they set themselves. Additionally, active research is underway on prompting techniques applying algorithmic approaches. Zhou et al.~\cite{zhou2022teaching} demonstrated enhanced inference performance in LLMs by applying in-context learning. Follow-up research is being conducted following CoT and ToT. For example, CoT-SC~\cite{wang2023self} is a study that selects results through voting from multiple instances of CoT, GoT~\cite{besta2023graph} secures flexibility by enabling the generation of graph-like thought nodes, and XoT~\cite{ding2023everything} uses the thought tree while Monte Carlo tree search and refines the tree with reinforcement learning. However, these attempts are closer to additional learning for LLMs, and more research is needed to ascertain whether fundamental improvements in LLMs' reasoning abilities are achievable.
\subsection{Future Direction After Solving ARC}
\label{sec:future_direction}

Solving ARC tasks does not directly imply achieving human-level artificial intelligence. Moreover, there is a challenge in comparing task-solving approaches with those of humans. Thus, we suggest three alternatives to more accurately measure human-level inference abilities.

\subsubsection{Using Different Benchmarks}
One limitation of ARC is its simple environment. SQA3D~\cite{ma2022sqa3d}, for instance, addresses inference tasks in a 3D domain by extending them into question-answering tasks using simulators like ScanNet~\cite{dai2017scannet}. Additionally, benchmarks such as TGIF-QA~\cite{jang2017tgif}, MovieQA~\cite{tapaswi2016movieqa}, TVQA~\cite{lei2018tvqa}, and STAR~\cite{wu2021star}, which append question-answering to videos, have been proposed. Such benchmarks mimicking real-world inference scenarios could serve as supplements to measure complex abstractions not covered by ARC.

\subsubsection{Quantification of ARC Task-Solving Processes}
Chollet, the creator of ARC, argued that ARC maximizes generality while minimizing prior and experience~\cite{chollet2019ARC}, but these components have not been quantitatively evaluated. As a result, the quantitative assessment of factors such as the generality achieved by models solving ARC, the level of prior knowledge, and the components of prior knowledge remains elusive. One possible way to quantitatively evaluate the process of solving ARC tasks is to quantify the model's achievement of prior, experience, and generality.

\subsubsection{Adding Evaluation Methods to Compare Task-Solving Processes with Human Approaches}
Recent ARC research has focused on finding ways for AI to solve tasks. However, there are doubts about how similar these solutions are to those of humans. The initial paper by Johnson et al.~\cite{johnson2021fast} analyzed human ARC solutions. Subsequently, LARC~\cite{acquaviva2022LARC} was proposed to analyze how tasks are solved through the language-based explanation of human solutions. Tools for facilitating the collection of human data are also continuously being developed. Kim et al.~\cite{Kim2022playgrounds}, for instance, have analyzed how tasks are solved through O2ARC. Based on these studies, we suggest calculating each ARC task's correctness and adding similarity with human data to the evaluation.

\iffalse
정리하자면, 1) 사람 수준의 문제 풀이 능력을 ARC로 검증하기 어렵다는 점, 2) 사람과 유사한 방식으로 문제를 푸는지 비교하기 어렵다는 점에서, ARC를 풀었다는 것이 곧바로 사람 수준의 인공지능 개발을 의미하진 않을 것이다. 아래에선 사람 수준의 인공지능 개발을 위해 ARC 벤치마크를 개선하는 다른 방법들을 제안한다.

-	ARC 문제의 변형
1)	다른 벤치마크 사용
ARC 문제의 한계 중 하나는 복잡한 환경에서의 abstraction을 정확히 측정하지 못한다는 데 있다. SQA3D는 3차원 도메인에서의 추론 문제로, ScanNet과 비슷한 3차원 시뮬레이터를 추론이 가능하도록 question-answering task로 확장한 것이 특징이다. 실제 세계 비디오에 Question&Answering을 덧붙인 TGIF-QA, MovieQA, TVQA, STAR 벤치마크 등이 제안되기도 했다. 위와 같은 실생활을 모사한 추론 벤치마크는 ARC가 포함하지 않는 복잡한 abstraction을 측정하는 척도로 보조할 수 있을 것으로 보인다.

2)	Generality, Prior 등을 수치적으로 계산할 필요가 있다. (제거?)
ARC를 제안한 숄레는 ARC가 Generality를 최대화하고, Prior와 Experience를 최소화한 문제라고 주장했다. 그러나 각 구성 요소는 수치적으로 평가된 바 없다. 그 탓에 ARC를 푼 모델의 Generality가 어느 정도고, Prior는 어느 정도인지, 또 Prior는 어떤 요소를 가지는지 등 모델이 ARC를 풀어가는 세부 상황을 정량적으로 평가할 수 없는 실정이다.
ARC 풀이 과정을 정량적으로 평가할 한 가지 가능한 방법은 모델이 성취한 prior, experience, generality를 수치화하는 방법이다. (조금 더 구체적으로 어떻게 수치화 할지 예시 추가)

3)	사람이 문제 푸는 과정과 비교해 문제를 푸는 과정을 평가하는 방법을 추가한다.

최근의 ARC 연구는 인공지능이 풀 수 있는 방법을 찾아내는 방향에 집중하고 있다. 그러나, 그 풀이가 사람과 얼마나 비슷한지에 관해선 의문이 남는다. Fast and Flexible은 사람의 ARC 풀이를 분석한 초기 논문이다. 이후, 어떤 과정을 통해 푸는지 분석하기 위해 사람들이 자신의 풀이를 언어로 설명하게 한 LARC가 제안됐다. 사람 데이터 수집을 원활히 하기 위한 도구 역시 지속적으로 개발되고 있다. Sundong et al.은 O2ARC를 통해 어떻게 문제를 푸는지 분석한 바 있다. ARC의 각 문제에 대해 단순히 맞고 틀리고를 계산하는 것뿐 아니라, 이러한 사람 데이터와의 유사도를 추가할 것을 제안한다.
\fi
\subsection{Recent Research Trends on the Reasoning Abilities of LLMs}
\label{sec:recent_research_trends}

\revise{In this paper, we utilized the ARC to evaluate and enhance the reasoning capabilities of LLMs. 
ARC serves as a crucial benchmark for testing AI models' ability to perform human-like reasoning. 
Beyond ARC, datasets such as DROP~\cite{dua2019drop}, CommonsenseQA~\cite{talmor2019commonsenseqa}, BoolQ~\cite{clark2019boolq}, and GSM8K~\cite{cobbe2021gsm8k} provide invaluable resources to enhance the diverse reasoning capabilities of LLMs.}

\revise{Recent studies indicate that LLMs still exhibit significant limitations in their reasoning abilities despite their proficiency in language-based tasks. LLMs still exhibit significant limitations in their reasoning abilities. Carvalho et al.~\cite{de2024show} found that LLMs struggle with reasoning and decision-making in tasks beyond their training data, particularly in non-linguistic tasks requiring strategic thinking and spatial reasoning. Similarly, Gendron et al.~\cite{gendron2024large} revealed poor performance on tasks requiring the identification and application of general patterns from limited examples. These studies collectively highlight that current LLMs, though advanced in linguistic tasks, are still far from achieving robust reasoning abilities across diverse domains.}

\revise{To address these limitations, several advanced approaches have been developed. 
These include reinforcement learning with human feedback~\cite{christiano2017rlpreference}, CoT prompting~\cite{wei2022chain}, reasoning-centric fine-tuning~\cite{lewkowycz2022reasoning}, incorporating knowledge graphs during pre-training~\cite{liu2020kbert}, and explainable AI techniques~\cite{biran2017explanation}. 
These approaches play a crucial role in advancing LLMs' reasoning capabilities across various domains.}

\revise{Moreover, recent research has introduced innovative approaches to further augment the reasoning capabilities of LLMs. These include multimodal learning techniques~\cite{singh2022flava}, adaptive learning strategies with human feedback~\cite{ouyang2022training}, and integration of programming languages with LLMs~\cite{gao2023pal}. 
These cutting-edge studies significantly contribute to systematically strengthening the multidimensional reasoning capabilities of LLMs.}

\section{Conclusions}
\label{sec: conclusion}

\revise{This study addresses the limitations of result-oriented analysis in LLMs' reasoning abilities by adopting the Language of Thought Hypothesis (LoTH). While recent LLMs have shown performance levels close to humans, experiments reveal significant gaps in planning and reasoning. Through LoTH's three components--logical coherence, compositionality, and productivity--we provide a structured approach to evaluate the reasoning process rather than just outcomes.}

\revise{Using the Abstraction and Reasoning Corpus (ARC) as our benchmark, we conducted three quantitative experiments:}
\begin{enumerate}
    \item \textbf{Logical Coherence:} \revise{Our analysis revealed significant gaps in both inferential and semantic coherence. While LLMs occasionally produced correct answers, they frequently failed to maintain logical consistency across similar problems and often derived correct results through flawed reasoning processes.}
    
    \item \textbf{Compositionality:} \revise{LLMs showed fundamental limitations in combining simple components to solve complex problems. Their performance degraded significantly with increasing task complexity, and they struggled with DSL selection even when provided with additional context, indicating weak compositional abilities.}
    
    \item \textbf{Productivity:} \revise{LLMs demonstrated significant weaknesses in a rule-based generation despite their known capabilities in creative tasks. They often resorted to mimicking observed patterns rather than truly understanding and applying abstract rules to generate valid new examples.}
\end{enumerate}

\revise{These findings suggest that current LLMs, despite their impressive performance metrics, lack fundamental reasoning capabilities when evaluated from a process-oriented perspective. To advance toward human-level artificial intelligence, future research should pursue three complementary directions. First, LLMs need enhancement in both abstraction knowledge and reasoning capabilities: this could involve developing better representation methods for implicit information extraction and exploring advanced prompting techniques to handle vast search spaces efficiently. Second, to ensure meaningful progress, we need to develop more comprehensive evaluation frameworks that can: (1) incorporate diverse benchmarks that better reflect real-world reasoning scenarios; (2) quantitatively measure solution processes beyond mere task completion; and (3) enable systematic comparisons between AI and human reasoning approaches. This study ultimately contributes to the field by providing a structured framework for evaluating and advancing AI reasoning capabilities, highlighting the importance of aligning AI development with human cognitive processes.}

% \revise{These findings suggest that current LLMs, despite their impressive performance metrics, lack fundamental reasoning capabilities when evaluated from a process-oriented perspective. To advance toward human-level artificial intelligence, future research should pursue three complementary directions. First, LLMs need enhancement in both abstraction knowledge and reasoning capabilities: this could involve developing better representation methods for implicit information extraction and exploring advanced prompting techniques to handle vast search spaces efficiently. Second, while ARC serves as a valuable benchmark for reasoning capabilities, we must acknowledge its limitations—it operates in a relatively simple environment and may not fully reflect the complexity of real-world problems that require various cognitive abilities beyond pure reasoning. Finally, to ensure meaningful progress, we need to develop more comprehensive evaluation frameworks that can: (1) incorporate diverse benchmarks that better reflect real-world reasoning scenarios, (2) quantitatively measure solution processes beyond mere task completion, and (3) enable systematic comparisons between AI and human reasoning approaches. This study ultimately contributes to the field by providing a structured framework for evaluating and advancing AI reasoning capabilities, while highlighting the importance of aligning AI development with human cognitive processes.}

\begin{acks}
This work was supported by the IITP (RS-2023-00216011, RS-2024-00445087, No. 2019-0-01842) and the NRF (RS-2024-00451162) grants funded by the Ministry of Science and ICT, Korea. Experiments are supported by the Accelerate Foundation Models Research Initiative, Microsoft. 
% Experiments are done with Azure OpenAI API, supported by the Accelerate Foundation Models Research Initiative, Microsoft.
\end{acks}

\bibliographystyle{ACM-Reference-Format}
\bibliography{ARC_reference}

\appendix
\newpage

\section{Supplementary Analysis}
\subsection{Comparing LLM and Human Problem Difficulty Perception}
\begin{figure}[ht]
    \includegraphics[width=\textwidth]{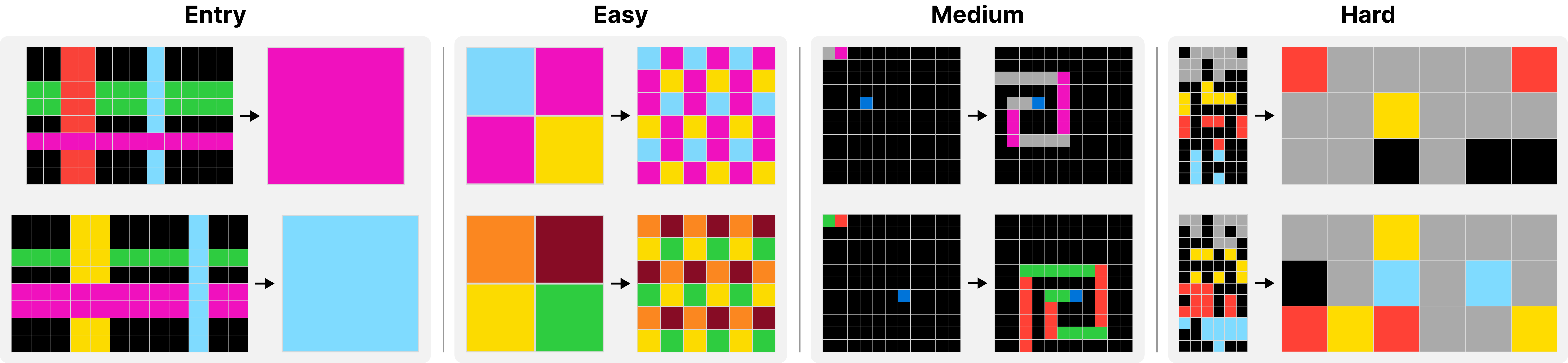}
    \captionsetup{width=.95\linewidth}
    \caption{Showcases of ARC tasks organized by human-perceived difficulty levels. These tasks illustrate the spectrum of complexity that humans use to rate problems, ranging from single-step `Entry' level tasks to multi-step `Hard' challenges. The difficulty classification reflects both the depth of inference required and the number of logical operations needed to reach a solution, paralleling the varying success rates of LLMs in tackling these tasks.}
    % \caption{Representative ARC tasks categorized by difficulty~\cite{borsky2021arcgame}. As in these examples, ARC tasks have differences in difficulty perceived by humans, and ARC tasks are classified according to the level of difficulty based on human standards.}
    \label{fig:logical_coherence/ARC-Game_example}
\end{figure}

Following the analysis in Section~\ref{sec:logical_coherence/semantic_coherence}, we analyzed problems that LLMs (Large Language Models) solve well and those they struggle with. Table~\ref{tbl:logical_coherence/category_performance} presents the accuracy of LLMs across problem difficulty levels classified by humans. The classification was based on the existing categorization, relying on perceived difficulty by humans~\cite{borsky2021arcgame}. As a result, we discovered a tendency where problems perceived as difficult by humans align closely with those challenging for LLMs. Difficult problems shared two commonalities: 1) they required lengthy inference processes to solve, and 2) they involved considering multiple simultaneous problems to extract information about changes. An example from Fig.~\ref{fig:logical_coherence/ARC-Game_example} illustrates this point: a task classified as `Entry' only requires a single step of coloring, while a task classified as `Hard' requires three steps: recognizing each object, identifying the priority of each object, and merging each object considering their priority. `Easy' and `Medium' are tasks that require relatively more complex steps than `Entry' and fewer steps than `Hard'. Considering these observations, it can be inferred that artificial intelligence possesses simple forms of visual logic that deal with only one of the four priors included in ARC: objectness, goal-directedness, numbers and counting, and basic geometry. However, it cannot handle complex combinations of logic that integrate these priors.

\begin{table}[ht]
\small
\centering
\caption{Analyzing LLMs' reasoning capabilities by task difficulty, following prior categorization~\cite{borsky2021arcgame}. The number of ARC tasks corresponding to each category is listed in the table, and the experiment was performed five times for each task.}
\begin{tabular}{c c c c c c c c} 
\toprule
\textbf{} & \textbf{Entry} & \textbf{Easy} & \textbf{Medium} & \textbf{Hard} \\
\midrule
Tasks & 2 & 20 & 46 & 14 \\
Trials & 10 & 100 & 230 & 70 \\
\midrule
CoT & 100.00\% & 30.00\% & 0.00\% & 0.00\% \\
LtM &  20.00\% & 19.00\% & 0.00\% & 2.85\% \\
ToT &  50.00\% & 22.00\% & 0.00\% & 0.00\% \\
\midrule
Average & 56.67\% & 23.67\% & 0.00\% & 0.95\% \\
\bottomrule
\end{tabular}
\label{tbl:logical_coherence/category_performance}
\end{table}
\subsection{Comparison of Augmentation Cost-Efficiency across GPT Versions}

In a follow-up experiment to our productivity study, we aimed to compare the cost-efficiency of GPT-3.5 and GPT4-32k when augmenting demonstration example tasks. This investigation was crucial to understanding the trade-offs between model performance and associated costs in real-world applications.

Our experimental setup began with the creation of a prompt describing the category. Using this prompt, we developed an Inverse Transformation Prompt (ITP) and proceeded to augment demonstration examples using both GPT-3.5-16k and GPT-4-32k models. Throughout this process, we meticulously logged all prompts given to the LLMs and their corresponding responses.

To analyze the cost implications, we tokenized the logged text using the tiktoken library. We then calculated the cost of generating a single valid demonstration example based on the per-token cost specified by the Azure OpenAI API. This approach allowed us to accurately assess the financial implications of using each model for demonstration example augmentation. Validation of the generated examples was a critical component of our experiment. We employed human reviewers to manually verify the quality and appropriateness of the outputs. These reviewers were tasked with confirming two key aspects:

\begin{enumerate}
\item Whether the results could be legitimately generated from the given rules.
\item If the generated results were unique, avoid repetition or trivial variations.
\end{enumerate}

This rigorous validation process ensured that our assessment of ``valid'' examples was thorough and meaningful in the context of practical applications.

% In a follow-up experiment to the productivity study, we compared the cost-efficiency of GPT-3.5 and GPT4-32k when augmenting example tasks. First, as shown in Table~\ref{tbl:appendix_productivity_category_prompt}, a prompt was created to describe the category. Using this, an ITP was made, and examples were augmented using GPT-3.5-16k and GPT-4-32k. During this process, the prompts given to the LLMs and the LLMs' answers were logged. The logged text was then tokenized using the tiktoken library, and the cost per token as specified by the Azure OpenAI API was used to calculate the cost of generating a single valid example. Validation was performed manually by human reviewers. They verified whether the results could be generated from the rules and if these results were unique.

% 먼저 Table~\ref{tbl:appendix_productivity_itp_prompt}에 보이는 것처럼 category에 대해 설명하는 prompt를 만들어 주었다. 이를 이용하여 ITP Prompt를 만들었고 이를 GPT-3.5-16k와 GPT-4-32k를 이용하여 task당 example를 2개씩 증강시켰다. 이때 LLMs에게 준 prompt와 LLMs의 answer를 logging하였다. tiktoken라이브러리를 활용하여 logged된 텍스트에 대해서 tokenizing을 한 후 azure openai api에 나와있는 token당 발생하는 비용을 계삲 이용하여 유효한 example을 1개 생성하는데 드는 비용을 계산하였다. 

\begin{table}[ht]
\small
\centering
\caption{Comparison of augmentation cost-efficiency across GPT versions. The experiment was conducted on each demonstration example pair within the 16 task categories of Concept ARC. This table shows the results of LLMs generating valid demonstration example pairs and costs.}
\begin{tabular}{l c c c c} 
\toprule
 & \textbf{Generated Example} & \textbf{Valid Example} & \textbf{Validity} & \textbf{Cost per Valid Example} \\
\midrule
\multirow{1}{*}{\textbf{GPT-3.5-16k}} & 346 & 24 & 6.94\% & \$0.0275 \\
\midrule
\multirow{1}{*}{\textbf{GPT-4-32k}} & 411 & 40 & 9.73\% & \$0.3925  \\
\bottomrule
\end{tabular}
\label{tbl:appendix_productivity_cost}
\end{table}

Analysis of the cost to generate valid demonstration examples, as illustrated in Table~\ref{tbl:appendix_productivity_cost}, reveals that while GPT-4-32k showed approximately 1.5 times higher performance in terms of validity compared to GPT-3.5-16k, its cost was nearly 20 times higher. This suggests that productivity gains may not scale linearly with model capability and cost, especially when generating outputs under complex constraints. Consequently, in scenarios requiring valid outputs under intricate constraints, GPT-3.5 might be preferable to GPT4-32k when considering the trade-off between performance improvement and cost increase. However, the low overall validity rate of less than 10\% for both models indicates that current LLMs still have significantly lower productivity compared to humans in such tasks. This finding suggests that merely upgrading to more advanced models is unlikely to fully resolve the productivity gap, highlighting the need for further research and development in enhancing LLM performance for complex, constrained tasks.

% Table~\ref{tbl:appendix_productivity_cost} illustrates the cost incurred to generate one valid demonstration example\footnote{GPT-3.5-16k (input cost: $0.5/1M$ tokens, output cost: $1.5/1$M tokens), GPT-4-32k (input cost: $10/1$M tokens, output cost: $30/1$M tokens)}. The results show that while GPT-4-32k demonstrated approximately 1.5 times higher performance in terms of validity, its cost was nearly 20 times higher. This suggests that when generating under complex constraints, productivity may not increase linearly with model capability and cost. Furthermore, these findings indicate that in situations where valid outputs need to be generated under complex constraints, using GPT-3.5 might be preferable to using GPT4-32k, considering the trade-off between performance improvement and cost increase. However, the low validity of less than 10\% suggests that LLMs still have lower productivity compared to humans, and that merely changing the model is unlikely to resolve this issue.
% \input{contents/6.6_human_analysis_tool}
\subsection{Limitations of ARC as a Benchmark for Human-Level AI}
\label{sec:discussion/limitation_of_arc}

Does solving ARC signify the completion of human-like AI? To answer this question, two doubts need to be appropriately addressed: 1) Will the ARC solver possess human-level problem-solving abilities? and 2) Will that solver think like humans to solve ARC? It is not easy to imagine how the ARC solver operates without human-level reasoning. At this point, what we can assume is that the model will have the three properties of LoTH, and the model could be capable of several types of reasoning included in ARC. With this hypothesis, we attempt to address the following questions.

\subsubsection{Will the Model Possess Human-Level Problem-Solving Abilities?}

Being capable of reasoning does not necessarily equate to having human-level problem-solving abilities. In other words, even if a model can reason to a level that can solve ARC, it may not have human-level problem-solving capabilities. Various tasks that humans face are generally more complex than ARC and involve various other cognitive factors besides reasoning. Therefore, even models that can solve ARC may have the following limitations compared to human-level problem-solving abilities.

First, with the current ARC criteria, it is still unknown whether the model that solved it can solve more complex types of tasks. This is because ARC tasks focus on just reasoning and are therefore presented in a relatively simple environment. Whether the reasoning ability learned through ARC would also work in more complex environments has not been revealed. Second, solving ARC does not imply the presence of other components of intelligence beyond reasoning. While reasoning is undoubtedly a core aspect of cognitive processes, it is not the entirety of intelligence. There is research showing that solving human-level complex tasks requires various cognitive abilities~\cite{gardner2011frames}.

\subsubsection{Will the Model Think Like Humans?}

Even if we assume that the ARC solver can reason in terms of LoTH, we cannot guarantee whether this solver's process is human-like for the following two reasons. Firstly, the current ARC provides a performance measure that rewards only for solving a task. It is important to recognize that such a measure might instigate the wrong purpose, leading to what is known as the King Midas problem~\cite{russell1995artificial}. This problem emphasizes the risk of AI achieving its given objective too literally, leading to unintended negative consequences, underscoring the importance of aligning AI's goals with human values and the broader context. The policy of rewarding only the results, excluding the solution process, makes it difficult to evaluate whether the solution process is similar to human reasoning. Therefore, models trained on current ARC likely differ in how they solve tasks compared to humans. The second reason is that directly comparing the reasoning processes of humans and language models is challenging. The process by which humans solve ARC tasks has not been investigated, making it unclear how the solving process differs between humans and artificial intelligence. Furthermore, there is a lack of metrics for comparing the solving processes, making direct comparisons difficult.
\\

\section{Experimental Detail}
\subsection{Logical Coherence}
\label{appendix:Logical Coherence}

The logical coherence study comprised two main experiments: a comparison on semantic coherence across prompting techniques and an assessment of the inferential coherence of LLMs. For the first experiment, prompting technique comparison, we randomly selected 100 tasks from the ARC evaluation set. We then applied three different prompting methods - Chain of Thought (CoT), Least to Most (LtM), and Tree of Thoughts (ToT) - to compare their effectiveness in maintaining semantic coherence.
% Logical Coherence 실험은 Comparison Across Prompting Techniques과 Inferential Coherence of LLMs로 구성이 되었다. Comparison Across Prompting Techniques 실험의 경우 ARC evaluation dataset에서 100개의 task를 random으로 뽑은 후 Chain of Thought(CoT), Least to Most(LtM), Tree of Thoughts(ToT)을 적용하여 semantic coherence 성능을 비교한 실험이다. 

The second experiment assessing the inferential coherence of the LLMs aims to assess whether the same logic can be consistently applied. Therefore, it is necessary to first confirm the tasks where the LLMs have understood the logic. To this end, we experimented using CoT prompting, which showed the best performance in the comparison across prompting techniques experiment, to solve the ARC training set. This experiment was repeated five times. The inferential coherence of the LLMs experiment was then conducted on tasks that were correctly solved at least once out of the five repetitions. The detailed task IDs and prompts used in each experiment are provided in \ref{sec:appendix_logical_coherence_section_task_id} and \ref{sec:appendix_logical_coherence_section_prompting_setting}, respectively.
% Inferential Coherence of LLMs 실험은 동일한 logic을 일관적으로 적요시킬 수 있는지 보는 것이기 때문에 먼저 LLM이 logic을 이해한 tasks에 대해서 확인해야 한다. 이를 위해 Compariosn Across Prompting Techniques 실험에서 성능이 가장 좋았던 CoT promting을 사용하여 ARC training dataset을 solving하는 실험을 진행하였고 이때 5번 반복하였다. 그리고 5번의 반복 중 최소 1번이상 맞춘 tasks를 대상으로 Inferential Coherence of LLMs 실험을 진행하였다. 각 실험에서 사용한 자세한 task id와 프롬프트는 각각 \ref{sec:appendix_logical_coherence_section_task_id}와 ~\ref{sec:appendix_logical_coherence_section_prompting_setting}에 자세히 나와있다.

\subsubsection{Task ID List for Each Experiment}
\label{sec:appendix_logical_coherence_section_task_id}

Task ID list selected for the experiment comparing logical coherence. The first experiment for comparison across prompting techniques was conducted on 100 ARC evaluation tasks, while the second experiment for the inferential coherence experiment of LLMs was carried out on 83 ARC training tasks.

\begin{tcolorbox}[breakable, colback=blue!5!white, colframe=blue!75!black, title=Task ID List: Comparison Across Prompting Techniques]
`3ed85e70', `0f63c0b9', `17cae0c1', `47996f11', `4acc7107', `0692e18c', `477d2879', `1c0d0a4b', `292dd178', `1990f7a8', `22a4bbc2', `4364c1c4', `2f0c5170', `17b80ad2', `03560426', `0c786b71', `3391f8c0', `42a15761', `0bb8deee', `1e97544e', `1c02dbbe', `4b6b68e5', `2a5f8217', `3194b014', `1acc24af', `0c9aba6e', `0e671a1a', `37d3e8b2', `0becf7df', `0607ce86', `3a301edc', `2546ccf6', `009d5c81', `31adaf00', `281123b4', `3d31c5b3', `423a55dc', `1d0a4b61', `1a2e2828', `319f2597', `3979b1a8', `12422b43', `140c817e', `0a2355a6', `19bb5feb', `332efdb3', `27a77e38', `2c0b0aff', `00dbd492', `2c737e3', `2072aba6', `48f8583b', `27f8ce4f', `14754a24', `32e9702f', `195ba7dc', `137f0df0', `184a9768', `29700607', `1c56ad9f', `15663ba9', `4c177718', `136b0064', `0a1d4ef5', `1d398264', `09c534e7', `2685904e', `48131b3c', `31d5ba1a', `2697da3f', `103eff5b', `12997ef3', `1e81d6f9', `25094a63', `08573cc6', `20981f0e', `4852f2fa', `2b01abd0', `2072aba6', `1a6449f1', `34b99a2b', `0b17323b', `15696249', `414297c0', `2753e76c', `12eac192', `0934a4d8', `310f3251', `358ba94e', `21f83797', `4aab4007', `351d6448', `45bbe264', `456873bc', `15113be4', `3490cc26', `3b4c2228', `00576224', `42918530', `45737921', `20818e16'
\end{tcolorbox}

\addvspace{10pt}

\begin{tcolorbox}[breakable, colback=blue!5!white, colframe=blue!75!black, title=Task ID List: Inferential Coherence of LLMs]
`017c7c7b', `025d127b', `08ed6ac7', `0dfd9992', `1bfc4729', `22eb0ac0', `239be575', `23b5c85d', `25d8a9c8', `25ff71a9', `272f95fa', `27a28665', `3618c87e', `3af2c5a8', `3bd67248', `3bdb4ada', `44f52bb0', `48d8fb45', `496994bd', `49d1d64f', `539a4f51', `53b68214', `5582e5ca', `5bd6f4ac', `6150a2bd', `62c24649', `67a3c6ac', `67e8384a', `68b16354', `6d0aefbc', `6f8cd79b', `6fa7a44f', `7447852a', `746b3537', `74dd1130', `7837ac64', `794b24be', `7b7f7511', `7f4411dc', `82819916', `88a62173', `8be77c9e', `8e1813be', `90c28cc7', `9172f3a0', `963e52fc', `97999447', `9dfd6313', `a416b8f3', `a65b410d', `a699fb00', `a85d4709', `a87f7484', `aabf363d', `b1948b0a', `b8cdaf2b', `b94a9452', `ba26e723', `ba97ae07', `bc1d5164', `bd4472b8', `bda2d7a6', `bdad9b1f', `c3f564a4', `c59eb873', `c8f0f002', `c9e6f938', `c9f8e694', `d0f5fe59', `d10ecb37', `d13f3404', `d2abd087', `d4469b4b', `d631b094', `d8c310e9', `d9fac9be', `dc433765', `e26a3af2', `e9afcf9a', `ea786f4a', `ef135b50', `f5b8619d', `f76d97a5'
\end{tcolorbox}

\subsubsection{Prompting Setting}
\label{sec:appendix_logical_coherence_section_prompting_setting}

The prompts used in comparison across prompting techniques and inferential coherence of LLMs are CoT, L2M, and ToT. These are detailed in Section~\ref{sec:appendix_logical_coherence_section_detailed_promptings}. In the prompts, parts enclosed in curly brackets indicate where the corresponding type should be inserted. Demo examples and test input refer to the demonstration examples and test input of the test example provided in the task to be solved. For instance, if the type is CoT Prompt, it consists of the CoT one-shot example, the task's demonstration examples, and test input. Regardless of the prompting method, all prompts are given a one-shot example. CoT solves the task using only the one-shot example, the task's demonstration examples, and test input. On the other hand, LtM and ToT use a decomposing prompt to obtain instructions for solving the problem through a decomposing stage. For LtM, the step-by-step solving prompt is then used to sequentially execute the instructions obtained through decomposing. The previously executed instructions and the resulting changed grid are included in this process. For ToT, the decomposing prompt is used to create multiple instruction candidates, and the ToT decomposing vote prompt is used to have the LLM select the most promising instruction candidate. The selected instructions are then processed through the step-by-step solving prompt to generate multiple candidate results for each instruction. The ToT step-by-step solving vote prompt is then used to select the grid that best reflects the instruction. This process is carried out step-by-step for all instructions.

% Comparison Across Prompting Techniques와 Inferential Coherence of LLMs에서 사용한 prompt는 Figure~\ref{fig:logical_coherence/allinone}과 같이 CoT', 'L2M', 'ToT이다. 
% Table~\ref{tbl:appendix_logical_cohernece_prompting_setting_table}에 나와 있는 것과 같다. Prompt에 curly brackets으로 둘러싸인 부분은 해당 Type을 insert한다라는 뜻이며 \{ Demo Examples \}와 \{ Test Input \}은 풀어야 할 task에서 제공되는 demostrration examples와 test example이다. 예를 들어 type이 CoT Prompt인 경우 CoT One-Shot Example과 task의 demostration examples', 'test input 등으로 구성된 prompt이다. 
% 프롬프트 방식과 상관 없이 모든 프롬프트에는 one-shot example이 주어진다. CoT는 One-Shot Example과 task의 demonstration examples과 test input만을 가지고 task를 해결한다. 반면 LtM과 ToT의 경우 decomposing 단계를 통해 문제를 해결하기 위한 instructions을 얻어야 함으로 Decomposing Prompt를 사용한다. 그 후, LtM의 경우 Step-By-Step Solving Prompt를 통해 decomposing을 통해 얻은 instructions을 하나씩 수행해나간다. 이때 이전에 수행한 instructions과 이에 대한 changed grid는  ToT의 경우 Decomposing Prompt를 통해 여러 개의 instructions 후보들을 만들고 ToT Decomposing Vote Prompt를 통해 LLM에게 가장 좋을 것 같은 instructions 후보를 선택하도록 하게 한다. 이렇게 얻어진 instructions을 Step-By-Step Solving Prompt를 통해 instruction을 한 개에 대한 여러 후보 결과들을 만든후에 ToT Step-By-Step Solving Vote Prompt를 통해 instruction이 잘 반영된 grid를 선택한다. 이 과정을 모든 instructions들에 대해서 step-by-step으로 진행한다.

\subsubsection{Detailed Promptings}
\label{sec:appendix_logical_coherence_section_detailed_promptings}

The logical coherence experiments employed various prompting techniques, including CoT, LtM, and ToT. CoT utilizes the CoT prompt, while LtM uses decomposing and step-by-step solving prompts. ToT incorporates decomposing, ToT decomposing vote, step-by-step solving, and ToT step-by-step solving vote prompts.

\begin{tcolorbox}[breakable, colback=gray!5!white, colframe=gray!75!black, title=CoT One-Shot Example Data]
If input grids are like that:  
 
[[0, 3, 0, 0, 0, 0], 

[0, 3, 0, 2, 0, 0], 

[0, 0, 0, 2, 0, 0], 

[0, 8, 0, 0, 2, 2],

[0, 0, 0, 0, 2, 2], 

[6, 6, 6, 0, 0, 0]], 

\addvspace{10pt}
 
then these grids change to output grids below:  
 
[[0, 0, 0, 0, 3, 0], 

[0, 0, 0, 0, 3, 2], 

[0, 0, 0, 0, 0, 2], 

[0, 0, 0, 8, 2, 2], 

[0, 0, 0, 0, 2, 2], 

[0, 0, 0, 6, 6, 6]].
\end{tcolorbox}

\addvspace{10pt}

\begin{tcolorbox}[breakable, colback=gray!5!white, colframe=gray!75!black, title=CoT One-Shot Example]
Do you know the ARC problem?  
 
\addvspace{10pt}
 
It is similar to below.  
 
\addvspace{10pt}
 
\textbf{\{CoT One-Shot Example Data\}}  
 
\addvspace{10pt}
 
You can understand the pattern of this problem with the input-output pair in Example 1. In the above case, you can infer as follows.  
 
\addvspace{10pt}
 
In Example 1, all objects move to the right, and then you can move the object to the right side.  
 
\addvspace{10pt}
 
Like this concept, `object', `count', `color', `move', `row', `column', etc., help you to understand the patterns of the problem and solve it.  
 
\addvspace{10pt}
 
Below is another pattern to solve. So you can understand the pattern with several examples and apply the problem's input to get the correct output.
\end{tcolorbox}

\begin{tcolorbox}[breakable, colback=gray!5!white, colframe=gray!75!black, title=CoT Prompt]
\textbf{\{CoT One-Shot Example\}}  
 
\addvspace{10pt}
 
\textbf{\{Demo Examples\}}  
 
\addvspace{10pt}
 
If input grids are like that:  
 
\addvspace{10pt}
 
\textbf{\{Test Input\}}  
 
\addvspace{10pt}
 
then output grids?
\end{tcolorbox}

\begin{tcolorbox}[breakable, colback=gray!5!white, colframe=gray!75!black, title=Decomposing One-Shot Example-Demo Examples]
Example 1  

\addvspace{10pt}
 
If input grids are like that:  
 
[[0, 0, 0, 0, 0, 0], 

[0, 0, 3, 0, 0, 0], 

[0, 3, 0, 3, 0, 0], 

[0, 0, 3, 0, 3, 0], 

[0, 0, 0, 3, 0, 0], 

[0, 0, 0, 0, 0, 0]]  

\addvspace{10pt}
 
then these grids change to output grids below:  
 
[[0, 0, 0, 0, 0, 0], 

[0, 0, 3, 0, 0, 0], 

[0, 3, 4, 3, 0, 0], 

[0, 0, 3, 4, 3, 0],

[0, 0, 0, 3, 0, 0], 

[0, 0, 0, 0, 0, 0]].  

\addvspace{10pt}
 
Example 2  

\addvspace{10pt}
 
If input grids are like that:  
 
[[0, 0, 0, 0, 0, 0, 0, 0, 0, 0],

[0, 0, 3, 0, 3, 0, 0, 0, 0, 0], 

[0, 0, 0, 3, 0, 3, 0, 0, 0, 0], 

[0, 0, 3, 0, 0, 0, 3, 0, 0, 0], 

[0, 0, 0, 0, 0, 3, 0, 3, 0, 0], 

[0, 0, 0, 3, 0, 3, 3, 0, 0, 0], 

[0, 0, 3, 3, 3, 0, 0, 0, 0, 0], 

[0, 0, 0, 3, 0, 0, 0, 0, 0, 0],

[0, 0, 0, 0, 0, 0, 0, 0, 0, 0],

[0, 0, 0, 0, 0, 0, 0, 0, 0, 0]]  

\addvspace{10pt}
 
then these grids change to output grids below:  
 
[[0, 0, 0, 0, 0, 0, 0, 0, 0, 0],

[0, 0, 3, 0, 3, 0, 0, 0, 0, 0], 

[0, 0, 0, 3, 0, 3, 0, 0, 0, 0], 

[0, 0, 3, 0, 0, 0, 3, 0, 0, 0], 

[0, 0, 0, 0, 0, 3, 4, 3, 0, 0], 

[0, 0, 0, 3, 0, 3, 3, 0, 0, 0], 

[0, 0, 3, 3, 3, 0, 0, 0, 0, 0], 

[0, 0, 0, 3, 0, 0, 0, 0, 0, 0], 

[0, 0, 0, 0, 0, 0, 0, 0, 0, 0], 

[0, 0, 0, 0, 0, 0, 0, 0, 0, 0]].
\end{tcolorbox}

\begin{tcolorbox}[breakable, colback=gray!5!white, colframe=gray!75!black, title=Decomposing One-Shot Example-Test Input]
Example-problem  

\addvspace{10pt}
 
If input grids are like that:  

[[0, 0, 0, 0, 0, 3, 0, 0, 0, 0], 

[0, 0, 0, 0, 3, 0, 0, 0, 0, 0], 

[0, 3, 3, 0, 3, 3, 0, 3, 0, 0], 

[3, 0, 0, 3, 0, 0, 3, 0, 3, 0],

[0, 0, 0, 3, 0, 0, 3, 3, 0, 0], 

[0, 0, 0, 3, 0, 0, 3, 0, 0, 0], 

[0, 0, 0, 3, 0, 0, 3, 0, 0, 0], 

[0, 0, 0, 0, 3, 3, 0, 3, 0, 0], 

[0, 0, 0, 0, 0, 0, 0, 0, 3, 0], 

[0, 0, 0, 0, 0, 0, 0, 0, 0, 0]]  

\addvspace{10pt}
 
then output grids?
\end{tcolorbox}

\begin{tcolorbox}[breakable, colback=gray!5!white, colframe=gray!75!black, title=Decomposing Prompt]
Do you know the ARC problem?  
 
\addvspace{10pt}
 
Each example has the same pattern and the quiz also has the same pattern with examples. So if you understand the pattern of examples, you can solve the quiz. It will help you to analyze the pattern if you decompose the pattern into some steps. I give an example that decomposes patterns into subquestions.  
 
\addvspace{10pt}
 
\textbf{\{Decomposing One-Shot Example-Demo Examples\}}  
 
\addvspace{10pt}
 
\textbf{\{Decomposing One-Shot Example-Test Input\}}  
 
\addvspace{10pt}
 
To solve the quiz, I think we should do something like below:  
 
\addvspace{10pt}
 
Q1: We need to identify the places surrounded by 3s in the input grid of example-quiz.  

Q2: Fill in the 4 in the location we found through Q1.  
 
\addvspace{10pt}
 
\textbf{\{Demo Examples\}}  
 
\addvspace{10pt}
 
\textbf{\{Test Input\}}  
 
\addvspace{10pt}
 
I want you to answer in the format below:  
 
\addvspace{10pt}
 
Q1: ....  

Q2: ....  

...  

QN: ....  
 
\addvspace{10pt}
 
N is the index of the last question.  

\addvspace{10pt}
 
(The answers to the last question should allow you to generate the output grid for the quiz, and you shouldn't solve the problem yet in this process. You should only create the subquestions for solving the problem.)
\end{tcolorbox}

\begin{tcolorbox}[breakable, colback=gray!5!white, colframe=gray!75!black, title=Step-By-Step Solving Prompt]
\textbf{\{CoT One-Shot Example\}}  
 
\addvspace{10pt}
 
\textbf{\{Demo Examples\}}  
 
\addvspace{10pt}
 
If input grids are like that:  
   
\textbf{\{Test Input\}}  
 
\addvspace{10pt}
 
then output grids?  
 
\addvspace{10pt}
 
\textbf{\{Previous Instructions\}}  
 
\addvspace{10pt}
 
\textbf{\{Previous Changed Grid\}}  
 
\addvspace{10pt}
 
\textbf{\{Current Instruction\}}  
\end{tcolorbox}

\begin{tcolorbox}[breakable, colback=gray!5!white, colframe=gray!75!black, title=ToT Decomposing Vote Prompt]
First, consider how to solve the problem below.  
  
\addvspace{10pt}
 
\textbf{\{Demo Examples\}}  
 
\addvspace{10pt}
 
\textbf{\{Test Input\}}  
 
\addvspace{10pt}
 
Then, given instruction and several choices, decide which choice is most promising. Analyze each choice in detail, then conclude in the last line ``The best choice is s", where s is the integer ID of the choice.
\end{tcolorbox}

\begin{tcolorbox}[breakable, colback=gray!5!white, colframe=gray!75!black, title=ToT Step-By-Step Solving Vote Prompt]
First, consider how to solve the problem below.  
  
\addvspace{10pt}
 
\textbf{\{Demo Examples\}}  
 
\addvspace{10pt}
 
\textbf{\{Test Input\}}  
 
\addvspace{10pt}
 
Then, given a question and some answers, you need to choose which answer is the best answer to the question. Analyze each choice in detail and then conclude on the last line, ``The best answer is `s'," where s is the integer ID of the answer.  
 
\addvspace{10pt}
 
\textbf{\{Previous Instructions\}}  
 
\addvspace{10pt}
 
\textbf{\{Previous Changed Grid\}}  
 
\addvspace{10pt}
 
\textbf{\{Current Instruction\}}  

\addvspace{10pt}
 
\textbf{\{Current Changed Grid\}}  
\end{tcolorbox}

% \newcolumntype{P}[1]{>{\centering\arraybackslash}p{#1}}
\newcommand{\tab}{\hspace*{1em}}

% \newpage
\subsection{Compositionality}
\label{appendix:Compositionality}

In the compositionality study, we conducted two experiments: the LLMs DSL understanding experiment to assess how well LLMs comprehend the provided DSLs, and an experiment to evaluate LLMs' compositionality ability. The LLMs DSL understanding experiment measures how accurately LLMs can generate the correct DSLs when given the answer for a task. The compositionality ability experiment examines whether LLMs can correctly select and use the necessary DSLs from those provided for problem-solving. Both experiments used the same set of tasks. Detailed information about the task IDs can be found in Table~\ref{sec:appendix_compositionality_task_id_list}, while the specific prompt details are available in Table~\ref{sec:appendix_compositionality_section_detailed_dsl_promptings} and Table~\ref{sec:appendix_compositionality_section_detailed_promptings}.

% Compositionality에서는 LLM이 제공된 DSL에 대해서 얼마나 이해하고 있는지 확인하는 LLMs DSL understanding 실험과 LLM의 compositionality ability를 확인하는 실험을 진행하였다. LLMs DSL understanding 실험의 경우, task에 해당하는 정답 DSLs을 주고 LLM이 정확하게 생성하는지 측정하는 실험이다. compositionality ability를 확인하는 실험은 LLM이 problem-solving을 위해서 제공된 dsls중 필요한 dsls들을 올바르게 선택하여 사용하는지 확인하는 실험이다. 이때 두 실험에서 사용한 tasks는 모두 동일하며 task id에 대한 자세한 내용은 Table~\ref{tbl:appendix_compositionality_task_id_list}를, Prompt 세부 내용은 Table~\ref{tbl:appendix_compositionality_dsl_function_prompt}과 Table~\ref{tbl:appendix_compositionality_prompt_setting}를 확인하면 된다.

\subsubsection{Task ID List}
\label{sec:appendix_compositionality_task_id_list}
Task ID list for the compositionality experiment comprising 158 tasks. From the total 800 ARC tasks, we selected only those problems where the input and output grid sizes were identical and could be solved with DSL sequence length within 10 using the given DSL for our experiment.

\begin{tcolorbox}[breakable, colback=blue!5!white, colframe=blue!75!black, title=Task ID List:  LLMs DSL Understanding \& Compositionality of LLMs]
`025d127b', `05f2a901', `08ed6ac7', `0ca9ddb6', `0d3d703e', `11852cab', `150deff5', `1b60fb0c', `1caeab9d', `1e0a9b12', `1f642eb9', `1f876c06', `2204b7a8', `22168020', `22233c11', `228f6490', `22eb0ac0', `253bf280', `25d487eb', `25d8a9c8', `25ff71a9', `29c11459', `2bee17df', `2c608aff', `2dd70a9a', `3345333e', `3618c87e', `36fdfd69', `3906de3d', `3aa6fb7a', `3bd67248', `3c9b0459', `3e980e27', `3eda0437', `40853293', `42a50994', `444801d8', `44d8ac46', `4938f0c2', `496994bd', `50846271', `508bd3b6', `50cb2852', `5168d44c', `54d82841', `5582e5ca', `56dc2b01', `56ff96f3', `5c0a986e', `60b61512', `6150a2bd', `623ea044', `63613498', `67385a82', `673ef223', `67a3c6ac', `67a423a3', `6855a6e4', `68b16354', `694f12f3', `6c434453', `6cf79266', `6d58a25d', `6d75e8bb', `6e02f1e3', `6e19193c', `6e82a1ae', `74dd1130', `760b3cac', `776ffc46', `794b24be', `7ddcd7ec', `7e0986d6', `7f4411dc', `810b9b61', `855e0971', `88a10436', `890034e9', `8d510a79', `90f3ed37', `928ad970', `93b581b8', `941d9a10', `952a094c', `9565186b', `99fa7670', `9dfd6313', `a2fd1cf0', `a3df8b1e', `a48eeaf7', `a5313dff', `a61f2674', `a65b410d', `a699fb00', `a78176bb', `a79310a0', `a85d4709', `a9f96cdd', `aabf363d', `ae3edfdc', `aedd82e4', `af902bf9', `b1948b0a', `b230c067', `b2862040', `b548a754', `b7249182', `b8cdaf2b', `ba97ae07', `bb43febb', `bda2d7a6', `bdad9b1f', `c0f76784', `c8f0f002', `cbded52d', `ce22a75a', `ce9e57f2', `d037b0a7', `d07ae81c', `d23f8c26', `d2abd087', `d406998b', `d43fd935', `d4a91cb9', `d4f3cd78', `d511f180', `d5d6de2d', `d6ad076f', `d89b689b', `d8c310e9', `d90796e8', `d9f24cd1', `dc433765', `ddf7fa4f', `ded97339', `e40b9e2f', `e48d4e1a', `e5062a87', `e509e548', `e73095fd', `e8dc4411', `e9614598', `e9afcf9a', `ea32f347', `ea786f4a', `ec883f72', `ed36ccf7', `ef135b50', `f25ffba3', `f76d97a5', `f8a8fe49', `fcc82909', `8f2ea7aa', `5521c0d9', `32597951', `98cf29f8', `0e206a2e', `a1570a43'
\end{tcolorbox}

\subsubsection{Types of DSLs used}
Each DSL was implemented as a Python function. As shown in Table~\ref{tbl:appendix_compositionality_dsl_list_table}, there are three types of DSLs using three parameter types. Color Change DSLs accept parameters such as Coordinate and Object. Coordinate-based Color Change DSLs include Pixel Color, X Line, Horizontal Line, Vertical Line, and Diagonal Line. For Object parameters, only the obj color DSL exists. Transformation DSLs use Object and Grid parameters. Object-based transformations include Rotate Left Obj, Rotate Right Obj, Horizontal Flip Obj, Vertical Flip Obj, and movement operations (Move Left, Move Right, Move Up, Move Down). Grid-based transformations include Rotate Left State, Rotate Right State, Horizontal Flip, and Vertical Flip. Lastly, the Complete DSL exists independently of parameters, indicating task completion before reaching DSL sequence length of 10. For tasks solved with exactly DSL sequence length of 10, the Complete DSL is unnecessary.

\begin{table}[h]
\small
\centering
\caption{DSL list: The DSL used in the compositionality experiment is categorized based on the type of target and the kind of functionality. The targets in the DSL are categorized into three types: Coordinate, Object, and Grid. The functions of the DSL include changing the color of a target (Color Change), moving a target (Transformation), and indicating the completion of the task at DSL sequence length shorter than 10 (Complete).}
\resizebox{\textwidth}{!}{
\begin{tabular}{c c c c}
\toprule
& \textbf{Coordinate} & \textbf{Object} & \textbf{Grid} \\
\midrule
\textbf{Color Change} & 
\begin{tabular}[c]{@{}c@{}}Pixel Color, X Line,\\ Horizontal Line, Vertical Line,\\ Diagonal Line\end{tabular} & 
Obj Color & 
X \\
\midrule
\multirow{2}{*}{\textbf{Transformation}} & 
X & 
\begin{tabular}[c]{@{}c@{}}Rotate Left Obj, Rotate Right Obj,\\ Horizontal Flip Obj,\\ Vertical Flip Obj,\\ Move Left, Move Right,\\ Move Up, Move Down\end{tabular} & 
\begin{tabular}[c]{@{}c@{}}Rotate Left State,\\ Rotate Right State,\\ Horizontal Flip, Vertical Flip\end{tabular} \\
\midrule
\multicolumn{1}{c}{\textbf{Complete}} & \multicolumn{3}{c}{Complete}  \\
\bottomrule
\end{tabular}
}
\label{tbl:appendix_compositionality_dsl_list_table}
\end{table}

% \newpage
% \subsection{Prompting Setting}
\subsubsection{Prompt Contents for LLMs with DSL Codes and Comments }

In the two experiments measuring compositionality and LLM's DSL understanding, we identified a set of 10 tasks that collectively required the use of all 15 DSLs at least once. This set was used to determine the optimal prompt for explaining DSLs to LLMs. We conducted experiments with four prompt variants: no DSL information, DSL code only, DSL comments only, and both DSL code and comments. The LLMs DSL understanding experiment was performed for these 10 tasks across all four prompt compositions. Results indicated that providing both code and comments yielded optimal performance. Consequently, we employed prompts containing both DSL code and comments for the LLM's DSL understanding and compositionality of LLMs experiments. Section~\ref{sec:appendix_compositionality_section_detailed_dsl_promptings} illustrates the prompt content where both code and comments were provided to the LLM.

% Compositionality와 LLM의 dsl understanding을 측정한 본문의 두 실험에서, 우리는 LLM에게 DSLs 설명하는 최적의 프롬프트를 찾고자 15개의 DSL을 최소 한 번씩은 사용해야 하는 task의 집합을 찾았다. 이 집합은 10개의 task로 구성이 되어 있으면 10개의 task에 대해서 모두 맞출려면 15개의 DSLs를 최소 1번은 사용해야 한다. 10개의 task를 이용하여 최적의 프롬프트의 구성을 찾는 실험을 진행하였다. 프롬프트 구성은 DSL에 대한 정보를 주지 않는 경우, DSLs 코드만 준 경우, DSLs의 주석만 준 경우, DSLs의 코드와 주석 모두 준 경우 등 총 4가지이다. 이 4가지 프롬프트 구성에 대해서 10개의 LLMs DSL understanding 실험을 진행해보았다. 결과적으로, 코드와 주석 모두를 준 경우가 성능이 가장 좋았다. 이러한 이유로 LLMs DSL understainding과 Compositionality of LLMs실험에서 프롬프트에 DSLs 코드와 주석을 함께 제공하는 프롬프틀 사용였다. Table~\ref{tbl:appendix_compositionality_dsl_function_prompt}은 LLM에게 코드와 주석을 모둔 준 경우에 해당하는 프롬프트 내용이다. 

\lstset{
    basicstyle=\normalfont,
    keywordstyle=\bfseries,
    columns=fullflexible,
    keepspaces=true
}

\subsubsection{Detailed DSL Promptings}
\label{sec:appendix_compositionality_section_detailed_dsl_promptings}

Prompts of DSL Function Codes and Comments

\begin{tcolorbox}[breakable, colback=gray!5!white, colframe=gray!75!black, title=DSL Prompt: Rotate Left State]
\# rotate\_left\_state function is a counterclockwise rotation about the given state. 

\# This function rotates a square grid (NxN) counterclockwise by 90 degrees. 

\# Parameters: 

\# - state: A 2D list representing the current grid state. 

\# Returns: 

\# - A new 2D list representing the grid after the counterclockwise rotation.  

\addvspace{10pt}

\begin{lstlisting}
def rotate_left_state(state):
    N = len(state)
    rotated_state = copy.deepcopy(state)
    if N == len(state[0]):
        temp_state = copy.deepcopy(state)
        for x in range(N):
            for y in range(N):
                rotated_state[N-1-y][x] = state[x][y]
    return rotated_state
\end{lstlisting}
\end{tcolorbox}

\begin{tcolorbox}[breakable, colback=gray!5!white, colframe=gray!75!black, title=DSL Prompt: Rotate Right State]
\# rotate\_right\_state function is a clockwise rotation about the given state. 

\# This function rotates a square grid (NxN) clockwise by 90 degrees. 

\# Parameters: 

\# - state: A 2D list representing the current grid state. 

\# Returns: 

\# - A new 2D list representing the grid after the clockwise rotation.  

\addvspace{10pt}

\begin{lstlisting}
def rotate_right_state(state):
    N = len(state)
    rotated_state = copy.deepcopy(state)
    if N == len(state[0]):
        for x in range(N):
            for y in range(N):
                rotated_state[y][N-1-x] = state[x][y]
    return rotated_state
\end{lstlisting}
\end{tcolorbox}

\begin{tcolorbox}[breakable, colback=gray!5!white, colframe=gray!75!black, title=DSL Prompt: Vertical Flip]
\# vertical\_flip function is a flip by x-axis about the given state. 

\# This function flips the grid vertically, swapping the top and bottom rows. 

\# Parameters: 

\# - state: A 2D list representing the current grid state. 

\# Returns: 

\# - A new 2D list representing the grid after the vertical flip.  

\addvspace{10pt}

\begin{lstlisting}
def vertical_flip(state):
    temp_state = copy.deepcopy(state)
    N = len(state)
    M = len(state[0])
    for i in range(N):
        for j in range(M):
            temp_state[N-1-i][j] = state[i][j]
    return temp_state
\end{lstlisting}
\end{tcolorbox}

\begin{tcolorbox}[breakable, colback=gray!5!white, colframe=gray!75!black, title=DSL Prompt: Horizontal Flip]
\# horizontal\_flip function is a flip by y-axis about the given state. 

\# This function flips the grid horizontally, swapping the left and right columns. 

\# Parameters: 

\# - state: A 2D list representing the current grid state. 

\# Returns: 

\# - A new 2D list representing the grid after the horizontal flip.  

\addvspace{10pt}

\begin{lstlisting}
def horizontal_flip(state):
    N = len(state)
    M = len(state[0])
    flipped_state = copy.deepcopy(state)
    for i in range(N):
        for j in range(M // 2):
            flipped_state[i][j], flipped_state[i][M-1-j] = state[i][M-1-j], state[i][j]
    return flipped_state
\end{lstlisting}
\end{tcolorbox}

\begin{tcolorbox}[breakable, colback=gray!5!white, colframe=gray!75!black, title=DSL Prompt: Move Right]
\# move\_right function moves all pixels in the selected object to the right by one column. 

\# Parameters: 

\# - state: A 2D list representing the current grid state. 

\# - object: A list of lists where each inner list contains the coordinates [x, y] of a pixel to move. 

\# Returns: 

\# - A new 2D list representing the grid after the object is moved to the right.  

\addvspace{10pt}

\begin{lstlisting}
def move_right(state, object): 
     move_state = copy.deepcopy(state) 
     new_obj = []   
     for x, y in object: 
          move_state[x][y] = 0   
     for x, y in object: 
          new_x, new_y = x, y + 1 
          if 0 <= new_x < len(state) and 0 <= new_y < len(state[0]): 
               move_state[new_x][new_y] = state[x][y] 
               new_obj.append([new_x, new_y])   
     return move_state 
\end{lstlisting}
\end{tcolorbox}

\begin{tcolorbox}[breakable, colback=gray!5!white, colframe=gray!75!black, title=DSL Prompt: Move Left]
\# move\_left function moves all pixels in the selected object to the left by one column. 

\# Parameters: 

\# - state: A 2D list representing the current grid state. 

\# - object: A list of lists where each inner list contains the coordinates [x, y] of a pixel to move. 

\# Returns: 

\# - A new 2D list representing the grid after the object is moved to the left.  

\addvspace{10pt}

\begin{lstlisting}
def move_left(state, object): 
     move_state = copy.deepcopy(state) 
     new_obj = []   
     for x, y in object: 
          move_state[x][y] = 0   
     for x, y in object: 
          new_x, new_y = x, y - 1 
          if 0 <= new_x < len(state) and 0 <= new_y < len(state[0]): 
               move_state[new_x][new_y] = state[x][y] 
               new_obj.append([new_x, new_y])   
     return move_state 
\end{lstlisting}
\end{tcolorbox}

\begin{tcolorbox}[breakable, colback=gray!5!white, colframe=gray!75!black, title=DSL Prompt: Move Up]
\# move\_up function moves all pixels in the selected object up by one row.  

\# Parameters: 

\# - state: A 2D list representing the current grid state. 

\# - object: A list of lists where each inner list contains the coordinates [x, y] of a pixel to move. 

\# Returns: 

\# - A new 2D list representing the grid after the object is moved up.  

\addvspace{10pt}

\begin{lstlisting}
def move_up(state, object): 
     move_state = copy.deepcopy(state) 
     new_obj = []  
     for x, y in object: 
          move_state[x][y] = 0  
     for x, y in object: 
          new_x, new_y = x-1, y 
          if 0 <= new_x < len(state) and 0 <= new_y < len(state[0]):  
               move_state[new_x][new_y] = state[x][y]  
               new_obj.append([new_x, new_y])   
     return move_state 
\end{lstlisting}
\end{tcolorbox}

\begin{tcolorbox}[breakable, colback=gray!5!white, colframe=gray!75!black, title=DSL Prompt: Move Down]
\# move\_down function moves all pixels in the selected object down by one row. 

\# Parameters: 

\# - state: A 2D list representing the current grid state.  

\# - object: A list of lists where each inner list contains the coordinates [x, y] of a pixel to move. 

\# Returns: 

\# - A new 2D list representing the grid after the object is moved down. 

\addvspace{10pt}

\begin{lstlisting}
def move_down(state, object): 
    move_state = copy.deepcopy(state) 
    new_obj = []   
    for x, y in object:  
        move_state[x][y] = 0   
    for x, y in object: 
        new_x, new_y = x + 1, y 
        if 0 <= new_x < len(state) and 0 <= new_y < len(state[0]): 
            move_state[new_x][new_y] = state[x][y] 
            new_obj.append([new_x, new_y])   
    return move_state 
\end{lstlisting}
\end{tcolorbox}

\begin{tcolorbox}[breakable, colback=gray!5!white, colframe=gray!75!black, title=DSL Prompt: Rotate Right Object]
\# rotate\_right\_obj function makes a clockwise rotation about the given object. 

\# This function rotates the selected object within the grid 90 degrees clockwise around its center. 

\# Parameters: 

\# - state: A 2D list representing the current grid state. 

\# - object: A list of lists where each inner list contains the coordinates [x, y] of a pixel in the object. 

\# Returns: 

\# - A new 2D list representing the grid after the object is rotated clockwise.  

\addvspace{10pt}

\begin{lstlisting}
def rotate_right_obj(state, object): 
    rotate_state = copy.deepcopy(state) 
    new_obj = [] 
    max_x = max(x for x,_ in object) 
    min_x = min(x for x,_ in object) 
    max_y = max(y for_, y in object) 
    min_y = min(y for_, y in object) 
    fixed_x = (max_x + min_x) // 2 
    fixed_y = (max_y + min_y) // 2 

    for x, y in object: 
        rotate_state[x][y] = 0 

    for x, y in object: 
        moved_x = y - fixed_y + fixed_x 
        moved_y = -x + fixed_x + fixed_y 
        if 0 <= moved_x < len(state) and 0 <= moved_y < len(state[0]): 
            rotate_state[moved_x][moved_y] = state[x][y] 
            new_obj.append([moved_x, moved_y]) 

    return rotate_state 
\end{lstlisting}
\end{tcolorbox}

\begin{tcolorbox}[breakable, colback=gray!5!white, colframe=gray!75!black, title=DSL Prompt: Rotate Left Object]
\# rotate\_left\_obj function makes a counterclockwise rotation about the given object. 

\# This function rotates the selected object within the grid 90 degrees counterclockwise around its center. 

\# Parameters: 

\# - state: A 2D list representing the current grid state. 

\# - object: A list of lists where each inner list contains the coordinates [x, y] of a pixel in the object. 

\# Returns: 

\# - A new 2D list representing the grid after the object is rotated counterclockwise.  

\addvspace{10pt}

\begin{lstlisting}
def rotate_left_obj(state, object): 
    rotate_state = copy.deepcopy(state) 
    new_obj = [] 
    max_x = max(x for x,_ in object) 
    min_x = min(x for x,_ in object) 
    max_y = max(y for_, y in object) 
    min_y = min(y for_, y in object) 
    fixed_x = (max_x + min_x) // 2 
    fixed_y = (max_y + min_y) // 2 

    for x, y in object: 
        rotate_state[x][y] = 0 

    for x, y in object: 
        moved_x = -y + fixed_y + fixed_x 
        moved_y = x - fixed_x + fixed_y 
        if 0 <= moved_x < len(state) and 0 <= moved_y < len(state[0]): 
            rotate_state[moved_x][moved_y] = state[x][y] 
            new_obj.append([moved_x, moved_y]) 

    return rotate_state 
\end{lstlisting}
\end{tcolorbox}

\begin{tcolorbox}[breakable, colback=gray!5!white, colframe=gray!75!black, title=DSL Prompt: Vertical Flip Object]
\# vertical\_flip\_obj function makes a vertical flip of the selected object.

\# This function flips the selected object within the grid vertically. 

\# Parameters: 

\# - state: A 2D list representing the current grid state. 

\# - object: A list of lists where each inner list contains the coordinates [x, y] of a pixel in the object. 

\# Returns: 

\# - A new 2D list representing the grid after the object is flipped vertically.  

\addvspace{10pt}

\begin{lstlisting}
def vertical_flip_obj(state, object): 
    flip_state = copy.deepcopy(state) 
    new_obj = [] 
    max_x = max(x for x,_ in object) 
    min_x = min(x for x,_ in object) 

    for x, y in object: 
        flip_state[x][y] = 0 

    for x, y in object: 
        flip_state[max_x + min_x - x][y] = state[x][y] 
        new_obj.append([max_x + min_x - x, y]) 

    return flip_state 
\end{lstlisting}
\end{tcolorbox}

\begin{tcolorbox}[breakable, colback=gray!5!white, colframe=gray!75!black, title=DSL Prompt: Horizontal Flip Object]
\# horizontal\_flip\_obj function makes a horizontal flip of the selected object. 

\# This function flips the selected object within the grid horizontally. 

\# Parameters: 

\# - state: A 2D list representing the current grid state. 

\# - object: A list of lists where each inner list contains the coordinates [x, y] of a pixel in the object. 

\# Returns: 

\# - A new 2D list representing the grid after the object is flipped horizontally.  

\addvspace{10pt}

\begin{lstlisting}
def horizontal_flip_obj(state, object): 
    flip_state = copy.deepcopy(state) 
    new_obj = [] 
    max_y = max(y for_, y in object) 
    min_y = min(y for_, y in object) 

    for x, y in object: 
        flip_state[x][y] = 0 

    for x, y in object: 
        flip_state[x][max_y + min_y - y] = state[x][y] 
        new_obj.append([x, max_y + min_y - y]) 

    return flip_state 
\end{lstlisting}
\end{tcolorbox}

\begin{tcolorbox}[breakable, colback=gray!5!white, colframe=gray!75!black, title=DSL Prompt: X Line]
\# X\_line function makes a diagonal X-line in all directions from a given pixel until they reach the end of the grid. 

\# Parameters:

\# - state: A 2D list representing the current grid state. 

\# - r: The row index of the starting pixel. 

\# - c: The column index of the starting pixel. 

\# - color: The color to be used for the X-line. 

\# Returns: 

\# - A new 2D list representing the grid after the X-line is drawn.  

\addvspace{10pt}

\begin{lstlisting}
def X_line(state, r, c, color): 
    X_state = copy.deepcopy(state) 
    x_move = [-1, 1] 
    y_move = [-1, 1] 

    for i in x_move: 
        for j in y_move: 
            moved_x, moved_y = r + i, c + j 
            while 0 <= moved_x < len(state) and 0 <= moved_y < len(state[0]): 
                X_state[moved_x][moved_y] = color 
                moved_x += i 
                moved_y += j 

    return X_state 
\end{lstlisting}
\end{tcolorbox}

\begin{tcolorbox}[breakable, colback=gray!5!white, colframe=gray!75!black, title=DSL Prompt: Horizontal Line]
\# horizontal\_line function draws a horizontal line between two pixels if they are on the same row. 

\# Parameters: 

\# - state: A 2D list representing the current grid state. 

\# - r1: The row index of the first pixel. 

\# - c1: The column index of the first pixel. 

\# - r2: The row index of the second pixel. 

\# - c2: The column index of the second pixel. 

\# - color: The color to be used for the line. 

\# Returns: 

\# - A new 2D list representing the grid after the horizontal line is drawn.  

\addvspace{10pt}

\begin{lstlisting}
def horizontal_line(state, r1, c1, r2, c2, color): 
    line_state = copy.deepcopy(state) 
    if r1 == r2: 
        if c1 < c2: 
            if c2 < len(state[0]): 
                for i in range(c1+1, c2): 
                    line_state[r1][i] = color 
        else: 
            if c1 < len(state[0]): 
                for i in range(c2+1, c1): 
                    line_state[r1][i] = color 
    return line_state 
\end{lstlisting}
\end{tcolorbox}

\begin{tcolorbox}[breakable, colback=gray!5!white, colframe=gray!75!black, title=DSL Prompt: Vertical Line]
\# vertical\_line function draws a vertical line between two pixels if they are in the same column. 

\# Parameters: 

\# - state: A 2D list representing the current grid state. 

\# - r1: The row index of the first pixel. 

\# - c1: The column index of the first pixel. 

\# - r2: The row index of the second pixel. 

\# - c2: The column index of the second pixel. 

\# - color: The color to be used for the line. 

\# Returns: 

\# - A new 2D list representing the grid after the vertical line is drawn.  

\addvspace{10pt}

\begin{lstlisting}
def vertical_line(state, r1, c1, r2, c2, color): 
    line_state = copy.deepcopy(state) 
    if c1 == c2: 
        if r1 < r2: 
            if r2 < len(state): 
                for i in range(r1+1, r2): 
                    line_state[i][c1] = color 
        else: 
            if r1 < len(state): 
                for i in range(r2+1, r1): 
                    line_state[i][c1] = color 
    return line_state 
\end{lstlisting}
\end{tcolorbox}

\begin{tcolorbox}[breakable, colback=gray!5!white, colframe=gray!75!black, title=DSL Prompt: Diagonal Line]
\# diagonal\_line function draws a diagonal line between two pixels if they form a 45-degree angle. 

\# Parameters: 

\# - state: A 2D list representing the current grid state. 

\# - r1: The row index of the first pixel. 

\# - c1: The column index of the first pixel. 

\# - r2: The row index of the second pixel. 

\# - c2: The column index of the second pixel. 

\# - color: The color to be used for the line. 

\# Returns: 

\# - A new 2D list representing the grid after the diagonal line is drawn. 

\addvspace{10pt}

\begin{lstlisting}
def diagonal_line(state, r1, c1, r2, c2, color): 
    line_state = copy.deepcopy(state) 
    if abs(r1 - r2) == abs(c1 - c2): 
        dr = 1 if r2 > r1 else -1 
        dc = 1 if c2 > c1 else -1 
        r, c = r1 + dr, c1 + dc 
        while r != r2 and c != c2: 
            line_state[r][c] = color 
            r += dr 
            c += dc 
    return line_state 
\end{lstlisting}
\end{tcolorbox}

\begin{tcolorbox}[breakable, colback=gray!5!white, colframe=gray!75!black, title=DSL Prompt: Object Color]
\# obj\_color function changes the color of the selected object. 

\# Parameters: 

\# - state: A 2D list representing the current grid state. 

\# - object: A list of lists where each inner list contains the coordinates [x, y] of a pixel in the object. 

\# - color: The new color to be applied to the object. 

\# Returns: 

\# - A new 2D list representing the grid after the object's color is changed.  

\addvspace{10pt}

\begin{lstlisting}
def obj_color(state, object, color): 
    color_state = copy.deepcopy(state) 
    for x, y in object: 
        color_state[x][y] = color 
    return color_state 
\end{lstlisting}
\end{tcolorbox}

\begin{tcolorbox}[breakable, colback=gray!5!white, colframe=gray!75!black, title=DSL Prompt: Pixel Color]
\# pixel\_color function changes the color of the selected pixel. 

\# Parameters: 

\# - state: A 2D list representing the current grid state. 

\# - r: The row index of the pixel to change. 

\# - c: The column index of the pixel to change. 

\# - color: The new color to be applied to the pixel.

\# Returns: 

\# - A new 2D list representing the grid after the pixel's color is changed.  

\addvspace{10pt}

\begin{lstlisting}
def pixel_color(state, r, c, color): 
    temp_state = copy.deepcopy(state) 
    temp_state[r][c] = color 
    return temp_state 
\end{lstlisting}
\end{tcolorbox}

\begin{tcolorbox}[breakable, colback=gray!5!white, colframe=gray!75!black, title=DSL Prompt: Complete]
\# complete function returns the current state as the final answer of the quiz. 

\# Parameters:

\# - state: A 2D list representing the current grid state. 

\# Returns: 

\# - The same 2D list representing the final grid state.  

\addvspace{10pt}

\begin{lstlisting}
def complete(state): 
    return state 
\end{lstlisting}
\end{tcolorbox}

\subsubsection{Prompt of Compositionality Experiment}

Both the LLMs DSL understanding and compositionality of LLMs experiments utilized the prompt structure outlined in Section~\ref{sec:appendix_compositionality_section_detailed_promptings}. The Introduction ARC Prompt provides a comprehensive overview of ARC, while the DSL Usage Example Prompt illustrates DSL application. The DSL prompt, comprising the prompts of DSL function codes and comments from Section~\ref{sec:appendix_compositionality_section_detailed_dsl_promptings} and the DSL usage example prompt, offers a comprehensive DSL explanation. The task prompt includes demonstration examples, test input, object information (coordinates of objects obtained through PnP in dictionary format), and output format guidelines. In the case of the LLMs DSL understanding prompt, unlike the task prompt, the DSLs path for the task is provided. The CoT prompt included the introduction ARC prompt and DSL prompt. In the case of the LLMs DSL understanding experiment, the LLMs DSL understanding prompt was used, while in the case of the compositionality of LLMs experiment, the task prompt was used. In the compositionality experiments, the CoT prompt was utilized.

% LLMs DSL understanding 실험과 Compositionality of LLMS 실험 모두 LLM에게 Table~\ref{tbl:appendix_compositionality_prompt_setting}에 있는 prompt를 사용하였다. Introduction ARC Prompt는 ARC에 대한 전반적인 설명을 담고 있으며 DSL Usage Example Prompt는 DSL 사용 예시에 대한 것이다. DSL Prompt는 Table~\ref{tbl:appendix_compositionality_dsl_function_prompt}의 prompts인 Prompts of DSL Function Codes and Comments과 DSL Usage Example Prompt를 사용하여 DSL에 대한 전반적인 설명이 담겨져있다. Task Prompt는 주어진 task에 대한 demostraction examples와 test input, obejct info 그리고 LLM이 output이 어떤 형태인지를 알려주는 내용이 담겨있다. object info는 해당 grid에서 PnP를 통해 얻은 object의 coordination이 dictionary형태로 담겨져 있다.마지막으로 CoT Prompt는 Introduction ARC Prompt, DSL Prompt, Task Prompt가 합쳐진 것으로 Compositionality 실험들에서는 CoT Prompt를 사용하였다.

\subsubsection{Detailed Promptings}
\label{sec:appendix_compositionality_section_detailed_promptings}

Composition of prompt contents used in the compositionality experiments.

\begin{tcolorbox}[breakable, colback=gray!5!white, colframe=gray!75!black, title=Introduction ARC Prompt, width=\textwidth]
Do you know ARC problem?  

\addvspace{10pt}

ARC is a quiz, and if we can solve this problem, we can understand and utilize several concepts such as ``object", ``count", ``color", ``move", ``row", ``column", etc.  

\addvspace{10pt}

ARC problems give you some examples to understand these patterns. You can understand the pattern below with several examples and then apply the quiz's input to get the correct output.  
\end{tcolorbox}

\begin{tcolorbox}[breakable, colback=gray!5!white, colframe=gray!75!black, title=DSL Usage Example Prompt, width=\textwidth]
In this example grid,  

[[0, 0, 0, 0, 0, 0, 0, 0, 0, 0],  

 [0, 0, 0, 0, 0, 0, 0, 5, 5, 0],  

 [0, 5, 5, 0, 0, 0, 0, 5, 5, 0],  

 [0, 0, 5, 5, 0, 0, 0, 0, 0, 0],  

 [0, 0, 0, 0, 0, 0, 0, 0, 0, 0],  

 [0, 0, 0, 0, 0, 0, 0, 0, 0, 5],  

 [0, 0, 0, 0, 0, 5, 5, 0, 0, 5],  

 [0, 5, 0, 0, 0, 0, 0, 0, 0, 5],  

 [0, 5, 0, 0, 5, 0, 0, 0, 0, 0],  

 [0, 0, 0, 5, 5, 0, 0, 0, 0, 0]]  

\addvspace{10pt}

there are 6 objects  

Object1: [[1, 7], [1, 8], [2, 7], [2, 8]]  

Object2: [[2, 1], [2, 2], [3, 2], [3, 3]]  

Object3: [[5, 9], [6, 9], [7, 9]]  

Object4: [[6, 5], [6, 6]]  

Object5: [[7, 1], [8, 1]]  

Object6: [[8, 4], [9, 3], [9, 4]]  

\addvspace{10pt}

If you apply ``rotate\_right\_obj(state, object2)", the result becomes  

[[0, 0, 0, 0, 0, 0, 0, 0, 0, 0],  

 [0, 0, 5, 0, 0, 0, 0, 5, 5, 0],  

 [0, 5, 5, 0, 0, 0, 0, 5, 5, 0],  

 [0, 5, 0, 0, 0, 0, 0, 0, 0, 0],  

 [0, 0, 0, 0, 0, 0, 0, 0, 0, 0],  

 [0, 0, 0, 0, 0, 0, 0, 0, 0, 5],  

 [0, 0, 0, 0, 0, 5, 5, 0, 0, 5],  

 [0, 5, 0, 0, 0, 0, 0, 0, 0, 5],  

 [0, 5, 0, 0, 5, 0, 0, 0, 0, 0],  

 [0, 0, 0, 5, 5, 0, 0, 0, 0, 0]]  
\end{tcolorbox}

\begin{tcolorbox}[breakable, colback=gray!5!white, colframe=gray!75!black, title=DSL Prompt, width=\textwidth]
\textbf{\{Prompts of DSL Function Codes and Comments\}}  

\addvspace{10pt}

Arguments for the DSLs are mainly ``state" and ``object", but some require ``color", ``row", ``column", etc. ``state" is the current state of the grid, which is the entire grid.  

\addvspace{10pt}

\textbf{"object"} is the list of coordinates of the object; there may be multiple objects in the grid, but no DSL requires multiple objects. \textbf{``color"} is the color of the pixel in the grid, which is a number between 0 and 9. \textbf{``row"} and \textbf{``column"} are the coordinate numbers of a pixel in the grid.  

\addvspace{10pt}

You can choose from here and apply the DSL to solve the problem. You must input the appropriate arguments to the DSL, or it will not work.  

\addvspace{10pt}

\textbf{\{DSL Usage Example Prompt\}}  

\addvspace{10pt}

Please choose the DSL from the list above and provide the proper arguments to solve the problem.  
\end{tcolorbox}

\begin{tcolorbox}[breakable, colback=gray!5!white, colframe=gray!75!black, title=Task Prompt, width=\textwidth]
\textbf{\{Demo Examples\}} \& \textbf{\{Test Input\}}  

\addvspace{10pt}

You must solve the given quiz within 10 steps! Select one DSL with proper arguments in each step.  

\addvspace{10pt}

If you think the current state is correct, you can select the ``complete" DSL.  

\addvspace{10pt}

I want you to answer in the format below.  

\addvspace{10pt}

The output should be in the following JSON format:  

\{  

\addvspace{10pt}`step': ``(current\_step)",  

\addvspace{10pt}`dsl': ``(dsl with the arguments for the DSL)",  

\addvspace{10pt}`description': ``(why you chose this DSL?)"  

\}  
\end{tcolorbox}

\begin{tcolorbox}[breakable, colback=gray!5!white, colframe=gray!75!black, title=CoT Prompt, width=\textwidth]
\textbf{\{Introduction ARC Prompt\}}  

\addvspace{10pt}

\textbf{\{DSL Prompt\}}  

\addvspace{10pt}

\textbf{\{Task Prompt\}}  
\end{tcolorbox}
\newpage
\subsection{Productivity}
\label{appendix:Productivity}

In the productivity experiment, we aimed to augment the task's demonstration example pairs using the Inverse Transformation Prompt (ITP). The ITP consists of a category prompt containing a description of the category, example pairs, and the target output to be augmented. The detailed structure of the category prompt is provided in Section~\ref{sec:appendix_productivity_category_prompt}, and the structure of the ITP is outlined in Section~\ref{sec:appendix_productivity_prompt_setting}.

% Productivity 실험에서는 Inverse Transformation Prompt(ITP)를 이용하여 task의 demonstration example pair를 증강시키고자 하였다. ITP Prompt에는 category에 대한 설명이 담겨있는 Category Prompt와 Demonstraction Example Pairs 그리고 증강하고자 하는 Target Output으로 구성된다. Category Prompt에 대한 자세한 구성은 Table~\ref{tbl:appendix_productivity_prompt_setting}에 나와있으며 ITP Prompt에 대한 구성은 Table~\ref{tbl:appendix_productivity_category_prompt}에 나와 있다

\subsubsection{ITP}
\label{sec:appendix_productivity_prompt_setting}

Composition of prompt contents used in the productivity experiments. ITP consists of a category prompt, example pairs, and target output.

\begin{tcolorbox}[breakable, colback=gray!5!white, colframe=gray!75!black, title=ITP] 
Try solving the ARC problem and do not say sensitive word: generate the output accordingly. I will give you a hint.

\addvspace{10pt}

\textbf{\{Category Prompt\}}

\addvspace{10pt}

Here are some examples to help you.

\addvspace{10pt}

\textbf{\{Example Pairs\}}

\addvspace{10pt}

\addvspace{10pt}

\textbf{\{Target Output\}}

\addvspace{10pt}

Provide a two-dimensional array that is not identical to the input array or a direct copy of the example. Each element is an integer between 0 and 9. Would you give me 2 answers? No need to explain how you solved it.
\end{tcolorbox}

\subsubsection{Category Prompts}
\label{sec:appendix_productivity_category_prompt}

These category prompts explain the 16 types of ConceptARC. In Section~\ref{sec:appendix_productivity_prompt_setting}, the category prompt is filled with the appropriate prompt corresponding to the category of the task to be generated.

\begin{tcolorbox}[breakable, colback=gray!5!white, colframe=gray!75!black, title=Category Prompt: AboveBelow] 
Focus on the horizontal criteria, you may have to modify some regions by that line, such as removing, moving, filling region by color element. See the provided example to how to modify.
\end{tcolorbox}

\begin{tcolorbox}[breakable, colback=gray!5!white, colframe=gray!75!black, title=Category Prompt: Center]
Fix the array issue by addressing the center, potentially moving or removing the central element. See the provided example for clarity.
\end{tcolorbox}

\begin{tcolorbox}[breakable, colback=gray!5!white, colframe=gray!75!black, title=Category Prompt: CleanUp]
Distort the shapes in areas where they are polygonal or completely filled, adding noise or disturbances to disrupt the complete shapes.
\end{tcolorbox}

\begin{tcolorbox}[breakable, colback=gray!5!white, colframe=gray!75!black, title=Category Prompt: CompleteShape]
Distort the perfectly shaped objects identified in the input image. Introduce noise to these identified objects to easily generate diverse outputs.
\end{tcolorbox}

\begin{tcolorbox}[breakable, colback=gray!5!white, colframe=gray!75!black, title=Category Prompt: Copy]
Delete one identical object from the output. Refer to the example to identify which one to remove. Consider deleting the object located in a position-indicating space.
\end{tcolorbox}

\begin{tcolorbox}[breakable, colback=gray!5!white, colframe=gray!75!black, title=Category Prompt: Count]
Create an input image based on the provided count-related problem. Focus on details like object or color count, as shown in the example.
\end{tcolorbox}

\begin{tcolorbox}[breakable, colback=gray!5!white, colframe=gray!75!black, title=Category Prompt: ExtendToBoundary]
In the input image, find lines connected to boundaries with different colors. Transform these lines into a different shape. The example illustrates how to make this transformation.
\end{tcolorbox}

\begin{tcolorbox}[breakable, colback=gray!5!white, colframe=gray!75!black, title=Category Prompt: ExtractObjects]
Generate an output image with objects from the given input. Refer to examples for guidance. Hint: Extract objects when inferring input from output.
\end{tcolorbox}

\begin{tcolorbox}[breakable, colback=gray!5!white, colframe=gray!75!black, title=Category Prompt: FilledNotFilled]
When inferring the input from the output, focus on situations where the inner part of an object contains empty space or another object. Examples provide guidance for creating the output image.
\end{tcolorbox}

\begin{tcolorbox}[breakable, colback=gray!5!white, colframe=gray!75!black, title=Category Prompt: HorizontalVertical]
Focus on horizontal and vertical relations, representing them with colors or preserving one direction while eliminating the other. Examples illustrate the approach.
\end{tcolorbox}

\begin{tcolorbox}[breakable, colback=gray!5!white, colframe=gray!75!black, title=Category Prompt: InsideOutside]
Address the inside-outside relationship, either by selecting items inside or outside in the input or determining quantities. Use the boundary as a reference. Examples offer guidance.
\end{tcolorbox}

\begin{tcolorbox}[breakable, colback=gray!5!white, colframe=gray!75!black, title=Category Prompt: MoveToBoundary]
Objects in the input may be shifted to one side, and in the output, they are displaced either horizontally or vertically. Infer the direction from examples and choose the displacement freely.
\end{tcolorbox}

\begin{tcolorbox}[breakable, colback=gray!5!white, colframe=gray!75!black, title=Category Prompt: Order]
This is about randomly rearranging initially ordered objects while representing their original positions through a specific rule. Examine the examples to understand how to achieve this.
\end{tcolorbox}

\begin{tcolorbox}[breakable, colback=gray!5!white, colframe=gray!75!black, title=Category Prompt: SameDifferent]
You'll notice that only specific-shaped objects are extracted in the input image. Create additional objects in the zero-represented space. Examples provide guidance on how to proceed.
\end{tcolorbox}

\begin{tcolorbox}[breakable, colback=gray!5!white, colframe=gray!75!black, title=Category Prompt: TopBottom2D]
Objects are in a 2D space. Check changes in the top and bottom. The input may have shifted or require removing top/bottom indicators. Look at examples for specifics.
\end{tcolorbox}

\end{document}